\documentclass{article}
\usepackage{lmodern}
\PassOptionsToPackage{table}{xcolor}
\usepackage{iclr2027_conference,times}
\definecolor{citecolor}{HTML}{0071bc}
\definecolor{refcolor}{HTML}{750014}
\usepackage[pagebackref=false,breaklinks=true,colorlinks,citecolor=citecolor,linkcolor=refcolor,bookmarks=false]{hyperref}
\usepackage{url}
\ifdefined\LaTeXML\else\usepackage{fontawesome5}\fi
\usepackage{booktabs}
\usepackage{multirow}
\usepackage{graphicx}
\usepackage{wrapfig}

\usepackage{amsmath,amsfonts,bm}

\def\eqref#1{equation~\ref{#1}}

\def\1{\bm{1}}

\DeclareMathAlphabet{\mathsfit}{\encodingdefault}{\sfdefault}{m}{sl}
\SetMathAlphabet{\mathsfit}{bold}{\encodingdefault}{\sfdefault}{bx}{n}

\usepackage{amsthm}
\newtheorem{lemma}{Lemma}
\newcommand{\ms}[2]{#1{\scriptsize$\pm#2$}}
\newcommand{\cbar}{\ensuremath{\overline{\mathrm{C}}}}
\definecolor{oursgreen}{HTML}{D2EBD2}
\definecolor{deltablue}{HTML}{E6F1FB}
\definecolor{groupgrey}{HTML}{ECECEC}
\usepackage{tabularx}
\usepackage{enumitem}
\newcolumntype{R}{>{\raggedleft\arraybackslash}X}
\newcolumntype{G}{>{\columncolor{oursgreen}\raggedleft\arraybackslash}X}
\newcolumntype{D}{>{\columncolor{deltablue}\boldmath\raggedleft\arraybackslash}X}

\iclrfinalcopy  %

\title{Later Is Better: Token Reduction for ViTs\\ Under Distribution Shift}

\ifdefined\LaTeXML
\author{Hyeongheon Cha \\ KAIST \\ \texttt{hyeongheon@kaist.ac.kr}
\and Hyungjun Yoon \\ KAIST \\ \texttt{hyungjun.yoon@kaist.ac.kr}
\and Sung-Ju Lee \\ KAIST \\ \texttt{profsj@kaist.ac.kr}}
\date{\href{https://github.com/chahh9808/LaterIsBetter}{\includegraphics[height=1em,alt={GitHub}]{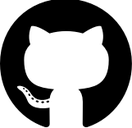}~\texttt{chahh9808/LaterIsBetter}}}
\else
\author{Hyeongheon Cha\textsuperscript{1} \quad Hyungjun Yoon\textsuperscript{1} \quad Sung-Ju Lee\textsuperscript{1} \\
 \textsuperscript{1} KAIST \\
 \texttt{\{hyeongheon,hyungjun.yoon,profsj\}@kaist.ac.kr} \\
 \href{https://github.com/chahh9808/LaterIsBetter}{\textcolor{black}{\faGithub\ \texttt{chahh9808/LaterIsBetter}}}
 }
\fi

\begin{document}
\maketitle
 \vspace{-10pt}

\begin{abstract}
Training-free token reduction accelerates vision transformers by removing redundant tokens across layers, recovering most of the original accuracy at a fraction of the compute. These methods, however, are designed and evaluated primarily on clean data, and under real-world distribution shift their accuracy gap to the uncompressed model widens with the removal rate. We show that this gap is governed by the reduction \emph{schedule}, the depth profile of removal, usually left fixed as an implementation detail. Concretely, we introduce a one-parameter \emph{late-concentrated} power-law schedule that consistently improves out-of-distribution accuracy over flat at no extra inference cost. On ImageNet-C with DeiT-S, the late schedule closes $83\%$ of that gap at a $26\%$ compute reduction ($+1.17$pp), and $99\%$ of it at a lighter $7\%$ reduction ($+0.26$pp). The gain cannot be attributed to retaining more tokens or using extra compute: held to flat's compute, the late schedule removes more tokens in total and leaves fewer tokens at the end, yet still wins. Single-layer probes point to a mechanism: earlier reductions perturb features that pass through more remaining layers, front-loading reduction error in depth. The effect is broad, holding across five token-reduction methods (ToMe, EViT, ATS, ATC, PiToMe), nine backbones, all ImageNet-C corruption types, eight further shift suites, and two further modalities, video and vision-language QA. It is also specific to shift, still positive on clean and rising monotonically to $\sim$4$\times$ that at the highest severity~5. The schedule keeps its gain under six test-time adaptation methods, and needs no per-input or per-domain tuning.
\end{abstract}

\section{Introduction}

Vision transformers (ViTs)~\citep{dosovitskiy2021vit} split an image into patches, embed each as a token, and process the resulting sequence through a stack of transformer layers. Their compute scales with the number of tokens, steeply so in self-attention, whose cost is quadratic in the sequence length. Because many of these tokens are redundant, token-reduction methods cut this cost by removing tokens as the image passes through the network, shrinking the sequence the deeper layers must process: pruning unimportant tokens~\citep{liang2022evit}, merging similar ones~\citep{bolya2023tome,tran2024pitome}, clustering them into groups~\citep{haurum2024atc}, or sampling a subset to keep~\citep{fayyaz2022ats}, all recovering most of the original accuracy at a fraction of the compute. 
Because they require no retraining and act at test time, with no extra data, they are widely studied as plug-in efficiency tools for existing ViT pipelines.

\begin{figure}[t]
\centering
\vspace{-25pt}
\includegraphics[width=0.88\linewidth]{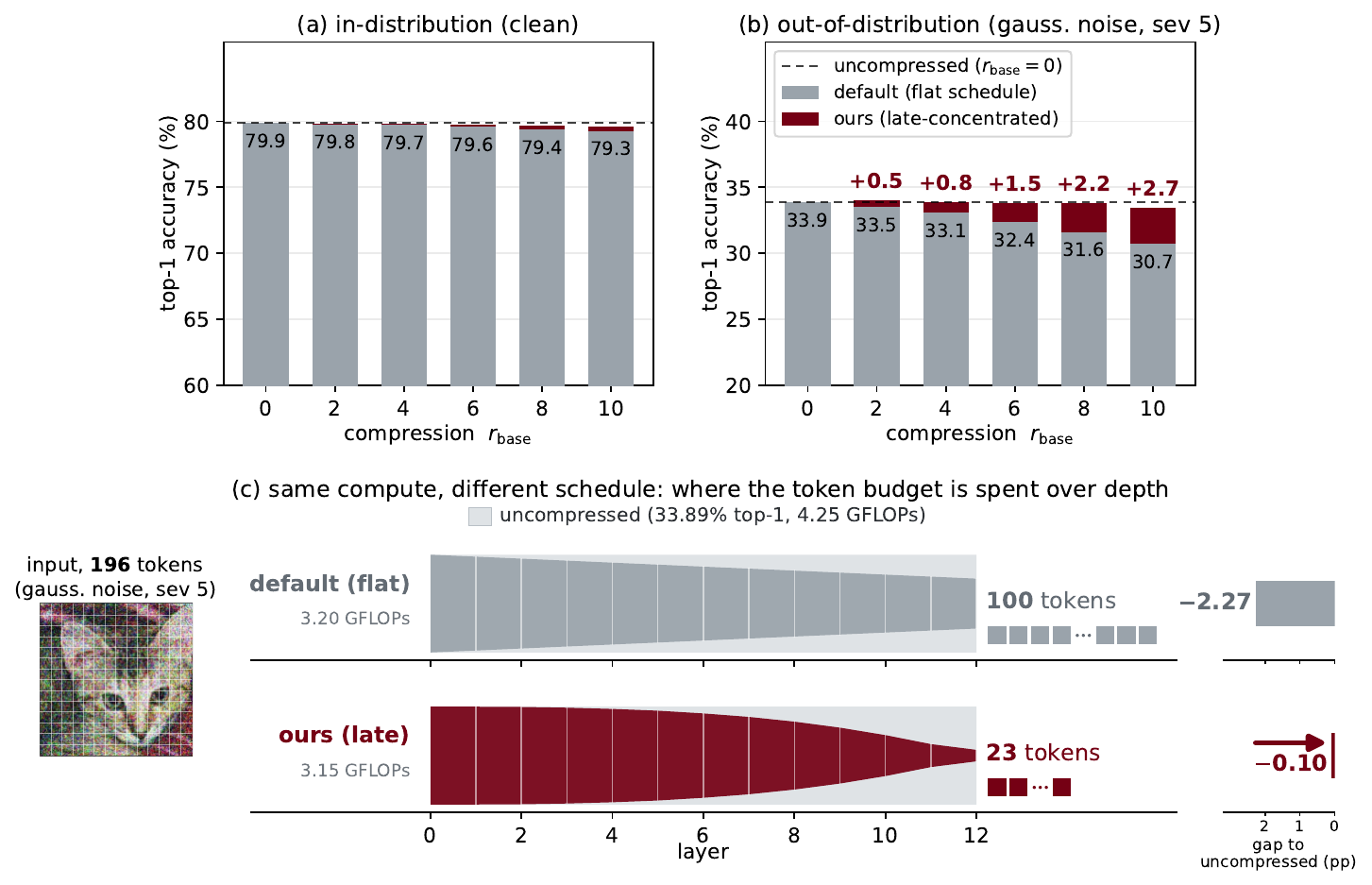}
\vspace{-14pt}
\caption{Token reduction increasingly sacrifices out-of-distribution accuracy; a
\emph{late-concentrated schedule} recovers it at \emph{no extra compute}. \textbf{(a,\,b)} DeiT-S
top-1 vs.\ the per-layer token removal rate $r_{\mathrm{base}}$ on clean and
gaussian-noise ImageNet-C. \textbf{(c)} each ribbon's height is the tokens remaining, so the pale band above
is what was removed; flat removes the same number at every layer, ours follows the late-concentrated
power law. Ours ends with far
fewer tokens, yet closer to uncompressed.}
\label{fig:compdrop}
\vspace{-10pt}
\end{figure}

These methods, however, are designed and evaluated primarily on clean, in-distribution images, while the deployments they target rarely see such inputs. Sensor noise, motion blur, weather, and compression artifacts produce a distribution shift away from the training data~\citep{hendrycks2019imagenetc,mintun2021cbar,kar20223dcc}. Whether token reduction, tuned for clean efficiency, retains its accuracy under such real-world shift is largely unexamined. Figure~\ref{fig:compdrop}a,b answers this for the widely used ToMe~\citep{bolya2023tome}: the more aggressive the reduction, the more accuracy it sacrifices under shift, while clean accuracy is barely touched.

The remedy lies in a choice these methods leave implicit. A token-reduction method is defined not only by how many tokens it removes, but by \emph{where} along the network's depth it removes them, that is, its \emph{schedule}. Existing methods largely use a flat schedule that removes the same number at every layer~\citep{bolya2023tome,haurum2024atc}, treated as a fixed implementation detail, with tuning effort targeting the per-layer rate or input-adaptive rules rather than the depth profile of removal itself. The deployment shift is unknown in advance and unlabeled at test time. We therefore ask whether, in this training-free regime, the schedule is itself a lever for robustness at matched inference cost.

We find that it is, and that concentrating removal \emph{late} is the direction that helps, while the mirror-image early-concentrated schedule is the most damaging of the three. The probes in Section~\ref{sec:mechtoken} point to amplification: a token-reduction step perturbs the features it acts on, and the earlier it acts, the more layers amplify that perturbation before it reaches the classifier. Removing later therefore distorts the output less, and shift widens the gap by raising that amplification. Concretely, we propose a \emph{late-concentrated} power-law schedule, a per-layer removal rate that increases with depth (Eq.~\ref{eq:schedule}, Figure~\ref{fig:compdrop}c). We use a single fixed exponent at our main operating point, with no per-input or per-domain tuning. On DeiT-S and ImageNet-C this recovers $83\%$ of the accuracy that token reduction loses under shift at a $26\%$ compute reduction ($+1.17$pp over flat), and $99\%$ of it at a lighter $7\%$ reduction ($+0.26$pp). It also removes more tokens in total than flat, yet still improves accuracy, so neither a larger token budget nor extra compute can account for the gain.

The benefit holds across five training-free reducers and nine backbones from four pretraining recipes, all ImageNet-C corruption types, eight further shift suites (ImageNet-A/R/V2/Sketch, DomainNet-126, ImageNet-\cbar{}/3DCC, NaBirds-C), and two modalities beyond image classification, video on Kinetics and vision-language QA on GQA-C. The schedule also composes with six test-time adaptation methods, which adapt the model to the same shift. And the gain rises monotonically with corruption severity, $\sim$4$\times$ from a small $+0.28$pp on clean images to $+1.17$pp at severity~5, so it is a robustness property rather than a general accuracy improvement.

We make the following contributions:
\begin{itemize}[leftmargin=0.7cm]
  \item We identify the reduction schedule (the depth profile of removal) as an under-characterized
    determinant of ViT robustness to distribution shift, and show that at matched compute a
    late-concentrated power law recovers most of the accuracy that reduction costs under shift.
  \item We provide a first-order diagnostic account of the gain by measuring per-layer distortion and
    perturbation amplification, finding both to be front-loaded in depth.
  \item We show the effect is broad, holding across five training-free reducers, nine backbones,
    all ImageNet-C corruptions, eight further shift suites, two other modalities, growing
    with shift severity, and composing with six test-time adaptation methods.
\end{itemize}

\section{Related Work}
\label{app:related}

\paragraph{Token reduction for efficient vision transformers.}
A large body of work accelerates ViTs by discarding redundant tokens. Pruning methods drop
low-importance tokens: EViT~\citep{liang2022evit} reorganizes tokens by their attention to the class
token and fuses the inattentive ones. Merging methods instead combine similar tokens:
ToMe~\citep{bolya2023tome} merges a fixed number of token pairs per layer via bipartite soft
matching, and PiToMe~\citep{tran2024pitome} merges progressively using proportional attention.
Adaptive schemes vary the number of tokens per input: ATC~\citep{haurum2024atc} clusters tokens into
groups, while ATS~\citep{fayyaz2022ats} samples a subset to keep by significance.
Dynamic-sparsification methods such as DynamicViT~\citep{rao2021dynamicvit} and A-ViT~\citep{yin2022avit}
instead learn an input-dependent number of tokens to keep, but require training the reduction policy
and so do not apply when the model cannot be retrained on the shift. We therefore scope this work to
\emph{training-free} token reduction, which operates purely at inference and is the relevant regime under
distribution shift, and use the five methods above throughout.

Depth allocation has been considered mostly for clean-data efficiency or adaptive reduction policies, not as a controlled robustness lever under distribution shift. ToMe random-searches clean ImageNet merging schedules and reports throughput-dependent preferences, including late-heavy schedules at low throughput, flat schedules at intermediate throughput, and decreasing schedules at high throughput. Other recent methods introduce depth dependence in different ways: ATM~\citep{lee2025atm} varies similarity thresholds with depth, DyMU~\citep{wang2025dymu} studies dynamic merging for VLMs, DSM~\citep{heo2023dsm} delays global merging by skipping early blocks, and DiffRate~\citep{chen2023diffrate}/LTMP~\citep{bonnaerens2023ltmp} learn layer-wise rates or thresholds with training-time clean objectives. Our question is different: for fixed training-free reducers, with no target labels, adaptation, or retraining, how should a static schedule allocate removal under distribution shift at matched inference cost? We answer this with a one-parameter late-concentrated family calibrated to use no more GFLOPs than the flat baseline; under that conservative cost constraint, late removes more tokens in total yet consistently improves OOD accuracy.

\paragraph{Robustness to distribution shift.}
ImageNet-C~\citep{hendrycks2019imagenetc} measures robustness to common corruptions at graded severities and is the standard distribution-shift
benchmark. ViTs~\citep{dosovitskiy2021vit,touvron2021deit} exhibit nontrivial robustness to such
corruptions, and a line of work studies how their architecture and training affect this
robustness~\citep{bhojanapalli2021understanding,paul2022vit}. The robustness of token-reduced ViTs,
however, is largely unexamined: efficiency methods are tuned for clean accuracy, and whether, and
how, token reduction should differ under shift is open. We show that the schedule specifically
governs shift robustness, with gains that grow with corruption severity and stay small on clean
data.

\paragraph{Test-time adaptation, and adaptation for token reduction.}
Test-time adaptation (TTA) improves robustness by updating model parameters on unlabeled test
inputs~\citep{wang2022tent}. Most related to us, recent work combines token reduction with TTA: NAViA~\citep{xiong2025navia} adds learnable class-token biases to
recover the mutual information that token aggregation discards, and TCA~\citep{wang2025tca} performs
VLM-specific token condensation with a CLIP/SigLIP test-time cache of domain-anchor tokens and logit
self-correction. Both report that token reduction costs accuracy under existing TTA and answer it by
adding to the model, via backpropagation or a maintained cache, while leaving the reduction schedule
itself unchanged. We instead reshape the schedule alone, with no gradients, cache, or added parameters, and gain
over the flat schedule both use; being inference-only, it is orthogonal to TTA and composes with it (Section~\ref{sec:tta}).
A direct comparison is not straightforward. TCA is coupled to CLIP/SigLIP zero-shot inference, so
porting it to the supervised DeiT/ViT classifiers we study would define a new TCA-inspired method
rather than a drop-in baseline, and evaluating it natively would need a separate zero-shot protocol;
NAViA, closer to our setting, has no official implementation, so any comparison would rest on our reimplementation, not the authors' method. We therefore treat both as complementary, not competing: because they reduce
tokens on a flat schedule by default, our reshaping could in principle be applied within either,
changing where tokens are merged across depth, not how adaptation operates.
\vspace{30pt}

\section{Method: Late-Concentrated Schedule}
\label{sec:method}

\begin{figure}[t]
\centering
\vspace{-15pt}
\includegraphics[width=0.97\linewidth]{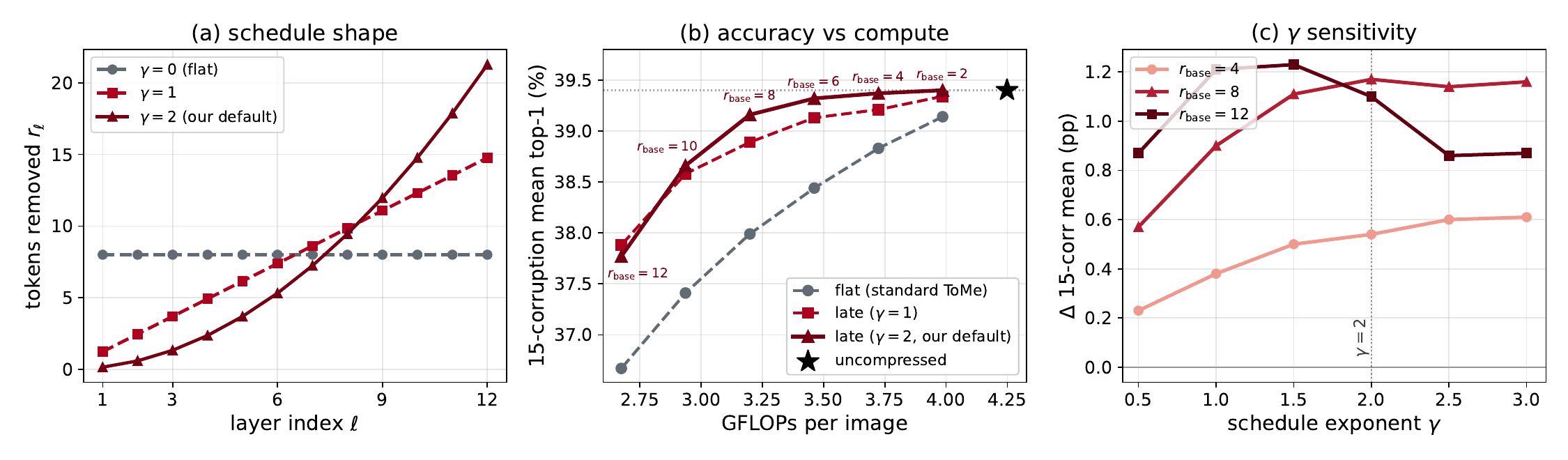}
\vspace{-15pt}
\caption{The late-concentrated schedule and its effect (DeiT-S, ImageNet-C severity~5).
  \textbf{(a)} schedule shape: tokens removed per layer $r_\ell\!\propto\!(\ell/L)^{\gamma}$ vs.\ layer
  index (at $r_{\mathrm{base}}{=}8$), for $\gamma{=}0$ (flat), $1$, and our default $2$; larger
  $\gamma$ defers removal to later layers.
  \textbf{(b)} 15-corruption-mean top-1 vs.\ compute (GFLOPs), each curve a schedule swept over
  compression (points labeled by $r_{\mathrm{base}}$); $\star$ marks the uncompressed model.
  \textbf{(c)} gain over flat vs.\ $\gamma$, one curve per compression level
  $r_{\mathrm{base}}\!\in\!\{4,8,12\}$.}
\label{fig:calib}
\vspace{-10pt}
\end{figure}

\paragraph{Power-law schedule.}
A reduction schedule sets a removal count at each of the network's $L$ layers. We seek a one-parameter family that spans schedules from flat to strongly late-concentrated and whose compute cost is analytic in that parameter, so a schedule can be matched to a baseline's compute without search. We therefore propose a power-law schedule over normalized depth,
\begin{equation}
  r(\ell) = R \cdot \frac{(\ell/L)^{\gamma}}{\sum_{k=1}^{L}\,(k/L)^{\gamma}}, \qquad \ell = 1,\ldots,L,
  \label{eq:schedule}
\end{equation}
where $R$ is the total number of tokens removed across the network and the exponent $\gamma$ is the only parameter our method sets. We use a power law because it recovers the flat schedule as the special case $\gamma{=}0$ (a uniform $R/L$ removal at every layer, the default across training-free token reducers~\citep{bolya2023tome,liang2022evit,fayyaz2022ats,haurum2024atc}), stays monotone non-decreasing in depth for every $\gamma{>}0$ (no layer removes more than a later one), and redistributes the same budget $R$ continuously toward the later layers as $\gamma$ grows (Figure~\ref{fig:calib}a). It parameterizes late-concentration rather than asserting an optimal schedule; likewise, $\gamma$ is set to a robust default ($\gamma{=}2$), not tuned per task (Figure~\ref{fig:calib}c). The finding is the late-concentrated \emph{direction} and its intensity (Figure~\ref{fig:placement}), and we expect other monotone late-weighted profiles to behave similarly; a check at the main operating point supports this, with an exponential profile and a non-smooth two-level step (each at iso-GFLOPs) recovering gains comparable to the power law (Appendix~\ref{app:shapeform}). We round the schedule to integer per-layer counts and apply it identically across all five reducers, which then differ only in which tokens they select (rounding rule and per-layer cap in Appendix~\ref{app:impl}).

\paragraph{Calibration to the flat baseline's compute (iso-GFLOPs).}
We label each operating point by $r_{\mathrm{base}}$, the per-layer rate of the flat baseline it is matched against and the deployment choice of how much compute to save ($R{=}r_{\mathrm{base}}L$ for the flat baseline only); the late schedule's own budget is set by the calibration below and is larger, with per-layer mean $\bar{r}{=}R/L > r_{\mathrm{base}}$ (Appendix~\ref{app:calib}). Later-layer removals save less compute (fewer tokens cascade through fewer remaining layers). At the same $r_{\mathrm{base}}$, late-concentrated uses fewer GFLOPs (giga floating-point operations), permitting more (later) token removal to restore GFLOPs parity. All comparisons in this paper are at fair GFLOPs, not equal $r_{\mathrm{base}}$. These iso-GFLOPs comparisons are also matched in wall-clock: at the main operating point the late ($\gamma{=}2$) and flat schedules differ by under $0.5\%$ in measured throughput, so the comparison is fair in real latency and not only in analytic FLOPs (Appendix~\ref{app:latency}).

We calibrate each $(r_{\mathrm{base}},\gamma)$ by choosing the integer per-layer removal budget whose GFLOPs match the corresponding flat baseline; because GFLOPs is analytic in the schedule, this calibration needs no extra evaluation. When an exact match is not attainable at integer budgets, we always round the budget so that the late-concentrated schedule uses no more GFLOPs than its flat baseline (the full list of calibrated schedules is given in Appendix~\ref{app:calib}, Table~\ref{tab:isocalib}). The late-versus-flat comparison is therefore strictly conservative, and the gains cannot be attributed to a larger compute budget. Figure~\ref{fig:calib}b plots the resulting accuracy--compute frontier on the 15-corruption mean: the calibrated late-concentrated family Pareto-dominates the flat one across the entire compute range.

\section{Why Late Concentration Improves Robustness}
\label{sec:mechanism}
\label{sec:model}
\label{sec:mechtoken}
\vspace{-5pt}
We provide a first-order account for why late concentration is preferred in our measurements under distribution shift, and why the empirical optimum is interior rather than maximally late. (i)~A reduction's output distortion is front-loaded in depth, because earlier perturbations pass through more downstream layers. (ii)~The compute saved by a reduction also falls with depth, but more slowly, so moving budget later lowers distortion per unit of compute saved. (iii)~Per-layer capacity keeps the empirical optimum interior, because each layer can merge only a bounded number of similar tokens.

\begin{figure}[t]
\centering
\vspace{-15pt}
\includegraphics[width=\linewidth]{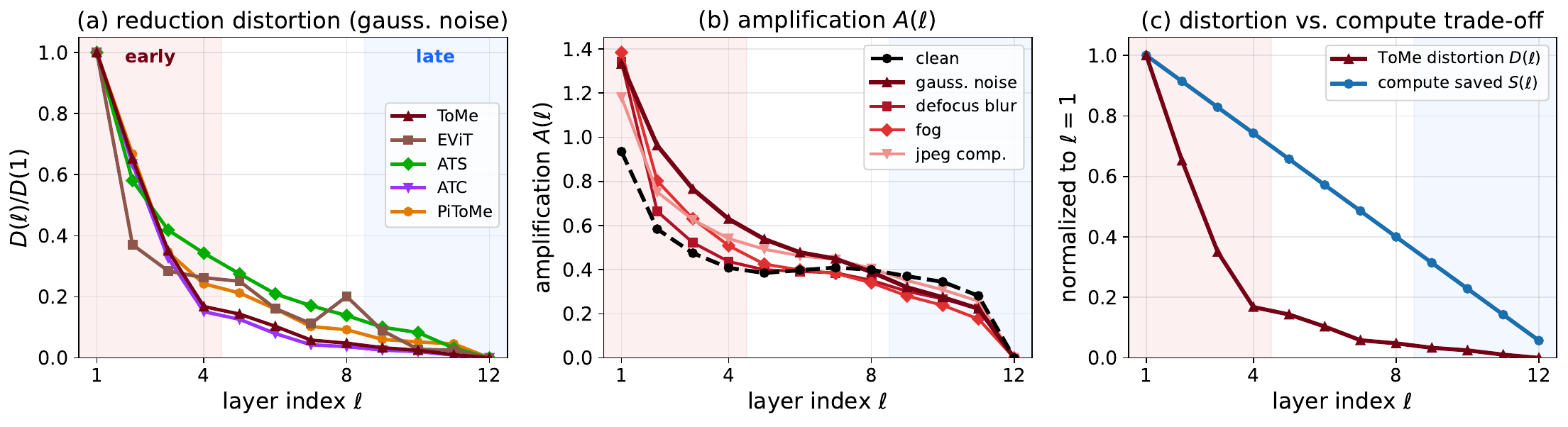}
\vspace{-25pt}
\caption{Per-layer reduction distortion and amplification (DeiT-S, ImageNet-C severity~5; means over
  $N{=}1024$ images, standard error $<\!2\%$). \textbf{(a)} The directly-measured distortion $D(\ell)$ of reducing
  at layer~$\ell$ only (relative output-logit change, gaussian noise), normalized to layer~1, for all five
  reducers. \textbf{(b)} The amplification $A(\ell)$ (clean and one corruption per ImageNet-C
  group). \textbf{(c)} $D(\ell)$ against the compute a removal at layer~$\ell$ saves, $S(\ell)$, both
  normalized; $D$ falls far faster than $S$.}
\label{fig:mechtoken}
\vspace{-5pt}
\end{figure}

\paragraph{The reduction distortion is front-loaded.}
A reduction step (merging, pruning, clustering, or sampling) removes or distorts token information at the layer where it acts, and the perturbation propagates forward to the classification readout; we call the resulting output error the distortion of reducing at layer $\ell$. We measure it directly and reducer-agnostically. For each training-free reducer we remove a fixed number of tokens at layer $\ell$ and nowhere else, so per-layer differences reflect depth alone, and read the relative logit change $D(\ell)=\mathbb{E}_x\,\lVert f^{(\ell)}(x)-f(x)\rVert/\lVert f(x)\rVert$ ($f$ unreduced, $f^{(\ell)}$ reduced at $\ell$ only), a removal-based sensitivity measure in the tradition of ablation-importance criteria~\citep{lecun1990obd,michel2019sixteen,lee2025atm}. For all five reducers (ToMe, EViT, ATS, ATC, PiToMe) $D(\ell)$ is steeply front-loaded (Figure~\ref{fig:mechtoken}a), and it stays front-loaded under every corruption, with the early/late distortion ratio exceeding $6\times$ in all $20$ reducer$\times$corruption cases (Spearman $\le\!-0.90$; per-cell values in Appendix~\ref{app:mechgen}). The distortion is near zero at the last layer ($D(\ell{=}12)\approx0$). Under the class-token readout a final-layer patch merge has no later layer through which to reach the class token, so it costs almost nothing, and the gain does not depend on this (Appendix~\ref{app:finalblock}). The account rests on single-layer probes, whereas a deployed schedule composes many reductions whose effects interact, each merge changing what later ones see. It still holds for the composed schedules: at equal compute the late schedule's mean per-case KL to the unreduced model is $0.029$ against $0.085$ for flat, and it changes the unreduced model's top-1 on $10.8\%$ of cases against $19.1\%$, lower on all $15$ corruptions and both backbones (Appendix~\ref{app:channels}, Figure~\ref{fig:predagree}).

\paragraph{Perturbation amplification induces the front-loading.}
The distortion is front-loaded because the network amplifies an early perturbation more: it passes through more downstream layers on its way to the classifier. We probe this gain directly. At layer $\ell$ we perturb each output patch token by $\delta_t=\alpha\,\lVert x_t\rVert\,u_t$, with $u_t$ an isotropic unit vector and $\alpha{=}0.1$ (the class token is left untouched), let it propagate to the end, and read the pre-logit class-token representation $z$. The amplification $A(\ell)=\mathbb{E}\big[\lVert z_{\mathrm{pert}}-z_{\mathrm{clean}}\rVert/\lVert z_{\mathrm{clean}}\rVert\big]/\alpha$, averaged over $5$ draws shared across layers, is a Jacobian-norm sensitivity probe~\citep{novak2018sensitivity} anchored at a hidden layer. Linearizing the downstream layers (Lemma~\ref{lem:amp}, Appendix~\ref{app:probe}) sends a perturbation at layer $\ell$ to the readout through the composed Jacobian $\prod_{k>\ell}J_k$; the isotropic probe measures the average-case gain of that product, and its operator norm bounds that gain from above. Each deeper layer drops one factor, so the gain is non-increasing in $\ell$ whenever those layers are on average non-contractive, the familiar depth-compounding of perturbations in deep networks~\citep{poole2016exponential}. Amplification is a property of the network, not of any reducer's choices, consistent with $D(\ell)$ being front-loaded for all five. Measured on clean inputs and one corruption from each ImageNet-C group (severity~5), $A(\ell)$ is front-loaded in every condition (Figure~\ref{fig:mechtoken}b; Spearman $\le\!-0.92$), with clean the flattest profile, so corruption steepens the amplification.

\paragraph{Capacity keeps the optimum interior, not maximally late.}
What a schedule spends is compute, not token count, so the governing quantity is not the raw distortion but the distortion per unit of compute saved, $\rho(\ell)=D(\ell)/S(\ell)$, where $S(\ell)$ is the marginal compute a removal at layer $\ell$ saves. ToMe, our default reducer, merges between attention and the MLP, so a removal saves that layer's own MLP and all downstream layers: $S(\ell)=c+(L-\ell)$ in units of a full layer, with $c\approx0.61$ the own-MLP fraction and $S(L)=c>0$. A THOP measurement confirms this near-linear profile, the last layer saving only $\approx6\%$ of what the first does. Figure~\ref{fig:mechtoken}c contrasts them: $D$ falls far faster than $S$, so late reductions are cheap in distortion yet save little compute, and $\rho$ falls with depth. That is why late concentration pays at equal compute. The ratio alone does not bound how late: since $A(L)=0$ the last layer reduces at near-zero distortion, $\rho(L)\to0$, and unconstrained $\rho$-minimization would pile the budget onto the latest layers. We attribute the \emph{interior} optimum to per-layer capacity instead. Each layer can merge only a bounded number of similar tokens, so the budget must spill beyond the few cheap late layers, and as a layer nears that limit the marginal distortion of each further merge rises, favoring a smooth spread over a few saturated layers. The account is first-order and does not pinpoint an optimal $\gamma$, but its predictions hold: the interior inverted-U in $\gamma$ (Table~\ref{tab:main}, Figure~\ref{fig:calib}c), its migration toward weaker concentration at larger removal budgets (Appendix~\ref{app:sweetspot}), and the late-concentrated, last-layer-heavy shape a schedule searched for clean accuracy takes (Appendix~\ref{app:learned}).

\paragraph{Why the advantage grows under shift.}
The gain grows monotonically with corruption severity (Section~\ref{sec:severity}); this is an empirical property, which the probes explain in direction, not magnitude. Token reduction sacrifices more accuracy under a stronger shift (Figure~\ref{fig:compdrop}), so a better allocation of the same budget has more to recover. The representational reason is that the network amplifies a reduction's perturbation more off the clean manifold. Self-attention is non-Lipschitz~\citep{kim2021lipschitz} and can develop larger Jacobians off-distribution (Appendix~\ref{app:ood-amp}); the amplification $A(\ell)$ rises under every corruption group (early $A(\ell)$ from $0.60$ on clean to $0.74$--$0.92$) and, compounding through more downstream layers, does so most in the early layers (Lemma~\ref{lem:amp}). The early reductions the flat schedule spends its budget on are therefore the most amplified under shift, so the late schedule recovers more. We do not read the probes as predictors of the per-corruption gain, whose magnitude also reflects how close the corrupted inputs sit to the decision boundary.

\paragraph{The gain is not a readout or insertion-point artifact.}
The late advantage survives dropping the final layers from both schedules, and it reproduces under global-average-pooling readout, which removes the free last layer but leaves the front-loaded distortion intact: on MAE ViT-B and ViT-L every one of the $15$ corruptions improves ($+1.28$ and $+1.22$). The strictly free final-layer saving is in any case only $0.2$--$0.7\%$ of total GFLOPs (Appendix~\ref{app:finalblock}).

\paragraph{Allocation outweighs the matching criterion.}
\begin{wrapfigure}{r}{0.35\linewidth}
  \centering
  \vspace{-15pt}
  \includegraphics[width=\linewidth]{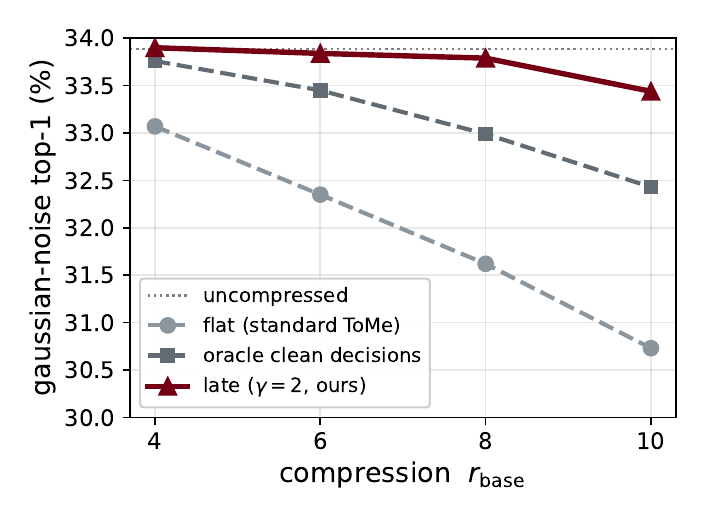}
  \vspace{-25pt}
  \caption{Oracle clean merge decisions recover less than the late
    schedule at every $r_{\mathrm{base}}$, the gap widening with compression.}
  \label{fig:cleanpath}
  \vspace{-10pt}
\end{wrapfigure}
Our account concerns where tokens are removed; a distinct possible source of the OOD penalty is which tokens are merged, since corrupted similarities could worsen merge decisions (the matching criterion). In an oracle control, for each corrupted image we force the per-layer merge decisions its clean counterpart would make (captured offline, unavailable at test time), holding the schedule and the compute fixed. Even the flat schedule with these oracle decisions recovers less OOD accuracy than the late schedule at every operating point (Figure~\ref{fig:cleanpath}, gaussian noise; full ImageNet-C in Appendix~\ref{app:cleanpath}). Allocation therefore outweighs the gain from this oracle's improved matching; a genuinely corruption-robust criterion, which this oracle does not upper-bound, remains a complementary axis (Section~\ref{sec:limitations}).

\section{Experiments}
\label{sec:experiments}

\paragraph{Setup.}
We evaluate token-reduction schedules across image-classification backbones (DeiT-S/B~\citep{touvron2021deit}, ViT-S/B/L~\citep{dosovitskiy2021vit}, CLIP ViT-B/L~\citep{radford2021clip}, and MAE ViT-B/L~\citep{he2022mae}), a video model (VideoMAE ViT-B~\citep{tong2022videomae}), and a vision-language model (LLaVA-1.5-7B~\citep{liu2024llava15}). The schedule is applied to five training-free reducers; ToMe~\citep{bolya2023tome} is the default because it is widely used and merges tokens rather than discarding them, while EViT, ATS, ATC, and PiToMe verify the effect is not specific to merging. Our primary benchmark is ImageNet-C at severity~5, and we test generalization on ImageNet-\cbar/3DCC, ImageNet-R, ImageNet-Sketch, NaBirds-C, Mini-Kinetics-C, and GQA-C. To probe composability with test-time adaptation we run Tent, EATA, SAR, FOA, SPA, and NEO on top of the schedule. Comparisons are inference-only (and deterministic) unless a TTA method is stated; we report the severity-5 15-corruption mean top-1 unless noted. Our main operating point is $r_{\mathrm{base}}{=}8$ with $\gamma{=}2$ (a $26\%$ compute reduction); we sweep compression from $r_{\mathrm{base}}{=}2$ to $12$ (Table~\ref{tab:main}), with $\gamma$ fixed at 2 rather than tuned per rate. Full implementation details are in Appendix~\ref{app:impl}.

\subsection{Late-Concentrated vs.\ Flat at iso-GFLOPs}

\begin{table}[t]
\centering
\vspace{-15pt}
\caption{\textbf{$\gamma$ sensitivity across operating points at iso-GFLOPs} (DeiT-S, ImageNet-C
  severity~5, 15-corruption mean top-1, \%). The GFLOPs column is the default $\gamma{=}2$ schedule's compute
  and its reduction against the uncompressed model ($39.40$ at $4.25$G; per-$\gamma$ schedules in
  Table~\ref{tab:isocalib}). flat is the const-$r_{\mathrm{base}}$ baseline in absolute top-1; the
  $\gamma$ columns are $\Delta$ = late $-$ flat. \colorbox{oursgreen}{Shaded} = the
  default $\gamma{=}2$; \textbf{bold} = main result ($r_{\mathrm{base}}{=}8$); \underline{underline}
  = per-row optimum $\gamma^{\star}$.}
\label{tab:main}
\footnotesize\setlength{\tabcolsep}{3pt}
\begin{tabularx}{\linewidth}{l l R R R R G R R}
\toprule
 & & & \multicolumn{6}{c}{late-concentrated $\gamma$ \;($\Delta$ vs.\ flat)} \\
\cmidrule(lr){4-9}
$r_{\mathrm{base}}$ & GFLOPs (cut) & flat & $0.5$ & $1.0$ & $1.5$ & $2.0$ & $2.5$ & $3.0$ \\
\midrule
$2$  & 3.97G ($-7\%$)  & 39.14 & $+0.09$ & $+0.20$ & $+0.22$ & $+0.26$ & $+0.27$ & $\underline{+0.28}$ \\
$4$  & 3.69G ($-13\%$) & 38.83 & $+0.23$ & $+0.38$ & $+0.50$ & $+0.54$ & $\underline{+0.61}$ & $+0.60$ \\
$6$  & 3.34G ($-21\%$) & 38.44 & $+0.34$ & $+0.69$ & $+0.81$ & $\underline{+0.88}$ & $+0.83$ & $+0.69$ \\
$8$  & 3.15G ($-26\%$) & 37.99 & $+0.57$ & $+0.90$ & $+1.11$ & $\underline{\mathbf{+1.17}}$ & $+1.14$ & $+1.16$ \\
$10$ & 2.90G ($-32\%$) & 37.41 & $+0.67$ & $+1.17$ & $\underline{+1.28}$ & $+1.25$ & $+1.15$ & $+0.92$ \\
$12$ & 2.67G ($-37\%$) & 36.67 & $+0.87$ & $+1.21$ & $\underline{+1.23}$ & $+1.10$ & $+0.86$ & $+0.87$ \\
\bottomrule
\end{tabularx}
\end{table}

\begin{figure}[b]
\centering
\vspace{-10pt}
\begin{minipage}[t]{0.45\linewidth}
  \centering
  \includegraphics[width=\linewidth]{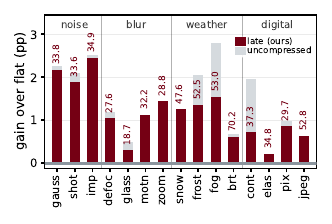}
  \vspace{-25pt}
  \caption{Per-corruption gain over flat at iso-GFLOPs (DeiT-S, $r_{\mathrm{base}}{=}8$, $\gamma{=}2$, ImageNet-C sev=5); uncompressed in gray. All 15 corruptions improve; labels are late's top-1.}
  \label{fig:percorruption}
\end{minipage}\hfill
\begin{minipage}[t]{0.52\linewidth}
  \centering
  \includegraphics[width=\linewidth]{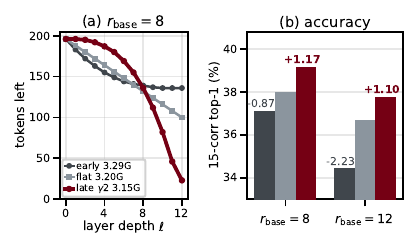}
  \vspace{-25pt}
  \caption{Schedule placement (DeiT-S, ImageNet-C sev=5). \textbf{(a)} tokens left per layer at
    $r_{\mathrm{base}}{=}8$, with each schedule's GFLOPs; the $r_{\mathrm{base}}{=}12$ profiles are in
    Appendix~\ref{app:placer12}. \textbf{(b)} top-1 at both rates, $\Delta$ vs.\ flat.}
  \label{fig:placement}
\end{minipage}
\end{figure}
The late-concentrated schedule at $r_{\mathrm{base}}{=}8$, $\gamma{=}2$ improves the corrupted mean by $+1.17$pp over flat (Table~\ref{tab:main}). That closes $83\%$ of the accuracy compression costs, leaving a $-0.24$pp gap to the uncompressed model against $-1.41$pp for flat, at a $\sim$26\% compute reduction. The gain is recovered predictions, not a compensating mix: flat destroys $8.5\%$ of what the unreduced model gets right and ours $3.9\%$ (Appendix~\ref{app:channels}). The gain grows with compression, from $+0.26$pp at the lightest $r_{\mathrm{base}}{=}2$ to $+1.25$pp at $r_{\mathrm{base}}{=}10$, and holds at $+1.10$pp at the aggressive $r_{\mathrm{base}}{=}12$ (Figure~\ref{fig:calib}b). The $\gamma$ sweep is an inverted-U, positive for every tested $\gamma\in[0.5,3.0]$ ($\geq +0.57$pp), with the optimum migrating from $\gamma^{\star}\!\approx\!3$ at $r_{\mathrm{base}}{=}2$ to $\approx\!1.5$ at $r_{\mathrm{base}}{=}12$; $\gamma{=}2$ stays within $0.13$pp of the per-rate optimum. All 15 corruptions improve (Figure~\ref{fig:percorruption}; $+0.21$pp on elastic transform to $+2.45$pp on impulse noise), as do the four outside ImageNet-C's standard set (Appendix~\ref{app:extrac}); a sign test rejects the no-effect null at $p<10^{-4}$. Depth allocation is a deployment-neutral lever: it requires no retraining or extra compute, and the gain holds for every tested $\gamma$ and every corruption rather than at a single tuned setting.

\paragraph{Early vs.\ flat vs.\ late schedule.}
To test the schedule's directionality, we evaluate the symmetric early-concentrated schedule (the same power law reflected to remove more early). The result is a monotone ordering, early $<$ flat $<$ late, at both operating points (Figure~\ref{fig:placement}). Concentrating removal early hurts ($-2.23$ to $-0.87$pp vs.\ flat), late helps ($+1.10$ to $+1.17$pp). This is the ordering predicted by the front-loaded per-step distortion (Section~\ref{sec:model}). Early reductions are the most damaging, so concentrating removal early is the worst allocation. Crucially the early deficit is attributable to the schedule, not compute or token budget. We calibrate the early-concentrated budget to flat's compute; it comes out slightly higher: $2.8\%$ more GFLOPs at $r_{\mathrm{base}}{=}8$ and $1.3\%$ more at $r_{\mathrm{base}}{=}12$. The early schedule also removes fewer tokens in total (60 vs.\ 96 at $r_{\mathrm{base}}{=}8$). More compute and less removal should both help, yet early is the worst of the three, losing to flat by $0.87$--$2.23$pp, while the late schedule uses the least compute and wins. Thus the ordering cannot be explained by the simple resource variables of retained token count or compute alone; the depth allocation of removal is the decisive factor in these comparisons. %

\subsection{The Gain Across Shift Severities}
\label{sec:severity}
\begin{wraptable}{r}{0.42\linewidth}
\centering
\vspace{-22pt}
\caption{Per-severity 15-corruption mean (\%), default $\gamma{=}2$, $r_{\mathrm{base}}{=}8$, DeiT-S;
  severity~0 is clean ImageNet val. $\Delta$ = late $-$ flat, $\Delta/\Delta_0$ its multiple of the
  clean gain.}
\label{tab:severity}
\fontsize{7.7}{9}\selectfont\setlength{\tabcolsep}{3pt}
\begin{tabularx}{\linewidth}{l R G D R}
\toprule
Sev. & flat & late & $\Delta$ & $\Delta/\Delta_0$ \\
\midrule
0 (clean) & 79.40 & 79.68 & $+0.28$ & 1.0$\times$ \\
1         & 70.52 & 71.04 & $+0.52$ & 1.9$\times$ \\
2         & 64.33 & 64.99 & $+0.66$ & 2.4$\times$ \\
3         & 59.06 & 59.89 & $+0.83$ & 3.0$\times$ \\
4         & 50.06 & 51.12 & $+1.06$ & 3.8$\times$ \\
5         & 37.99 & 39.16 & $+1.17$ & \textbf{4.2$\times$} \\
\bottomrule
\end{tabularx}
\vspace{-15pt}
\end{wraptable}
The improvement grows with the shift, rising monotonically with corruption severity (Table~\ref{tab:severity}). At late $\gamma{=}2$ it rises from $+0.28$pp on clean images to $+1.17$pp at severity~5, so the corrupted gain is $4.2\times$ the clean gain. The gain is therefore a robustness effect, not a general accuracy improvement. The $\gamma{=}1$ schedule traces the same monotone profile (Appendix~\ref{app:severityg1}). The nonzero clean gain is not a separate effect: the per-step reduction distortion $D(\ell)$ is front-loaded even on clean inputs (Section~\ref{sec:mechtoken}), so deferring removal helps by a small margin with no corruption, and shift inflates the early-layer distortion further, widening that same margin. The clean and corrupted gains are thus two readings of one allocation principle, the clean gain being the floor that shift steepens.

\subsection{Generalization Across Token-Reduction Methods}
\label{sec:baselines}

\begin{table}[!b]
\centering
\vspace{-15pt}
\caption{Late-concentrated schedule applied to five training-free token-reduction methods at the
  default $\gamma{=}2$. DeiT-S/B, ImageNet-C severity~5,
  15-corruption mean top-1 (\%), iso-GFLOPs; each
  header's \% is the $\gamma{=}2$ compute reduction vs.\ the uncompressed model. For
  each operating point we report flat, \textbf{late} (bold), and $\Delta$ = late $-$ flat (pp); The $\gamma{=}1$ and larger $r_{\mathrm{base}}$ results are in Appendix~\ref{app:baselinesg1}.}
\label{tab:baselines}
\fontsize{7.7}{9}\selectfont\setlength{\tabcolsep}{3pt}
\begin{tabularx}{\linewidth}{l R G D R G D R G D}
\toprule
 & \multicolumn{3}{c}{$r_{\mathrm{base}}{=}4$ ($-13\%$)} & \multicolumn{3}{c}{$r_{\mathrm{base}}{=}6$ ($-21\%$)} & \multicolumn{3}{c}{$r_{\mathrm{base}}{=}8$ ($-26\%$)} \\
\cmidrule(lr){2-4}\cmidrule(lr){5-7}\cmidrule(lr){8-10}
Method & flat & late & $\Delta$ & flat & late & $\Delta$ & flat & late & $\Delta$ \\
\midrule
\rowcolor{groupgrey} \multicolumn{10}{l}{\emph{DeiT-S}} \\
ToMe   & 38.83 & 39.37 & $+0.54$ & 38.44 & 39.32 & $+0.88$ & 37.99 & 39.16 & $+1.17$ \\
EViT   & 37.50 & 38.97 & $+1.47$ & 36.69 & 38.91 & $+2.22$ & 35.98 & 38.50 & $+2.52$ \\
ATS    & 34.95 & 35.43 & $+0.48$ & 34.50 & 35.21 & $+0.71$ & 33.92 & 34.88 & $+0.96$ \\
ATC    & 38.28 & 38.73 & $+0.45$ & 37.96 & 38.74 & $+0.78$ & 37.55 & 38.65 & $+1.10$ \\
PiToMe & 38.45 & 38.99 & $+0.54$ & 38.14 & 38.67 & $+0.53$ & 37.67 & 37.90 & $+0.23$ \\
\midrule
\rowcolor{groupgrey} \multicolumn{10}{l}{\emph{DeiT-B}} \\
ToMe   & 45.98 & 46.50 & $+0.52$ & 45.27 & 46.17 & $+0.90$ & 44.35 & 45.89 & $+1.54$ \\
EViT   & 44.52 & 45.80 & $+1.28$ & 43.57 & 45.38 & $+1.81$ & 42.45 & 44.62 & $+2.17$ \\
ATS    & 40.60 & 41.34 & $+0.74$ & 39.79 & 40.75 & $+0.96$ & 38.81 & 40.11 & $+1.30$ \\
ATC    & 45.41 & 45.96 & $+0.55$ & 44.79 & 45.73 & $+0.94$ & 43.97 & 45.47 & $+1.50$ \\
PiToMe & 45.25 & 45.78 & $+0.53$ & 44.56 & 44.84 & $+0.28$ & 43.69 & 43.97 & $+0.28$ \\
\bottomrule
\end{tabularx}
\vspace{5pt}
\end{table}

At iso-GFLOPs the late-concentrated schedule (default $\gamma{=}2$) improves all five training-free methods, on both DeiT-S and DeiT-B, at every operating point (Table~\ref{tab:baselines}): all 30 cells are positive ($+0.23$ to $+2.52$pp), so the same schedule used for ToMe transfers unchanged to the other methods. This is what Section~\ref{sec:mechanism} predicts, since the distortion $D(\ell)$ is front-loaded for every one of the five reducers (Figure~\ref{fig:mechtoken}a). The benefit spans method families, from attention-guided pruning (EViT~\citep{liang2022evit}) and merging (ToMe~\citep{bolya2023tome}, bipartite matching; PiToMe~\citep{tran2024pitome}, spectrum-preserving) to agglomerative clustering (ATC~\citep{haurum2024atc}) and adaptive sampling (ATS~\citep{fayyaz2022ats}), with EViT gaining most. The same five methods at $\gamma{=}1$, including a more aggressive operating point, are in Appendix~\ref{app:baselinesg1}, where the late schedule again beats flat in all cells, and Figure~\ref{fig:calibreducers} plots each method's accuracy against compute.

\subsection{Generalization Across Backbones and Pretraining}
\label{sec:arch}

\begin{table}[t]
\centering
\vspace{-5pt}
\begin{minipage}[t]{0.48\linewidth}
  \centering
  \caption{Generalization across backbones at the default $r_{\mathrm{base}}{=}8$ rate, a $24$--$25\%$ compute
    cut (ImageNet-C sev~5, 15-corruption mean top-1, \%). $\Delta$ = late $-$ flat, \textbf{bold} the default.}
    \vspace{2pt}
  \label{tab:arch}
  \resizebox{\linewidth}{!}{%
\begin{tabular}{l r r >{\columncolor{oursgreen}\bfseries\boldmath}r}
\toprule
Backbone & flat & late $\gamma{=}1$ ($\Delta$) & late $\gamma{=}2$ ($\Delta$) \\
\midrule
\rowcolor{groupgrey} \multicolumn{4}{l}{\emph{Supervised, distilled}}\\
DeiT-S & 37.99 & 38.89 ($+0.90$) & 39.16 ($+1.17$) \\
DeiT-B & 44.35 & 45.35 ($+1.00$) & 45.89 ($+1.54$) \\
\rowcolor{groupgrey} \multicolumn{4}{l}{\emph{Supervised, AugReg}}\\
ViT-S & 38.05 & 39.02 ($+0.97$) & 39.46 ($+1.41$) \\
ViT-B & 53.39 & 54.38 ($+0.99$) & 54.96 ($+1.57$) \\
ViT-L & 58.82 & 60.32 ($+1.50$) & 60.62 ($+1.80$) \\
\rowcolor{groupgrey} \multicolumn{4}{l}{\emph{Contrastive image-text}}\\
CLIP ViT-B & 43.68 & 44.72 ($+1.04$) & 45.07 ($+1.39$) \\
CLIP ViT-L & 57.29 & 58.10 ($+0.81$) & 58.41 ($+1.12$) \\
\rowcolor{groupgrey} \multicolumn{4}{l}{\emph{Self-supervised, global pooling}}\\
MAE ViT-B & 42.81 & 43.77 ($+0.96$) & 44.09 ($+1.28$) \\
MAE ViT-L & 52.97 & 53.96 ($+0.99$) & 54.19 ($+1.22$) \\
\bottomrule
\end{tabular}%
  }
\end{minipage}\hfill
\begin{minipage}[t]{0.5\linewidth}
  \centering
  \caption{Generalization across shift types and modalities (flat vs.\ late $\gamma{=}2$, iso-GFLOPs,
    accuracy \%). $\Delta$ = late $-$ flat (pp); the same schedule is applied to each system's ViT backbone.}
  \label{tab:generalization}
  \resizebox{\linewidth}{!}{%
\begin{tabular}{l r >{\columncolor{oursgreen}}r >{\columncolor{deltablue}\boldmath}r}
\toprule
Setting & flat & late & $\Delta$ \\
\midrule
\rowcolor{groupgrey} \multicolumn{4}{l}{\emph{Natural shift}}\\
ImageNet-A & 25.31 & 26.55 & $+1.24$ \\
ImageNet-R & 45.90 & 46.57 & $+0.67$ \\
ImageNet-V2 & 69.38 & 69.97 & $+0.59$ \\
ImageNet-Sketch & 32.89 & 33.33 & $+0.44$ \\
DomainNet-126 & 65.14 & 65.84 & $+0.70$ \\
\rowcolor{groupgrey} \multicolumn{4}{l}{\emph{Synthetic corruption}}\\
ImageNet-\cbar{} & 47.28 & 48.74 & $+1.45$ \\
ImageNet-3DCC & 41.47 & 42.38 & $+0.91$ \\
NaBirds-C & 31.06 & 31.83 & $+0.77$ \\
\rowcolor{groupgrey} \multicolumn{4}{l}{\emph{Other modalities}}\\
Mini-Kinetics-C (VideoMAE) & 66.17 & 66.79 & $+0.62$ \\
GQA-C (LLaVA) & 54.96 & 55.20 & $+0.24$ \\
\bottomrule
\end{tabular}%
}
\end{minipage}
\vspace{-5pt}
\end{table}

Table~\ref{tab:arch} applies the schedule at iso-GFLOPs to nine backbones spanning four pretraining recipes, three scales, and both readouts. The default $\gamma{=}2$ beats flat on all nine ($+1.12$ to $+1.80$pp) and $\gamma{=}1$ on all nine. The effect does not fade as the backbone grows larger or more accurate, ViT-L being both the most accurate ($58.82$ flat) and the one that gains most ($+1.80$). Nor does it depend on how the backbone was trained. The three ViT-B rows span $10.6$pp of flat accuracy under three pretraining recipes yet their gains stay within $0.29$pp, and the MAE rows keep theirs with neither supervised pretraining nor a class token. CLIP ViT-L carries $256$ tokens rather than $196$, and the schedule, defined over normalized depth, transfers to it unchanged. Appendix~\ref{app:crossbackbone} adds the per-corruption results and the aggressive $r_{\mathrm{base}}{=}12$ rate.

\subsection{Generalization Across Shift Types and Modalities}
\label{sec:shifttypes}

The schedule generalizes well beyond synthetic ImageNet-C. We apply the same iso-GFLOPs flat-versus-late comparison to eight further shift suites and two other modalities (Table~\ref{tab:generalization}): five natural shifts (ImageNet-A~\citep{hendrycks2021nae}, ImageNet-R, ImageNet-V2~\citep{recht2019imagenetv2}, ImageNet-Sketch, and DomainNet-126~\citep{peng2019domainnet} averaged over its three target domains), two harder synthetic suites (ImageNet-\cbar{}~\citep{mintun2021cbar}, ImageNet-3DCC~\citep{kar20223dcc}) with a fine-grained task (NaBirds-C), and two non-image modalities (video on Mini-Kinetics-C~\citep{yi2021benchmarking}, vision-language QA on GQA-C~\citep{hudson2019gqa,usama2025vlmcorruptions}). The image settings share a common ViT-S backbone, the non-image modalities their native models (VideoMAE, LLaVA). The gain is positive in all ten settings. Other backbones and per-corruption results are in Appendices~\ref{app:nabirds},~\ref{app:rsketch} and~\ref{app:cbar3dcc}.

The gains are largest on ImageNet-\cbar{} ($+1.45$) and ImageNet-A ($+1.24$) and smallest on the style shifts (ImageNet-R $+0.67$, Sketch $+0.44$); they do not track how much accuracy the shift costs, since near-clean ImageNet-V2 gains more than Sketch. ImageNet-A is where the schedule helps most across backbones, $+2.01$pp on average against the $+1.42$pp those same five gain on ImageNet-C, and on three of them the reduced model passes the uncompressed one. Where we dump per-image predictions, on ImageNet-A, ImageNet-V2 and each DomainNet domain, every gain in the table is significant under an exact McNemar test~\citep{mcnemar1947}, with the intervals in Appendix~\ref{app:naturalshift}, and NaBirds-C and DomainNet-126 show the effect surviving fine-tuning to a new label space. The benefit also crosses modality. On video we run VideoMAE~\citep{tong2022videomae} under the six spatial Mini-Kinetics-C corruptions with ToMe at iso-GFLOPs ($+0.62$ at severity~5 and $+0.25$ at severity~3, so the video gain grows with severity as well, and it again concentrates on the structure-degrading corruptions; per-corruption in Appendix~\ref{app:video}). On a vision-language model, the furthest transfer we test, the schedule acts only on the CLIP vision encoder that feeds LLaVA-1.5-7B~\citep{liu2024llava15}, and corrupted GQA accuracy still improves at iso-GFLOPs ($+0.24$pp).

\subsection{Composability with Test-Time Adaptation}
\label{sec:tta}

The late-concentrated schedule changes no parameters, so it can be layered under test-time adaptation (TTA), which addresses the same shift by updating the model on unlabeled test inputs~\citep{wang2022tent}. We ask whether its robustness gain persists when a TTA method runs on top, at iso-GFLOPs. We run six methods: three minimize entropy over normalization parameters (Tent~\citep{wang2022tent}, EATA~\citep{niu2022eata}, SAR~\citep{niu2023sar}); the forward-only FOA~\citep{niu2024foa} searches input prompts with an evolution strategy; the self-bootstrapping SPA~\citep{niu2025spa} trains a predictor on two geometry-preserving views; and NEO~\citep{murphy2026neo} optimizes nothing and re-centers the classifier's input. All six run on ViT-S and ViT-B, the backbones this line of adaptation work evaluates on, and on both the flat ($\gamma{=}0$) and the late schedule at the main operating point ($r_{\mathrm{base}}{=}8$, $\gamma{=}2$); we report the 15-corruption mean over 3 seeds (Table~\ref{tab:tta}). Learning rates are frozen and shared by the two schedules; these, the backbone choice and the rest of the protocol are in Appendix~\ref{app:impl}, with $r_{\mathrm{base}}{=}4$ in Appendix~\ref{app:ttavitb}.

\begin{table}[t]
\vspace{-10pt}
\centering
\caption{Late-concentrated schedule layered under TTA methods. ImageNet-C severity~5,
  15-corruption mean top-1, $r_{\mathrm{base}}{=}8$, iso-GFLOPs; each method is
  run on the flat and the late ($\gamma{=}2$) schedule, which cuts GFLOPs by $26\%$ against the same
  method on full tokens (Appendix~\ref{app:ttacost}). $\Delta$ = late $-$ flat (pp); $\pm$ = std over 3 seeds.}
\label{tab:tta}
\resizebox{\linewidth}{!}{%
\begin{tabular}{l r >{\columncolor{oursgreen}}r >{\columncolor{deltablue}\boldmath}r r >{\columncolor{oursgreen}}r >{\columncolor{deltablue}\boldmath}r}
\toprule
 & \multicolumn{3}{c}{ViT-S} & \multicolumn{3}{c}{ViT-B} \\
\cmidrule(lr){2-4}\cmidrule(lr){5-7}
Method & flat & late & $\Delta$ & flat & late & $\Delta$ \\
\midrule
Source (inference-only)        & 38.05 & 39.46 & $+1.41$ & 53.39 & 54.96 & $+1.57$ \\
\midrule
Tent~\citep{wang2022tent}      & \ms{48.47}{0.00} & \ms{52.04}{0.00} & \ms{$+3.57$}{0.00} & \ms{54.49}{0.00} & \ms{55.72}{0.00} & \ms{$+1.23$}{0.00} \\
EATA~\citep{niu2022eata}       & \ms{52.65}{0.16} & \ms{53.66}{0.26} & \ms{$+1.00$}{0.38} & \ms{59.09}{0.62} & \ms{60.11}{0.37} & \ms{$+1.01$}{0.43} \\
SAR~\citep{niu2023sar}         & \ms{51.52}{0.00} & \ms{51.96}{0.00} & \ms{$+0.43$}{0.00} & \ms{59.71}{0.00} & \ms{61.07}{0.00} & \ms{$+1.35$}{0.00} \\
FOA~\citep{niu2024foa} & \ms{42.42}{0.48} & \ms{42.81}{0.30} & \ms{$+0.39$}{0.60} & \ms{59.80}{0.10} & \ms{60.41}{0.14} & \ms{$+0.61$}{0.09} \\
SPA~\citep{niu2025spa}        & \ms{51.22}{0.09} & \ms{52.98}{0.12} & \ms{$+1.76$}{0.21} & \ms{63.53}{0.08} & \ms{64.81}{0.06} & \ms{$+1.28$}{0.13} \\
NEO~\citep{murphy2026neo}     & \ms{40.97}{0.00} & \ms{41.97}{0.00} & \ms{$+1.00$}{0.00} & \ms{56.81}{0.00} & \ms{58.16}{0.00} & \ms{$+1.34$}{0.00} \\
\bottomrule
\end{tabular}%
}
\vspace{-10pt}
\end{table}

At $r_{\mathrm{base}}{=}8$, $\gamma{=}2$ the schedule improves over flat for every method on both backbones ($+0.39$ to $+3.57$pp, including the gradient-free FOA and NEO), and over both operating points it is positive in all 24 cells (Table~\ref{tab:ttavitb}). The gain is additive rather than absorbed: on top of the adapted baseline the schedule still adds $+1.36$pp on ViT-S and $+1.14$pp on ViT-B. SPA has the strongest flat baseline ($63.53$ on ViT-B) and still gains $+1.28$, so the effect is not an artifact of a weak baseline: at $26\%$ fewer GFLOPs its late schedule comes within $0.1$pp of full-token accuracy on ViT-S and $0.6$pp on ViT-B, against $1.9$pp for flat on both (Appendix~\ref{app:ttavitb}). Under test-time adaptation, the late-concentrated schedule therefore keeps the compute saving and improves on the flat schedule by about a point, leaving each method's mechanism untouched.

\section{Limitations and Conclusion}
\label{sec:limitations}

\paragraph{Limitations.}
Our first-order amplification account predicts the direction and ordering of the effects, borne out in every test, but not their per-corruption magnitude, which also reflects proximity to the decision boundary (Section~\ref{sec:mechtoken}), though the amplification it predicts rises under every corruption group. The schedule is static by design, which keeps it tuning-free; adapting $\gamma$ to the estimated shift is a refinement, not a requirement for the reported gains.

\paragraph{Conclusion.}

We showed that the token-reduction schedule, how the removal budget is distributed over depth, is a first-order determinant of a token-reduced vision transformer's robustness to distribution shift. At no extra inference cost, a simple late-concentrated power-law schedule recovers $83\%$ of the accuracy gap that token reduction opens against the uncompressed model ($+1.17$pp at a $26\%$ compute reduction), and $99\%$ at a lighter $7\%$ reduction ($+0.26$pp; DeiT-S, ImageNet-C). 
The gain cannot be explained by retaining more tokens or spending extra compute; it follows from where the removal budget is placed.
We trace this to a front-loaded error budget: the earlier a reduction acts the more the network amplifies its error, and shift widens that gap. The effect is broad: nine backbones, eight further shift suites, two other modalities, a gain that grows $\sim$4$\times$ from clean to severity~5, and composability with six test-time adaptation methods, all from one static schedule applied at test time with no tuning.

\section*{AI Use Statement}

We used generative AI tools to help interpret results, edit the text for grammar and readability, draft the figure scripts, and clean up code. The authors checked every interpretation and every figure against the underlying data, reviewed every edited passage, and verified that all AI-edited code reproduces the reported numbers. We take responsibility for the final content of this work, including text, claims or artifacts produced with the aid of generative AI.

\section*{Ethics Statement}

This work uses only publicly available benchmarks and publicly released model checkpoints. It involves no human subjects, collects no new data, and introduces no new dataset. The contribution is methodological and inherits, and neither adds to nor removes, the known biases and limitations of the models and datasets it is applied to. We do not foresee ethical concerns beyond those general to machine learning research on public benchmarks.

\section*{Reproducibility Statement}

\textbf{Code and data.} 
The code is available at \url{https://github.com/chahh9808/LaterIsBetter}, with the schedule and its iso-GFLOPs calibration, the evaluation pipeline, and a launcher for every image-classification table and figure in the paper. It also lists the DomainNet-126 and NaBirds source checkpoints we finetuned with their SHA-256; those checkpoints and the video and vision-language harnesses will be released with the code. Every dataset used is a public benchmark: the ImageNet-1k validation set, ImageNet-C, ImageNet-A, ImageNet-R, ImageNet-V2, ImageNet-Sketch, DomainNet-126, ImageNet-\cbar, ImageNet-3DCC, NaBirds, Mini-Kinetics-C, and GQA. Nothing is collected or relabelled for this work beyond applying the published corruption suites.

\textbf{Models.} Every backbone but one set is a public checkpoint, named with its exact identifier in Appendix~\ref{app:impl}: DeiT-S/B and ViT-S/B/L from \texttt{timm} pretrained on ImageNet-1k (DeiT-S is the default for the main analysis), CLIP ViT-B/16 and ViT-L/14 finetuned on ImageNet-1k, MAE-pretrained ViT-B/L that read out by global average pooling, VideoMAE ViT-B finetuned on Kinetics-400, and LLaVA-1.5-7B with a CLIP ViT-L/14 encoder. The exception is DomainNet-126, where no public source checkpoint covers all five backbones, so we finetune each on the \emph{real} domain with one recipe given in Appendix~\ref{app:impl}.

\textbf{The schedule is specified in closed form.} The only thing we change in a reducer is the per-layer removal budget of Eq.~\ref{eq:schedule}, which is a function of $(r_{\mathrm{base}},\gamma,L)$ alone: no training, no per-input decision, and no learned parameter. The continuous counts are rounded to integers by largest remainder and capped at $\lfloor n_\ell/2\rfloor$, the same cap for every reducer (Appendix~\ref{app:impl}). Because GFLOPs is analytic in the schedule, the iso-GFLOPs calibration needs no measurement: for each $(r_{\mathrm{base}},\gamma)$ we read off the integer budget whose GFLOPs match the flat baseline, always rounding so the late schedule uses no more GFLOPs than flat, with GFLOPs from a single THOP counter. Table~\ref{tab:isocalib} lists every calibrated schedule used in the paper, so the exact per-layer counts behind any reported number can be reconstructed without rerunning the calibration.

\textbf{Determinism and reported variance.} Inference-only evaluation is fully deterministic: we set \texttt{cudnn.deterministic}, enable deterministic algorithms, and use a deterministic scatter-mean in the merge. The flat-versus-late comparisons therefore carry no run variance, which is why the inference-only tables report single numbers rather than error bars. The test-time-adaptation experiments (Section~\ref{sec:tta}) are the stochastic ones and are run over 3 seeds ($42,43,44$), with seed standard deviations reported alongside the means.

\textbf{Evaluation scope.} The main results use the complete standard benchmarks, so there is no subsampling uncertainty: the full 50k images per ImageNet-C corruption at severity~5, the 50k ImageNet validation set for clean accuracy, the full ImageNet-A, ImageNet-R, ImageNet-V2 and ImageNet-Sketch sets, the four DomainNet-126 splits, and the $24{,}633$ NaBirds-C test images. Two suites are exceptions, marked wherever they are reported: ImageNet-\cbar{} and ImageNet-3DCC use the standard RobustBench~\citep{croce2021robustbench} 5k-image-per-corruption subset, and Mini-Kinetics-C uses $2500$ clips over its six spatial corruptions. In the vision-language experiment the encoder schedule is matched to within $0.03\%$ of the flat schedule's GMACs (Appendix~\ref{app:impl}).

\textbf{Fairness of the compute matching.} The iso-GFLOPs comparisons are also matched in wall-clock: at the main operating point the late ($\gamma{=}2$) and flat schedules differ by under $0.5\%$ in measured throughput (batch~64, single GPU), so the comparison is fair in real latency and not only in analytic FLOPs (Appendix~\ref{app:latency}).

\textbf{Hyperparameters.} Token reduction here is training-free and inference-only, so it adds no hyperparameter beyond the schedule itself. For test-time adaptation, each method's learning rate is chosen once per method and backbone from a flat-schedule stability protocol and then frozen before the flat-versus-late comparison, with no per-schedule tuning; every other setting is that method's official default. Appendix~\ref{app:impl} lists them all.

\bibliographystyle{iclr2027_conference}
\bibliography{references}

@inproceedings{kim2021lipschitz,
  title     = {The {L}ipschitz Constant of Self-Attention},
  author    = {Kim, Hyunjik and Papamakarios, George and Mnih, Andriy},
  booktitle = {International Conference on Machine Learning (ICML)},
  year      = {2021}
}

@inproceedings{he2022mae,
  title     = {Masked Autoencoders Are Scalable Vision Learners},
  author    = {He, Kaiming and Chen, Xinlei and Xie, Saining and Li, Yanghao and Doll{\'a}r, Piotr and Girshick, Ross},
  booktitle = {Proceedings of the IEEE/CVF Conference on Computer Vision and Pattern Recognition (CVPR)},
  year      = {2022}
}

@inproceedings{bolya2023tome,
  title     = {Token Merging: Your {V}i{T} But Faster},
  author    = {Bolya, Daniel and Fu, Cheng-Yang and Dai, Xiaoliang and Zhang, Peizhao and Feichtenhofer, Christoph and Hoffman, Judy},
  booktitle = {International Conference on Learning Representations (ICLR)},
  year      = {2023}
}

@inproceedings{liang2022evit,
  title     = {Not All Patches are What You Need: Expediting Vision Transformers via Token Reorganizations},
  author    = {Liang, Youwei and Ge, Chongjian and Tong, Zhan and Song, Yibing and Wang, Jue and Xie, Pengtao},
  booktitle = {International Conference on Learning Representations (ICLR)},
  year      = {2022}
}

@inproceedings{tran2024pitome,
  title     = {Accelerating Transformers with Spectrum-Preserving Token Merging},
  author    = {Tran, Hoai-Chau and Nguyen, Duy M. H. and Nguyen, Duy M. and Nguyen, Trung-Tin and Le, Ngan and Xie, Pengtao and Sonntag, Daniel and Zou, James Y. and Nguyen, Binh T. and Niepert, Mathias},
  booktitle = {Advances in Neural Information Processing Systems (NeurIPS)},
  year      = {2024}
}

@inproceedings{fayyaz2022ats,
  title     = {Adaptive Token Sampling for Efficient Vision Transformers},
  author    = {Fayyaz, Mohsen and Koohpayegani, Soroush Abbasi and Jafari, Farnoush Rezaei and Sengupta, Sunando and Joze, Hamid Reza Vaezi and Sommerlade, Eric and Pirsiavash, Hamed and Gall, J{\"u}rgen},
  booktitle = {European Conference on Computer Vision (ECCV)},
  year      = {2022}
}

@inproceedings{haurum2024atc,
  title     = {Agglomerative Token Clustering},
  author    = {Haurum, Joakim Bruslund and Escalera, Sergio and Taylor, Graham W. and Moeslund, Thomas B.},
  booktitle = {European Conference on Computer Vision (ECCV)},
  year      = {2024}
}

@inproceedings{touvron2021deit,
  title     = {Training Data-Efficient Image Transformers \& Distillation through Attention},
  author    = {Touvron, Hugo and Cord, Matthieu and Douze, Matthijs and Massa, Francisco and Sablayrolles, Alexandre and J{\'e}gou, Herv{\'e}},
  booktitle = {International Conference on Machine Learning (ICML)},
  year      = {2021}
}

@inproceedings{hendrycks2019imagenetc,
  title     = {Benchmarking Neural Network Robustness to Common Corruptions and Perturbations},
  author    = {Hendrycks, Dan and Dietterich, Thomas},
  booktitle = {International Conference on Learning Representations (ICLR)},
  year      = {2019}
}

@inproceedings{dosovitskiy2021vit,
  title     = {An Image is Worth 16x16 Words: Transformers for Image Recognition at Scale},
  author    = {Dosovitskiy, Alexey and Beyer, Lucas and Kolesnikov, Alexander and Weissenborn, Dirk and Zhai, Xiaohua and Unterthiner, Thomas and Dehghani, Mostafa and Minderer, Matthias and Heigold, Georg and Gelly, Sylvain and Uszkoreit, Jakob and Houlsby, Neil},
  booktitle = {International Conference on Learning Representations (ICLR)},
  year      = {2021}
}

@inproceedings{wang2022tent,
  title     = {Tent: Fully Test-Time Adaptation by Entropy Minimization},
  author    = {Wang, Dequan and Shelhamer, Evan and Liu, Shaoteng and Olshausen, Bruno and Darrell, Trevor},
  booktitle = {International Conference on Learning Representations (ICLR)},
  year      = {2021}
}

@inproceedings{xiong2025navia,
  title     = {Neutralizing Token Aggregation via Information Augmentation for Efficient Test-Time Adaptation},
  author    = {Xiong, Yizhe and Zhou, Zihan and Liang, Yiwen and Chen, Hui and Lin, Zijia and Xu, Xinhao and Hao, Tianxiang and Zhang, Fan and Han, Jungong and Ding, Guiguang},
  booktitle = {European Conference on Computer Vision (ECCV)},
  year      = {2026}
}

@inproceedings{wang2025tca,
  title     = {Is Less More? Exploring Token Condensation as Training-free Test-time Adaptation},
  author    = {Wang, Zixin and Gong, Dong and Wang, Sen and Huang, Zi and Luo, Yadan},
  booktitle = {Proceedings of the IEEE/CVF International Conference on Computer Vision (ICCV)},
  year      = {2025}
}

@inproceedings{chen2023diffrate,
  title     = {{DiffRate}: Differentiable Compression Rate for Efficient Vision Transformers},
  author    = {Chen, Mengzhao and Shao, Wenqi and Xu, Peng and Lin, Mingbao and Zhang, Kaipeng and Chao, Fei and Ji, Rongrong and Qiao, Yu and Luo, Ping},
  booktitle = {Proceedings of the IEEE/CVF International Conference on Computer Vision (ICCV)},
  year      = {2023}
}

@article{bonnaerens2023ltmp,
  title   = {Learned Thresholds Token Merging and Pruning for Vision Transformers},
  author  = {Bonnaerens, Maxim and Dambre, Joni},
  journal = {Transactions on Machine Learning Research (TMLR)},
  year    = {2023}
}

@inproceedings{rao2021dynamicvit,
  title     = {{DynamicViT}: Efficient Vision Transformers with Dynamic Token Sparsification},
  author    = {Rao, Yongming and Zhao, Wenliang and Liu, Benlin and Lu, Jiwen and Zhou, Jie and Hsieh, Cho-Jui},
  booktitle = {Advances in Neural Information Processing Systems (NeurIPS)},
  year      = {2021}
}

@inproceedings{yin2022avit,
  title     = {{A-ViT}: Adaptive Tokens for Efficient Vision Transformer},
  author    = {Yin, Hongxu and Vahdat, Arash and Alvarez, Jose M. and Mallya, Arun and Kautz, Jan and Molchanov, Pavlo},
  booktitle = {Proceedings of the IEEE/CVF Conference on Computer Vision and Pattern Recognition (CVPR)},
  year      = {2022}
}

@inproceedings{bhojanapalli2021understanding,
  title     = {Understanding Robustness of Transformers for Image Classification},
  author    = {Bhojanapalli, Srinadh and Chakrabarti, Ayan and Glasner, Daniel and Li, Daliang and Unterthiner, Thomas and Veit, Andreas},
  booktitle = {Proceedings of the IEEE/CVF International Conference on Computer Vision (ICCV)},
  year      = {2021}
}

@inproceedings{paul2022vit,
  title     = {Vision Transformers Are Robust Learners},
  author    = {Paul, Sayak and Chen, Pin-Yu},
  booktitle = {Proceedings of the AAAI Conference on Artificial Intelligence (AAAI)},
  year      = {2022}
}

@inproceedings{tong2022videomae,
  title     = {{VideoMAE}: Masked Autoencoders are Data-Efficient Learners for Self-Supervised Video Pre-Training},
  author    = {Tong, Zhan and Song, Yibing and Wang, Jue and Wang, Limin},
  booktitle = {Advances in Neural Information Processing Systems (NeurIPS)},
  year      = {2022}
}

@article{kay2017kinetics,
  title     = {The Kinetics Human Action Video Dataset},
  author    = {Kay, Will and Carreira, Joao and Simonyan, Karen and Zhang, Brian and Hillier, Chloe and Vijayanarasimhan, Sudheendra and Viola, Fabio and Green, Tim and Back, Trevor and Natsev, Paul and Suleyman, Mustafa and Zisserman, Andrew},
  journal   = {arXiv preprint arXiv:1705.06950},
  year      = {2017}
}

@inproceedings{yi2021benchmarking,
  title     = {Benchmarking the Robustness of Spatial-Temporal Models Against Corruptions},
  author    = {Yi, Chenyu and Yang, Siyuan and Li, Haoliang and Tan, Yap-Peng and Kot, Alex C.},
  booktitle = {Advances in Neural Information Processing Systems (NeurIPS) Datasets and Benchmarks Track},
  year      = {2021}
}

@inproceedings{niu2022eata,
  title     = {Efficient Test-Time Model Adaptation without Forgetting},
  author    = {Niu, Shuaicheng and Wu, Jiaxiang and Zhang, Yifan and Chen, Yaofo and Zheng, Shijian and Zhao, Peilin and Tan, Mingkui},
  booktitle = {International Conference on Machine Learning (ICML)},
  year      = {2022}
}

@inproceedings{niu2023sar,
  title     = {Towards Stable Test-Time Adaptation in Dynamic Wild World},
  author    = {Niu, Shuaicheng and Wu, Jiaxiang and Zhang, Yifan and Wen, Zhiquan and Chen, Yaofo and Zhao, Peilin and Tan, Mingkui},
  booktitle = {International Conference on Learning Representations (ICLR)},
  year      = {2023}
}

@inproceedings{niu2024foa,
  title     = {Test-Time Model Adaptation with Only Forward Passes},
  author    = {Niu, Shuaicheng and Miao, Chunyan and Chen, Guohao and Wu, Pengcheng and Zhao, Peilin},
  booktitle = {International Conference on Machine Learning (ICML)},
  year      = {2024}
}

@inproceedings{niu2025spa,
  title     = {Self-Bootstrapping for Versatile Test-Time Adaptation},
  author    = {Niu, Shuaicheng and Chen, Guohao and Zhao, Peilin and Wang, Tianyi and Wu, Pengcheng and Shen, Zhiqi},
  booktitle = {International Conference on Machine Learning (ICML)},
  year      = {2025}
}

@inproceedings{murphy2026neo,
  title     = {{NEO}: No-Optimization Test-Time Adaptation through Latent Re-Centering},
  author    = {Murphy, Alexander and Danilowski, Michal and Chatterjee, Soumyajit and Ghosh, Abhirup},
  booktitle = {International Conference on Learning Representations (ICLR)},
  year      = {2026}
}

@article{usama2025vlmcorruptions,
  title   = {Analysing the Robustness of Vision-Language-Models to Common Corruptions},
  author  = {Usama, Muhammad and Asim, Syeda Aishah and Ali, Syed Bilal and Wasim, Syed Talal and Mansoor, Umair Bin},
  journal = {arXiv preprint arXiv:2504.13690},
  year    = {2025}
}

@inproceedings{liu2024llava15,
  title     = {Improved Baselines with Visual Instruction Tuning},
  author    = {Liu, Haotian and Li, Chunyuan and Li, Yuheng and Lee, Yong Jae},
  booktitle = {Proceedings of the IEEE/CVF Conference on Computer Vision and Pattern Recognition (CVPR)},
  year      = {2024}
}

@inproceedings{hudson2019gqa,
  title     = {{GQA}: A New Dataset for Real-World Visual Reasoning and Compositional Question Answering},
  author    = {Hudson, Drew A. and Manning, Christopher D.},
  booktitle = {Proceedings of the IEEE/CVF Conference on Computer Vision and Pattern Recognition (CVPR)},
  year      = {2019}
}

@inproceedings{mintun2021cbar,
  title     = {On Interaction Between Augmentations and Corruptions in Natural Corruption Robustness},
  author    = {Mintun, Eric and Kirillov, Alexander and Xie, Saining},
  booktitle = {Advances in Neural Information Processing Systems (NeurIPS)},
  year      = {2021}
}

@inproceedings{kar20223dcc,
  title     = {{3D} Common Corruptions and Data Augmentation},
  author    = {Kar, O{\u{g}}uzhan Fatih and Yeo, Teresa and Atanov, Andrei and Zamir, Amir},
  booktitle = {Proceedings of the IEEE/CVF Conference on Computer Vision and Pattern Recognition (CVPR)},
  year      = {2022}
}

@inproceedings{croce2021robustbench,
  title     = {{RobustBench}: a standardized adversarial robustness benchmark},
  author    = {Croce, Francesco and Andriushchenko, Maksym and Sehwag, Vikash and Debenedetti, Edoardo and Flammarion, Nicolas and Chiang, Mung and Mittal, Prateek and Hein, Matthias},
  booktitle = {Advances in Neural Information Processing Systems (NeurIPS) Datasets and Benchmarks Track},
  year      = {2021}
}

@inproceedings{lecun1990obd,
  author = {LeCun, Yann and Denker, John and Solla, Sara},
  booktitle = {Advances in Neural Information Processing Systems (NeurIPS)},
  editor = {D. Touretzky},
  pages = {},
  publisher = {Morgan-Kaufmann},
  title = {Optimal Brain Damage},
  volume = {2},
  year = {1989}
}

@inproceedings{michel2019sixteen,
  title     = {Are Sixteen Heads Really Better than One?},
  author    = {Michel, Paul and Levy, Omer and Neubig, Graham},
  booktitle = {Advances in Neural Information Processing Systems (NeurIPS)},
  year      = {2019}
}

@inproceedings{novak2018sensitivity,
  title     = {Sensitivity and Generalization in Neural Networks: An Empirical Study},
  author    = {Novak, Roman and Bahri, Yasaman and Abolafia, Daniel A. and Pennington, Jeffrey and Sohl-Dickstein, Jascha},
  booktitle = {International Conference on Learning Representations (ICLR)},
  year      = {2018}
}

@inproceedings{poole2016exponential,
  title     = {Exponential Expressivity in Deep Neural Networks through Transient Chaos},
  author    = {Poole, Ben and Lahiri, Subhaneil and Raghu, Maithra and Sohl-Dickstein, Jascha and Ganguli, Surya},
  booktitle = {Advances in Neural Information Processing Systems (NeurIPS)},
  year      = {2016}
}

@inproceedings{pennington2017isometry,
  title     = {Resurrecting the Sigmoid in Deep Learning through Dynamical Isometry: Theory and Practice},
  author    = {Pennington, Jeffrey and Schoenholz, Samuel S. and Ganguli, Surya},
  booktitle = {Advances in Neural Information Processing Systems (NeurIPS)},
  year      = {2017}
}

@inproceedings{vanhorn2015nabirds,
  title     = {Building a Bird Recognition App and Large Scale Dataset with Citizen Scientists: The Fine Print in Fine-Grained Dataset Collection},
  author    = {Van Horn, Grant and Branson, Steve and Farrell, Ryan and Haber, Scott and Barry, Jessie and Ipeirotis, Panos and Perona, Pietro and Belongie, Serge},
  booktitle = {Proceedings of the IEEE Conference on Computer Vision and Pattern Recognition (CVPR)},
  year      = {2015}
}

@inproceedings{hendrycks2021many,
  title     = {The Many Faces of Robustness: A Critical Analysis of Out-of-Distribution Generalization},
  author    = {Hendrycks, Dan and Basart, Steven and Mu, Norman and Kadavath, Saurav and Wang, Frank and Dorundo, Evan and Desai, Rahul and Zhu, Tyler and Parajuli, Samyak and Guo, Mike and Song, Dawn and Steinhardt, Jacob and Gilmer, Justin},
  booktitle = {Proceedings of the IEEE/CVF International Conference on Computer Vision (ICCV)},
  year      = {2021}
}

@inproceedings{wang2019sketch,
  title     = {Learning Robust Global Representations by Penalizing Local Predictive Power},
  author    = {Wang, Haohan and Ge, Songwei and Lipton, Zachary and Xing, Eric P.},
  booktitle = {Advances in Neural Information Processing Systems (NeurIPS)},
  year      = {2019}
}

@inproceedings{hendrycks2021nae,
  title     = {Natural Adversarial Examples},
  author    = {Hendrycks, Dan and Zhao, Kevin and Basart, Steven and Steinhardt, Jacob and Song, Dawn},
  booktitle = {Proceedings of the IEEE/CVF Conference on Computer Vision and Pattern Recognition (CVPR)},
  year      = {2021}
}

@inproceedings{recht2019imagenetv2,
  title     = {Do {ImageNet} Classifiers Generalize to {ImageNet}?},
  author    = {Recht, Benjamin and Roelofs, Rebecca and Schmidt, Ludwig and Shankar, Vaishaal},
  booktitle = {International Conference on Machine Learning (ICML)},
  year      = {2019}
}

@inproceedings{peng2019domainnet,
  title     = {Moment Matching for Multi-Source Domain Adaptation},
  author    = {Peng, Xingchao and Bai, Qinxun and Xia, Xide and Huang, Zijun and Saenko, Kate and Wang, Bo},
  booktitle = {Proceedings of the IEEE/CVF International Conference on Computer Vision (ICCV)},
  year      = {2019}
}

@article{mcnemar1947,
  title   = {Note on the Sampling Error of the Difference Between Correlated Proportions or Percentages},
  author  = {McNemar, Quinn},
  journal = {Psychometrika},
  volume  = {12},
  number  = {2},
  pages   = {153--157},
  year    = {1947}
}

@inproceedings{heo2023dsm,
  title     = {Training-Free Acceleration of {ViTs} with Delayed Spatial Merging},
  author    = {Heo, Jung Hwan and Azizi, Seyedarmin and Fayyazi, Arash and Pedram, Massoud},
  booktitle = {Workshop on Efficient Systems for Foundation Models (ES-FoMo) at the International Conference on Machine Learning (ICML)},
  year      = {2024}
}

@inproceedings{lee2025atm,
  title     = {Lossless Token Merging Even Without Fine-Tuning in Vision Transformers},
  author    = {Lee, Jaeyeon and Choi, Dong-Wan},
  booktitle = {European Conference on Artificial Intelligence (ECAI)},
  year      = {2025}
}

@inproceedings{wang2025dymu,
  title     = {{DyMU}: Dynamic Merging and Virtual Unmerging for Efficient Variable-Length {VLMs}},
  author    = {Wang, Zhenhailong and Purushwalkam, Senthil and Xiong, Caiming and Savarese, Silvio and Ji, Heng and Xu, Ran},
  booktitle = {Advances in Neural Information Processing Systems (NeurIPS)},
  year      = {2025}
}

@inproceedings{shang2025prumerge,
  title     = {{LLaVA-PruMerge}: Adaptive Token Reduction for Efficient Large Multimodal Models},
  author    = {Shang, Yuzhang and Cai, Mu and Xu, Bingxin and Lee, Yong Jae and Yan, Yan},
  booktitle = {Proceedings of the IEEE/CVF International Conference on Computer Vision (ICCV)},
  year      = {2025}
}

@inproceedings{radford2021clip,
  title     = {Learning Transferable Visual Models From Natural Language Supervision},
  author    = {Radford, Alec and Kim, Jong Wook and Hallacy, Chris and Ramesh, Aditya and Goh, Gabriel and Agarwal, Sandhini and Sastry, Girish and Askell, Amanda and Mishkin, Pamela and Clark, Jack and Krueger, Gretchen and Sutskever, Ilya},
  booktitle = {International Conference on Machine Learning (ICML)},
  year      = {2021}
}

\clearpage
\appendix

\section{Implementation Details}
\label{app:impl}

\textbf{Backbones.} Image classification uses DeiT-S/B (\texttt{deit\_small/base\_patch16\_224}) and
ViT-S/B/L (\texttt{vit\_small/base/large\_patch16\_224}) from \texttt{timm}, pretrained on
ImageNet-1k; DeiT-S is the default for the main analysis. Table~\ref{tab:arch} adds two
contrastively pretrained backbones finetuned on ImageNet-1k
(\texttt{vit\_base\_patch16\_clip\_224.openai\_ft\_in1k} and
\texttt{vit\_large\_patch14\_clip\_224.openai\_ft\_in1k}, the latter carrying $256$ tokens rather
than $196$). The twelve-layer backbones there run at $r_{\mathrm{base}}{=}8$; on the
twenty-four-layer ones that per-layer rate would cut nearly twice the compute, so we match the
compute cut instead, at $r_{\mathrm{base}}{=}4$ for the ViT-L/16 models and $5$ for CLIP ViT-L/14, a
$24$--$25\%$ reduction in every case, with the late budget calibrated exactly as elsewhere. Video
uses VideoMAE ViT-B~\citep{tong2022videomae} finetuned on Kinetics-400~\citep{kay2017kinetics},
tokenizing a clip into $1568$ space-time tokens (tubelet embedding, no class token). Vision-language
QA uses LLaVA-1.5-7B~\citep{liu2024llava15} with a CLIP ViT-L/14 vision encoder ($576$ visual
tokens). The schedule acts inside this encoder; of prior VLM token reduction,
PruMerge~\citep{shang2025prumerge} reduces the encoder's output and DyMU~\citep{wang2025dymu} merges
inside it with a constant per-layer target. MAE-pretrained ViT-B and ViT-L~\citep{he2022mae} finetuned on ImageNet-1k (official
checkpoints) classify by global average pooling over the patch tokens rather than a class token;
they appear both in Table~\ref{tab:arch} and as the readout control of
Appendix~\ref{app:finalblock}. For DomainNet-126 no public source checkpoint covers all five
backbones, so we finetune each one on the \emph{real} domain with a single recipe (5 epochs, AdamW,
lr $5{\times}10^{-5}$ cosine, label smoothing $0.1$, each model's own normalization), reaching
$91.6$ to $94.3$ on the held-out real split. The recipe is fixed once and shared by all five, and
the schedule comparison that follows is inference-only on the frozen weights, so the flat and the
late arm see the identical model and no part of the finetuning can prefer one to the other. The
held-out real split is the control for the finetuning itself, and the gain is absent there. We
release these checkpoints with the code.

\textbf{Token reduction and iso-GFLOPs.} The power-law schedule (Eq.~\ref{eq:schedule}) is realized
with ToMe bipartite soft matching by default, and with EViT~\citep{liang2022evit},
ATS~\citep{fayyaz2022ats}, ATC~\citep{haurum2024atc}, and PiToMe~\citep{tran2024pitome} for the
cross-method study (the same schedule, each method's own scoring). The per-layer budget reaches each
reducer through its native interface (a keep-rate for EViT and ATS, a merge count for ToMe and
PiToMe, a cluster count for ATC), so the schedule is identical and only the scoring rule differs;
ToMe and PiToMe use proportional attention, the merge is reweighted by a token-survival correction,
and ATC clusters with average linkage. The continuous per-layer counts of Eq.~\ref{eq:schedule} are
rounded to integers by largest remainder, and, because ToMe's bipartite matching removes at most
half the tokens present at a layer, each count is capped at $\lfloor n_\ell/2\rfloor$; we apply the
same cap to every reducer so the realized schedule stays identical across methods. At strongly
late-concentrated settings this cap lowers the deepest layers, where few tokens remain, below the
nominal power law (the faster late drop in Figure~\ref{fig:placement}), so removal cannot be pushed
arbitrarily late. GFLOPs are computed with a single THOP counter; since GFLOPs is analytic in the
schedule, each $(r_{\mathrm{base}},\gamma)$ is calibrated by reading off the integer per-layer
budget whose GFLOPs match the flat baseline, rounded so the late schedule never exceeds the flat
GFLOPs (Table~\ref{tab:isocalib}). Inference is fully deterministic (\texttt{cudnn.deterministic},
deterministic algorithms, and a deterministic scatter-mean in the merge), so schedule comparisons
carry no seed variance. Tent, SAR and NEO have no stochastic step, so their three seeds coincide
(std $0.00$).

\textbf{Benchmarks.} ImageNet-C~\citep{hendrycks2019imagenetc}: 15 corruptions, severity~5, 50k
images per corruption; clean $=$ the 50k ImageNet validation set.
ImageNet-A~\citep{hendrycks2021nae} ($7{,}500$ natural adversarial images, evaluated through its
$200$-class mask), ImageNet-R~\citep{hendrycks2021many} (renditions), the matched-frequency split of
ImageNet-V2~\citep{recht2019imagenetv2} ($10{,}000$ images) and
ImageNet-Sketch~\citep{wang2019sketch} use their full sets. DomainNet-126~\citep{peng2019domainnet}:
models finetuned on \emph{real} and evaluated on sketch ($24{,}147$), clipart ($18{,}523$) and
painting ($30{,}042$), with the held-out real split ($6{,}962$) as an in-domain control.
ImageNet-\cbar{}~\citep{mintun2021cbar} and ImageNet-3DCC~\citep{kar20223dcc} use the standard
RobustBench~\citep{croce2021robustbench} $5$k-image-per-corruption subset at severity~5. NaBirds-C:
DeiT-S and ViT-S finetuned on NaBirds~\citep{vanhorn2015nabirds} ($555$ species), evaluated under
the 15 ImageNet-C corruptions at severity~5 ($24{,}633$ test images). Mini-Kinetics-C
\citep{yi2021benchmarking}: the six \emph{spatial} corruptions at severity~5 over $2500$ clips
spanning $400$ classes (the six temporal corruptions are out of scope).
GQA-C~\citep{usama2025vlmcorruptions}: GQA~\citep{hudson2019gqa} testdev-balanced under corruption;
because VLM inference is costly we evaluate the three corruptions that benchmark flags as hardest
(frost, impulse noise, glass blur) at severity~5, with the encoder schedule matched to within
$0.03\%$ of the flat schedule's GMACs.

\textbf{Test-time adaptation (Section~\ref{sec:tta}).} Tent, EATA, SAR, FOA, SPA and NEO are
run over 3 seeds ($42,43,44$) at full 50k (FOA is gradient-free, optimizing input prompts with
CMA-ES; NEO performs no optimization at all). We use
ViT rather than the DeiT backbones of the earlier sections because these adaptation methods were
developed and tuned primarily on ViT (SAR and FOA report on ViT-Base), so evaluating them in their
intended regime avoids confounding the comparison with per-backbone re-tuning. The learning rate is
selected once per method/backbone from the const-schedule stability protocol and then frozen before
the flat-versus-late comparison (no per-schedule tuning): $5\times10^{-4}$, except EATA at its
official rate ($1\times10^{-4}$ on ViT-S, $1\times10^{-3}$ on ViT-B) and Tent on ViT-B at
$1\times10^{-4}$. SPA also runs at $5\times10^{-4}$ on both backbones: its official rate
($1\times10^{-2}$, stated for ViT-B) diverges here, taking the ViT-S flat baseline to near-chance
after the first $1{,}280$ images and collapsing the ViT-B one in 8 of the 15 corruption$\times$seed
cells it was checked on; $5\times10^{-4}$ is the largest rate whose flat baseline survives every
corruption and seed. NEO has no learning
rate and no optimizer, so nothing is selected for it. EATA estimates Fisher information from
$2{,}000$ samples; Tent, SAR and NEO have no seed-dependent randomness and so are deterministic
across seeds. All methods take a single update
step with SGD (momentum $0.9$); SAR uses sharpness-aware minimization ($\rho{=}0.05$). The
reliable-sample entropy threshold is $E_0{=}0.4\ln 1000\!\approx\!2.76$ (EATA, SAR), EATA's
redundancy threshold is $0.05$. FOA optimizes $3$ input prompts with CMA-ES. SPA uses its
official ViT settings, a $0.2$ low-frequency amplitude mask and an average high-frequency noise
fraction of $0.4$, with the predictor at $5\times$ the norm-layer rate; NEO has no settings. These
are each method's official defaults, apart from SPA's learning rate, noted above.
\section{Per-Severity Detail: \texorpdfstring{$\gamma{=}1$}{gamma=1} vs.\ \texorpdfstring{$\gamma{=}2$}{gamma=2}}
\label{app:severityg1}

Table~\ref{tab:severityg1} gives the full per-severity breakdown for both exponents (companion to
the $\gamma{=}2$ body Table~\ref{tab:severity}). Both schedules' gains grow monotonically with
severity; $\gamma{=}1$ rises from $+0.27$ (clean) to $+0.90$pp (severity~5), $\gamma{=}2$ from
$+0.28$ to $+1.17$pp. The sev=0 ($\gamma{=}2$) value matches the clean improvement in
Table~\ref{tab:main}.

\begin{table}[ht]
\small
\centering
\caption{Per-severity 15-corruption mean (\%), DeiT-S, $r_{\mathrm{base}}{=}8$, iso-GFLOPs. flat =
  const ($\gamma{=}0$); $\Delta$ = late $-$ flat for $\gamma{=}1$ and the default $\gamma{=}2$; sev=0 is
  clean ImageNet val.}
\label{tab:severityg1}
\begin{tabular*}{\linewidth}{@{\extracolsep{\fill}}l r rr rr}
\toprule
Sev & flat & $\gamma{=}1$ & $\Delta_{\gamma1}$ & $\gamma{=}2$ & $\Delta_{\gamma2}$ \\
\midrule
0 (clean) & 79.40 & 79.67 & $+0.27$ & 79.68 & $+0.28$ \\
1         & 70.52 & 70.95 & $+0.43$ & 71.04 & $+0.52$ \\
2         & 64.33 & 64.82 & $+0.49$ & 64.99 & $+0.66$ \\
3         & 59.06 & 59.71 & $+0.65$ & 59.89 & $+0.83$ \\
4         & 50.06 & 50.88 & $+0.82$ & 51.12 & $+1.06$ \\
5         & 37.99 & 38.89 & $+0.90$ & 39.16 & \textbf{$+1.17$} \\
\bottomrule
\end{tabular*}
\end{table}

\section{Cross-Backbone Per-Corruption Results}
\label{app:crossbackbone}

At the aggressive $r_{\mathrm{base}}{=}12$ rate (a $37\%$ compute reduction) the default
$\gamma{=}2$ still beats flat on all four backbones evaluated there, by $+0.85$ to $+1.30$pp, and
$\gamma{=}1$ is the stronger exponent, reaching $+1.78$pp (Table~\ref{tab:archr12}). This is the
migration toward weaker concentration at larger removal budgets that Section~\ref{sec:mechanism}
predicts and Figure~\ref{fig:calib}c shows.

\begin{table}[ht]
\centering
\caption{Cross-backbone generalization at the aggressive rate $r_{\mathrm{base}}{=}12$ (ImageNet-C
  sev~5, 15-corruption mean top-1, \%, iso-GFLOPs). $\Delta$ = late $-$ flat.}
\label{tab:archr12}
\footnotesize
\begin{tabular}{l r r r}
\toprule
Backbone & flat & late $\gamma{=}1$ ($\Delta$) & late $\gamma{=}2$ ($\Delta$) \\
\midrule
DeiT-S & 36.67 & 37.88 ($+1.21$) & 37.77 ($+1.10$) \\
DeiT-B & 41.44 & 42.96 ($+1.52$) & 42.74 ($+1.30$) \\
ViT-S  & 36.07 & 37.48 ($+1.41$) & 36.92 ($+0.85$) \\
ViT-B  & 50.93 & 52.71 ($+1.78$) & 52.15 ($+1.22$) \\
\bottomrule
\end{tabular}
\end{table}

\begin{table}[ht]
\centering
\caption{Per-corruption top-1 (\%), ViT-B and ViT-S, sev=5, 50k samples per corruption, at the
  main operating point $r_{\mathrm{base}}{=}8$. flat = const ($\gamma{=}0$); late = $\gamma{=}2$.
  $\Delta$ = late $-$ flat. The late schedule uses no more GFLOPs than flat on each backbone, and all
  15 corruptions improve on both.}
\label{tab:crossbackbone}
\begin{tabular}{lrrr rrr}
\toprule
 & \multicolumn{3}{c}{ViT-B} & \multicolumn{3}{c}{ViT-S} \\
\cmidrule(lr){2-4}\cmidrule(lr){5-7}
Corruption & flat & late & $\Delta$ & flat & late & $\Delta$ \\
\midrule
gaussian\_noise    & 53.64 & 56.30 & $+$2.66 & 31.51 & 33.38 & $+$1.87 \\
shot\_noise        & 53.85 & 56.16 & $+$2.31 & 30.93 & 32.54 & $+$1.61 \\
impulse\_noise     & 54.11 & 56.92 & $+$2.82 & 31.08 & 32.93 & $+$1.85 \\
defocus\_blur      & 44.74 & 46.98 & $+$2.23 & 30.26 & 31.78 & $+$1.52 \\
glass\_blur        & 35.04 & 35.57 & $+$0.53 & 22.61 & 23.00 & $+$0.39 \\
motion\_blur       & 50.99 & 52.55 & $+$1.56 & 39.64 & 41.01 & $+$1.37 \\
zoom\_blur         & 42.48 & 44.17 & $+$1.69 & 30.78 & 31.92 & $+$1.14 \\
snow               & 60.54 & 61.48 & $+$0.94 & 41.90 & 43.17 & $+$1.27 \\
frost              & 58.78 & 60.39 & $+$1.60 & 41.28 & 43.26 & $+$1.98 \\
fog                & 62.44 & 64.53 & $+$2.08 & 44.51 & 46.40 & $+$1.90 \\
brightness         & 76.74 & 77.27 & $+$0.53 & 69.16 & 70.19 & $+$1.03 \\
contrast           & 29.92 & 32.44 & $+$2.52 & 14.45 & 16.32 & $+$1.87 \\
elastic\_transform & 45.46 & 45.81 & $+$0.35 & 35.50 & 36.80 & $+$1.30 \\
pixelate           & 65.17 & 66.39 & $+$1.22 & 52.42 & 53.78 & $+$1.36 \\
jpeg\_compression  & 66.96 & 67.42 & $+$0.47 & 54.78 & 55.44 & $+$0.66 \\
\midrule
\textbf{Mean}      & 53.39 & 54.96 & $\mathbf{+1.57}$ & 38.05 & 39.46 & $\mathbf{+1.41}$ \\
\bottomrule
\end{tabular}
\end{table}

\section{Natural Shift: ImageNet-A, ImageNet-V2, and DomainNet-126}
\label{app:naturalshift}

Table~\ref{tab:generalization} reports these settings on ViT-S. Table~\ref{tab:naturalshift} gives
all five backbones with paired statistics: for every setting we dump per-image correctness for the
flat and the late run on the identical images, then compare the two with a $95\%$ paired bootstrap
interval ($10^4$ resamples) and an exact McNemar test. ImageNet-A~\citep{hendrycks2021nae} uses its
$200$-class mask and ImageNet-V2~\citep{recht2019imagenetv2} the matched-frequency split. For
DomainNet-126~\citep{peng2019domainnet} each backbone is fine-tuned on the \emph{real} domain with
one recipe and evaluated on the three target domains, plus the held-out \emph{real} split as an
in-domain control. Uncompressed accuracy matches the published checkpoints within $0.1$pp
everywhere.

The late schedule wins on ImageNet-A on all five backbones by $+1.24$ to $+2.60$pp, the largest
gains we measure, and on ViT-B, DeiT-S and CLIP ViT-B the reduced model also passes the uncompressed
one ($49.83$ vs $49.67$, $19.73$ vs $18.68$, $48.13$ vs $47.31$) while using $74\%$ of its compute.
The DomainNet target domains are positive in $12$ of $15$ cells, the exceptions being DeiT-S on
sketch and clipart, which are the two most abstract domains; painting, the most photographic, is
positive on all five. The in-domain control is within $\pm0.24$pp everywhere and significant
nowhere, so the schedule neither helps nor hurts when the test distribution matches the training
one.

\begin{table}[ht]
\centering
\caption{Natural shift across backbones, $\Delta$ = late $\gamma{=}2$ $-$ flat (pp) at iso-GFLOPs,
  $r_{\mathrm{base}}{=}8$. $^{*}$ marks an exact McNemar $p<0.05$; those cells also have a $95\%$ paired
  bootstrap interval excluding zero. DN = DomainNet-126, whose \emph{real} column is the in-domain
  control.}
\label{tab:naturalshift}
\footnotesize
\begin{tabular}{l rr rrr r}
\toprule
 & \multicolumn{2}{c}{ImageNet} & \multicolumn{3}{c}{DN target domains} & DN in-domain \\
\cmidrule(lr){2-3}\cmidrule(lr){4-6}\cmidrule(lr){7-7}
Backbone & A & V2 & sketch & clipart & painting & real \\
\midrule
ViT-S      & $+1.24^{*}$ & $+0.59^{*}$ & $+0.63^{*}$ & $+0.56^{*}$ & $+0.90^{*}$ & $+0.04$ \\
ViT-B      & $+2.60^{*}$ & $+0.09$     & $+0.35^{*}$ & $+0.28^{*}$ & $+0.46^{*}$ & $-0.04$ \\
DeiT-S     & $+1.67^{*}$ & $+0.38^{*}$ & $-0.10$     & $-0.01$     & $+0.62^{*}$ & $+0.11$ \\
DeiT-B     & $+1.96^{*}$ & $+0.55^{*}$ & $+0.47^{*}$ & $+0.33^{*}$ & $+0.82^{*}$ & $+0.24$ \\
CLIP ViT-B & $+2.56^{*}$ & $+0.16$     & $+0.29^{*}$ & $+0.19^{*}$ & $+0.56^{*}$ & $-0.03$ \\
\midrule
mean       & $+2.01$ & $+0.35$ & $+0.33$ & $+0.27$ & $+0.67$ & $+0.06$ \\
\bottomrule
\end{tabular}
\end{table}

\section{Fine-Grained Shift: NaBirds-C}
\label{app:nabirds}

To test transfer beyond ImageNet-scale classification, we apply the same iso-GFLOPs
const-versus-late comparison to NaBirds-C: a DeiT-S and a ViT-S finetuned on
NaBirds~\citep{vanhorn2015nabirds} (555 fine-grained bird species), evaluated under the 15
ImageNet-C corruptions at severity~5 (24{,}633 test images), inference-only. Operating point: const
$r_{\mathrm{base}}{=}8$ (3.199G) vs.\ late $\gamma{=}2$ ($r_{\mathrm{base}}{=}8$, $\bar{r}{=}16$,
3.147G $\le$ flat). At the default $\gamma{=}2$ late beats flat on all 15 corruptions for both
backbones (Table~\ref{tab:nabirds}), mean $+0.92$pp (DeiT-S) and $+0.77$pp (ViT-S); for DeiT-S
contrast is near-degenerate at $\sim$2\%.

\begin{table}[ht]
\centering
\caption{NaBirds-C per-corruption top-1 (\%), DeiT-S and ViT-S (NaBirds-finetuned, 555 classes),
  sev=5, 24{,}633 images. flat = const ($r_{\mathrm{base}}{=}8$); late = $\gamma{=}2$ (the default,
  $\bar{r}{=}16$, 3.147G $\le$ flat). $\Delta$ = late $-$ flat.}
\label{tab:nabirds}
\begin{tabular}{lrrr rrr}
\toprule
 & \multicolumn{3}{c}{DeiT-S} & \multicolumn{3}{c}{ViT-S} \\
\cmidrule(lr){2-4}\cmidrule(lr){5-7}
Corruption & flat & late & $\Delta$ & flat & late & $\Delta$ \\
\midrule
gaussian\_noise   & 23.41 & 24.88 & $+$1.47 & 14.85 & 16.04 & $+$1.19 \\
shot\_noise       & 23.78 & 24.95 & $+$1.17 & 16.23 & 17.83 & $+$1.60 \\
impulse\_noise    & 26.02 & 27.47 & $+$1.45 & 15.60 & 16.76 & $+$1.16 \\
defocus\_blur     & 16.49 & 16.70 & $+$0.21 &  6.98 &  7.03 & $+$0.05 \\
glass\_blur       & 16.92 & 17.53 & $+$0.61 & 17.29 & 17.40 & $+$0.11 \\
motion\_blur      & 25.16 & 25.87 & $+$0.71 & 20.76 & 20.98 & $+$0.22 \\
zoom\_blur        & 33.78 & 34.75 & $+$0.97 & 36.74 & 37.34 & $+$0.60 \\
snow              & 44.46 & 45.83 & $+$1.37 & 46.56 & 47.18 & $+$0.62 \\
frost             & 46.81 & 48.22 & $+$1.41 & 35.83 & 36.15 & $+$0.32 \\
fog               & 45.56 & 46.13 & $+$0.57 & 38.26 & 39.41 & $+$1.15 \\
brightness        & 63.37 & 63.87 & $+$0.50 & 67.09 & 67.38 & $+$0.29 \\
contrast          &  2.17 &  2.18 & $+$0.01 & 18.11 & 19.38 & $+$1.27 \\
elastic\_transform& 45.00 & 46.45 & $+$1.45 & 47.89 & 49.21 & $+$1.32 \\
pixelate          & 33.15 & 34.28 & $+$1.13 & 38.26 & 39.17 & $+$0.91 \\
jpeg\_compression & 44.98 & 45.76 & $+$0.78 & 45.50 & 46.24 & $+$0.74 \\
\midrule
\textbf{Mean}     & 32.74 & 33.66 & $\mathbf{+0.92}$ & 31.06 & 31.83 & $\mathbf{+0.77}$ \\
\bottomrule
\end{tabular}
\end{table}

\section{Video: Per-Corruption Results}
\label{app:video}

Table~\ref{tab:video} gives the per-corruption Mini-Kinetics-C results behind
Table~\ref{tab:generalization} (VideoMAE ViT-B, six spatial corruptions, severities 3 and 5). The
gain concentrates on structure-degrading corruptions (shot noise, fog, rain) and is near-neutral on
mild photometric shifts (contrast, brightness, saturate), which leave accuracy near its clean value.
Clean single-clip, single-crop top-1 is 76.8\% on this 2500-clip subset (the published 80.9\% uses
multi-view inference). Mini-Kinetics-C also defines six temporal corruptions; these probe a
redundancy the spatial-token schedule does not act on, so we leave them out of scope.

\begin{table}[ht]
\centering
\caption{Cross-modality, per-corruption. VideoMAE ViT-B on Kinetics-400 with ToMe, inference-only,
  iso-GFLOPs (flat $r{=}48$ vs.\ late $\gamma{=}2$, both $\approx$142.6 GFLOPs, $\sim$20\% reduction),
  six spatial corruptions of Mini-Kinetics-C \citep{yi2021benchmarking}, 2500 clips.
  $\Delta$ = late $-$ flat.}
\label{tab:video}
\begin{tabular*}{\linewidth}{@{\extracolsep{\fill}}lrrrrrr}
\toprule
 & \multicolumn{3}{c}{Severity 3} & \multicolumn{3}{c}{Severity 5} \\
\cmidrule(lr){2-4}\cmidrule(lr){5-7}
Corruption & flat & late & $\Delta$ & flat & late & $\Delta$ \\
\midrule
Shot noise  & 67.40 & 68.28 & $+0.88$ & 50.04 & 51.68 & $+1.64$ \\
Fog         & 70.92 & 71.36 & $+0.44$ & 66.80 & 67.96 & $+1.16$ \\
Contrast    & 76.08 & 76.08 & $+0.00$ & 73.20 & 73.08 & $-0.12$ \\
Brightness  & 75.84 & 75.52 & $-0.32$ & 71.72 & 71.80 & $+0.08$ \\
Saturate    & 75.64 & 75.76 & $+0.12$ & 65.60 & 65.52 & $-0.08$ \\
Rain        & 73.48 & 73.88 & $+0.40$ & 69.68 & 70.68 & $+1.00$ \\
\midrule
Mean        & 73.23 & 73.48 & $+0.25$ & 66.17 & 66.79 & $+0.62$ \\
\bottomrule
\end{tabular*}
\end{table}

\section{ImageNet-R / ImageNet-Sketch: All Backbones}
\label{app:rsketch}

Table~\ref{tab:rsketchfull} gives the full ImageNet-R / ImageNet-Sketch results behind
Table~\ref{tab:generalization}, for the four supervised backbones at $r_{\mathrm{base}}{=}8$,
$\gamma{=}2$. All eight cells are positive (mean $+0.28$pp); the small backbones (DeiT-S, ViT-S)
gain more than the base backbones (DeiT-B, ViT-B). Inference-only, iso-GFLOPs (late $\le$ flat),
full sets.

\begin{table}[ht]
\centering
\caption{ImageNet-R / ImageNet-Sketch, all backbones. $r_{\mathrm{base}}{=}8$, late $\gamma{=}2$,
  iso-GFLOPs. $\Delta$ = late $-$ flat (top-1 pp). R = 30{,}000 images, Sketch = 50{,}889.}
\label{tab:rsketchfull}
\begin{tabular}{l rrr rrr}
\toprule
& \multicolumn{3}{c}{ImageNet-R} & \multicolumn{3}{c}{ImageNet-Sketch} \\
\cmidrule(lr){2-4}\cmidrule(lr){5-7}
Backbone & flat & late & $\Delta$ & flat & late & $\Delta$ \\
\midrule
DeiT-S & 42.14 & 42.45 & $+0.31$ & 29.40 & 29.68 & $+0.28$ \\
DeiT-B & 44.56 & 44.63 & $+0.07$ & 32.21 & 32.42 & $+0.21$ \\
ViT-S  & 45.90 & 46.57 & $+0.67$ & 32.89 & 33.33 & $+0.44$ \\
ViT-B  & 58.77 & 58.85 & $+0.08$ & 44.92 & 45.09 & $+0.17$ \\
\bottomrule
\end{tabular}
\end{table}

\section{ImageNet-\texorpdfstring{\cbar}{C-bar} and ImageNet-3DCC: Per-Corruption}
\label{app:cbar3dcc}

Tables~\ref{tab:cbar} and~\ref{tab:3dcc} give the per-corruption results behind
Table~\ref{tab:generalization} for the two harder synthetic corruption suites,
ImageNet-\cbar{}~\citep{mintun2021cbar} (10 corruptions) and ImageNet-3DCC~\citep{kar20223dcc} (12
corruptions), for DeiT-S and ViT-S at $r_{\mathrm{base}}{=}8$, late $\gamma{=}2$, iso-GFLOPs (late
$\le$ flat), severity~5, on the standard RobustBench 5k-image subset. The body table reports the
stronger-gaining ViT-S; both backbones improve on average on both suites.

\begin{table}[ht]
\centering
\caption{ImageNet-\cbar{} per-corruption top-1 (\%), DeiT-S and ViT-S, $r_{\mathrm{base}}{=}8$, late $\gamma{=}2$, iso-GFLOPs, severity~5, 5k-image subset. $\Delta$ = late $-$ flat.}
\label{tab:cbar}
\begin{tabular}{l rrr rrr}
\toprule
 & \multicolumn{3}{c}{DeiT-S} & \multicolumn{3}{c}{ViT-S} \\
\cmidrule(lr){2-4}\cmidrule(lr){5-7}
Corruption & flat & late & $\Delta$ & flat & late & $\Delta$ \\
\midrule
blue noise        & 50.36 & 50.62 & $+0.26$ & 53.08 & 55.10 & $+2.02$ \\
brownish noise    & 66.40 & 67.36 & $+0.96$ & 60.42 & 62.34 & $+1.92$ \\
caustic refr.     & 53.60 & 54.46 & $+0.86$ & 51.32 & 53.12 & $+1.80$ \\
checkerboard      & 62.66 & 63.58 & $+0.92$ & 64.06 & 65.20 & $+1.14$ \\
sine waves        & 31.42 & 31.70 & $+0.28$ & 24.62 & 26.42 & $+1.80$ \\
inv.\ sparkles    & 18.68 & 19.22 & $+0.54$ & 18.04 & 18.96 & $+0.92$ \\
perlin noise      & 69.42 & 70.22 & $+0.80$ & 65.64 & 66.88 & $+1.24$ \\
plasma noise      & 51.74 & 52.88 & $+1.14$ & 33.48 & 36.08 & $+2.60$ \\
single-freq.\ grey& 31.20 & 30.66 & $-0.54$ & 34.00 & 34.00 & $+0.00$ \\
sparkles          & 67.14 & 68.46 & $+1.32$ & 68.18 & 69.28 & $+1.10$ \\
\midrule
\textbf{Mean} & 50.26 & 50.92 & $\mathbf{+0.65}$ & 47.28 & 48.74 & $\mathbf{+1.45}$ \\
\bottomrule
\end{tabular}
\end{table}

\begin{table}[ht]
\centering
\caption{ImageNet-3DCC per-corruption top-1 (\%), DeiT-S and ViT-S, $r_{\mathrm{base}}{=}8$, late $\gamma{=}2$, iso-GFLOPs, severity~5, 5k-image subset. $\Delta$ = late $-$ flat.}
\label{tab:3dcc}
\begin{tabular}{l rrr rrr}
\toprule
 & \multicolumn{3}{c}{DeiT-S} & \multicolumn{3}{c}{ViT-S} \\
\cmidrule(lr){2-4}\cmidrule(lr){5-7}
Corruption & flat & late & $\Delta$ & flat & late & $\Delta$ \\
\midrule
bit error         & 15.50 & 16.04 & $+0.54$ & 16.70 & 16.98 & $+0.28$ \\
color quant       & 59.82 & 59.80 & $-0.02$ & 41.56 & 41.44 & $-0.12$ \\
far focus         & 50.96 & 51.62 & $+0.66$ & 53.30 & 54.14 & $+0.84$ \\
flash             & 39.28 & 39.84 & $+0.56$ & 38.98 & 39.70 & $+0.72$ \\
fog 3d            & 38.72 & 38.52 & $-0.20$ & 31.34 & 31.96 & $+0.62$ \\
h265 abr          & 43.12 & 43.46 & $+0.34$ & 45.94 & 46.44 & $+0.50$ \\
h265 crf          & 46.58 & 46.72 & $+0.14$ & 50.60 & 51.32 & $+0.72$ \\
iso noise         & 44.04 & 46.46 & $+2.42$ & 42.06 & 44.00 & $+1.94$ \\
low light         & 49.22 & 50.68 & $+1.46$ & 44.30 & 45.74 & $+1.44$ \\
near focus        & 59.30 & 60.44 & $+1.14$ & 63.04 & 64.50 & $+1.46$ \\
xy motion         & 28.22 & 29.22 & $+1.00$ & 34.44 & 35.62 & $+1.18$ \\
z motion          & 31.24 & 32.34 & $+1.10$ & 35.38 & 36.78 & $+1.40$ \\
\midrule
\textbf{Mean} & 42.17 & 42.93 & $\mathbf{+0.76}$ & 41.47 & 42.38 & $\mathbf{+0.91}$ \\
\bottomrule
\end{tabular}
\end{table}

\section{The Four ImageNet-C Corruptions Outside the Standard Fifteen}
\label{app:extrac}

ImageNet-C ships four further corruptions that the standard benchmark, and so the rest of this
paper, leaves out: speckle noise, Gaussian blur, spatter and saturate. We run them at the main
operating point to check that the fifteen reported throughout are not a favorable subset. The late
schedule wins on all four, by more on average ($+1.66$pp) than on the fifteen ($+1.17$pp), so the
schedule improves every corruption ImageNet-C defines.

\begin{table}[ht]
\centering
\caption{The four ImageNet-C corruptions outside the standard fifteen (DeiT-S, severity~5, 50k images
  each, $r_{\mathrm{base}}{=}8$, iso-GFLOPs, inference-only). $\Delta$ = late $-$ flat.}
\label{tab:extrac}
\small
\begin{tabular}{l r >{\columncolor{oursgreen}}r >{\columncolor{deltablue}\boldmath}r}
\toprule
Corruption & flat & late & $\Delta$ \\
\midrule
Speckle noise  & 41.67 & 43.98 & $+2.30$ \\
Gaussian blur  & 23.38 & 24.09 & $+0.71$ \\
Spatter        & 47.15 & 49.72 & $+2.57$ \\
Saturate       & 62.10 & 63.17 & $+1.08$ \\
\midrule
Mean           & 43.58 & 45.24 & $+1.66$ \\
\bottomrule
\end{tabular}
\end{table}

\section{Baseline Generalization at \texorpdfstring{$\gamma{=}1$}{gamma=1}}
\label{app:baselinesg1}

\begin{figure}[ht]
\centering
\includegraphics[width=\linewidth]{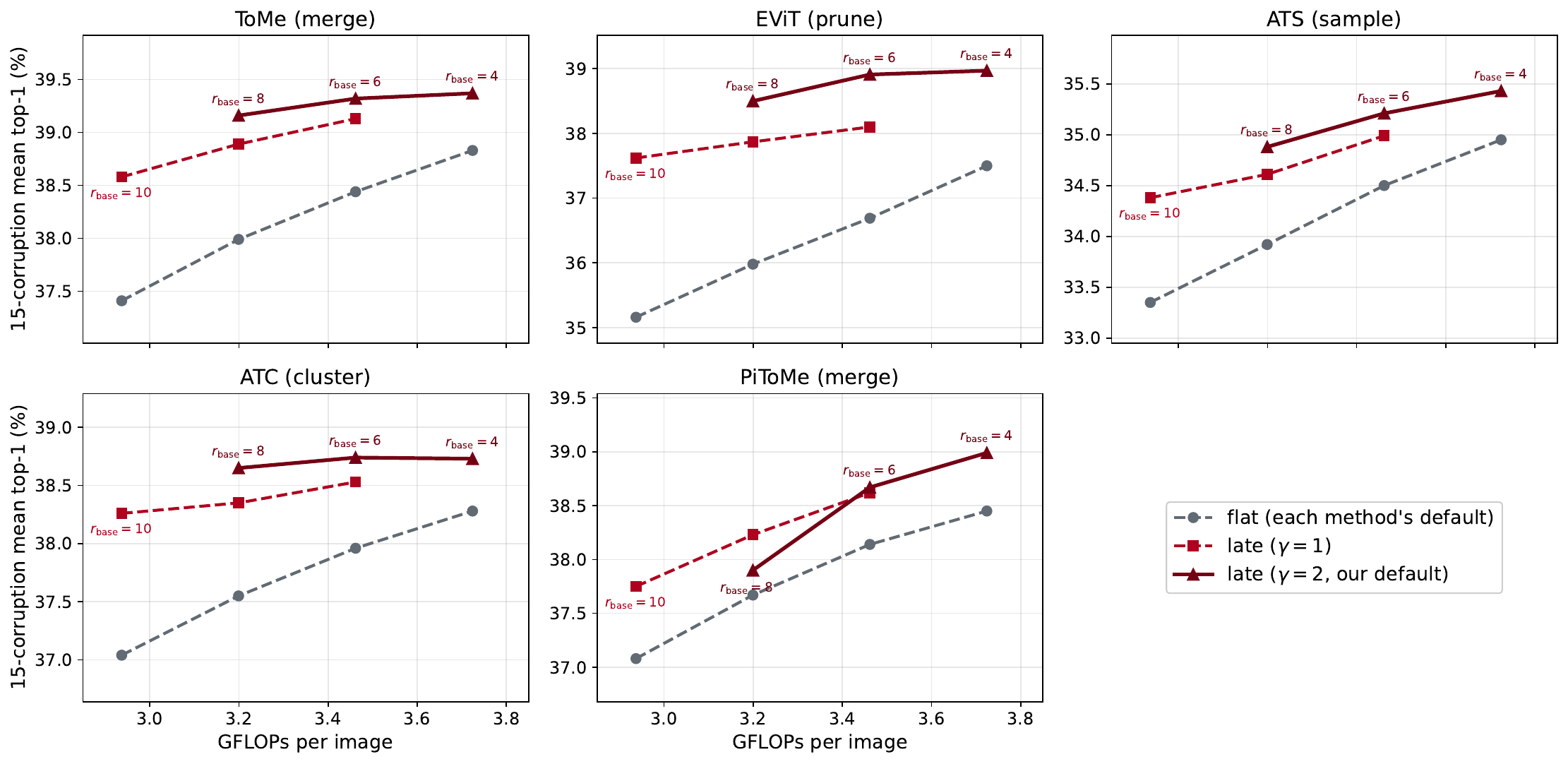}
\caption{15-corruption-mean top-1 vs.\ compute for each of the five training-free reducers (DeiT-S,
  ImageNet-C severity~5), each curve a schedule swept over compression (points labeled by
  $r_{\mathrm{base}}$). Flat and late $\gamma{=}2$ at $r_{\mathrm{base}}\in\{4,6,8\}$
  are the cells of Table~\ref{tab:baselines}, late $\gamma{=}1$ at $r_{\mathrm{base}}\in\{6,8,10\}$
  those of Table~\ref{tab:baselinesg1}; each late point is drawn at the GFLOPs of the flat schedule it
  is calibrated to (its own cost is at or below that).}
\label{fig:calibreducers}
\end{figure}

Table~\ref{tab:baselinesg1} is the $\gamma{=}1$ counterpart of the body Table~\ref{tab:baselines}:
the same five training-free reducers, but with the late schedule at $\gamma{=}1$ (the previous
default) and operating points $r_{\mathrm{base}}\in\{6,8,10\}$ that include the aggressive
$r_{\mathrm{base}}{=}10$. Late beats flat in all 30 cells here as well, so the schedule benefit is
robust to the exponent and persists at the aggressive rate (where $\gamma{=}1$ is the
operating-point optimum, Figure~\ref{fig:calib}c). Figure~\ref{fig:calibreducers} plots these cells
together with the body table's as the accuracy-versus-compute frontier of Figure~\ref{fig:calib}b,
one panel per reducer; for every reducer the late curves lie above flat at every budget.

\begin{table}[ht]
\centering
\caption{Late-concentrated at $\gamma{=}1$ across five training-free reducers (companion to the
  $\gamma{=}2$ body Table~\ref{tab:baselines}). DeiT-S/B, ImageNet-C sev=5, 15-corruption mean top-1
  (\%), full 50k, iso-GFLOPs (late $\le$ flat). $\Delta$ = \textbf{late} $-$ flat (pp); all 30 cells
  positive. Operating points $r_{\mathrm{base}}\in\{6,8,10\}$.}
\label{tab:baselinesg1}
\footnotesize\setlength{\tabcolsep}{3pt}
\begin{tabular*}{\linewidth}{@{\extracolsep{\fill}}l rrr rrr rrr}
\toprule
 & \multicolumn{3}{c}{$r_{\mathrm{base}}{=}6$} & \multicolumn{3}{c}{$r_{\mathrm{base}}{=}8$} & \multicolumn{3}{c}{$r_{\mathrm{base}}{=}10$} \\
\cmidrule(lr){2-4}\cmidrule(lr){5-7}\cmidrule(lr){8-10}
Method & flat & late & $\Delta$ & flat & late & $\Delta$ & flat & late & $\Delta$ \\
\midrule
\rowcolor{groupgrey} \multicolumn{10}{l}{\emph{DeiT-S}} \\
ToMe   & 38.44 & \textbf{39.13} & $+0.69$ & 37.99 & \textbf{38.89} & $+0.90$ & 37.41 & \textbf{38.58} & $+1.17$ \\
EViT   & 36.69 & \textbf{38.10} & $+1.41$ & 35.98 & \textbf{37.87} & $+1.89$ & 35.16 & \textbf{37.62} & $+2.46$ \\
ATS    & 34.50 & \textbf{34.99} & $+0.49$ & 33.92 & \textbf{34.61} & $+0.69$ & 33.35 & \textbf{34.38} & $+1.03$ \\
ATC    & 37.96 & \textbf{38.53} & $+0.57$ & 37.55 & \textbf{38.35} & $+0.80$ & 37.04 & \textbf{38.26} & $+1.22$ \\
PiToMe & 38.14 & \textbf{38.62} & $+0.48$ & 37.67 & \textbf{38.23} & $+0.56$ & 37.08 & \textbf{37.75} & $+0.67$ \\
\midrule
\rowcolor{groupgrey} \multicolumn{10}{l}{\emph{DeiT-B}} \\
ToMe   & 45.27 & \textbf{46.00} & $+0.73$ & 44.35 & \textbf{45.35} & $+1.00$ & 43.10 & \textbf{44.46} & $+1.36$ \\
EViT   & 43.57 & \textbf{44.91} & $+1.34$ & 42.45 & \textbf{44.02} & $+1.57$ & 41.11 & \textbf{43.45} & $+2.34$ \\
ATS    & 39.79 & \textbf{40.50} & $+0.71$ & 38.81 & \textbf{39.66} & $+0.85$ & 37.64 & \textbf{39.11} & $+1.47$ \\
ATC    & 44.79 & \textbf{45.48} & $+0.69$ & 43.97 & \textbf{44.95} & $+0.98$ & 42.81 & \textbf{44.55} & $+1.74$ \\
PiToMe & 44.56 & \textbf{45.06} & $+0.50$ & 43.69 & \textbf{43.96} & $+0.27$ & 42.43 & \textbf{43.06} & $+0.63$ \\
\bottomrule
\end{tabular*}
\end{table}
\section{Composability with TTA: Full Results}
\label{app:ttavitb}

This appendix gives the full $\gamma{=}2$ composability experiment of Section~\ref{sec:tta} on both
backbones: per-rate means at $r_{\mathrm{base}}\in\{4,8\}$ (Table~\ref{tab:ttavitb}) and the
per-corruption breakdown at $r_{\mathrm{base}}{=}8$ (Table~\ref{tab:ttapercorr}), and what each
method costs on the two token budgets (Table~\ref{tab:ttacost}). The late schedule
is the default $\gamma{=}2$ throughout. Learning rate is selected once per method/backbone from the
const-schedule stability protocol and then applied identically to both schedules (no per-schedule
tuning): $5\times10^{-4}$, except EATA at its official rate ($1\times10^{-4}$ ViT-S,
$1\times10^{-3}$ ViT-B) and ViT-B Tent at $1\times10^{-4}$; FOA is gradient-free (CMA-ES prompt
optimization) and NEO optimizes nothing, so neither has a rate to select.

\begin{table}[ht]
\centering
\caption{TTA $\times$ schedule, per-rate means ($\gamma{=}2$). ImageNet-C sev=5, 15-corruption mean
  top-1, 3 seeds. full = the same method with no reduction ($197$ tokens in every block);
  flat = const ($\gamma{=}0$); late = $\gamma{=}2$, iso-GFLOPs. $\Delta$ = late $-$
  flat; small $\pm$ values are seed stds. ($r_{\mathrm{base}}{=}8$ also appears in the body,
  Table~\ref{tab:tta}.)}
\label{tab:ttavitb}
\resizebox{\linewidth}{!}{%
\begin{tabular}{lrrrrrrr}
\toprule
 & & \multicolumn{3}{c}{$r_{\mathrm{base}}=4$} & \multicolumn{3}{c}{$r_{\mathrm{base}}=8$} \\
\cmidrule(lr){3-5}\cmidrule(lr){6-8}
Method & full & flat & late & $\Delta$ & flat & late & $\Delta$ \\
\midrule
\rowcolor{groupgrey} \multicolumn{8}{l}{\emph{ViT-S}} \\
Tent & \ms{50.19}{0.00} & \ms{49.27}{0.00} & \ms{50.84}{0.00} & \ms{$+1.57$}{0.00} & \ms{48.47}{0.00} & \ms{52.04}{0.00} & \ms{$\mathbf{+3.57}$}{0.00} \\
EATA & \ms{55.20}{0.02} & \ms{54.32}{0.11} & \ms{54.84}{0.13} & \ms{$+0.52$}{0.14} & \ms{52.65}{0.16} & \ms{53.66}{0.26} & \ms{$+1.00$}{0.38} \\
SAR  & \ms{53.39}{0.00} & \ms{52.77}{0.00} & \ms{53.22}{0.00} & \ms{$+0.45$}{0.00} & \ms{51.52}{0.00} & \ms{51.96}{0.00} & \ms{$+0.43$}{0.00} \\
FOA  & \ms{45.27}{0.23} & \ms{43.97}{0.28} & \ms{44.68}{0.15} & \ms{$+0.71$}{0.36} & \ms{42.42}{0.48} & \ms{42.81}{0.30} & \ms{$+0.39$}{0.60} \\
SPA  & \ms{53.09}{0.29} & \ms{52.38}{0.11} & \ms{53.12}{0.06} & \ms{$+0.74$}{0.15} & \ms{51.22}{0.09} & \ms{52.98}{0.12} & \ms{$\mathbf{+1.76}$}{0.21} \\
NEO  & \ms{42.66}{0.00} & \ms{42.10}{0.00} & \ms{42.55}{0.00} & \ms{$+0.45$}{0.00} & \ms{40.97}{0.00} & \ms{41.97}{0.00} & \ms{$+1.00$}{0.00} \\
\midrule
\rowcolor{groupgrey} \multicolumn{8}{l}{\emph{ViT-B}} \\
Tent & \ms{56.32}{0.00} & \ms{55.86}{0.00} & \ms{56.24}{0.00} & \ms{$+0.38$}{0.00} & \ms{54.49}{0.00} & \ms{55.72}{0.00} & \ms{$+1.23$}{0.00} \\
EATA & \ms{59.81}{1.78} & \ms{59.37}{2.76} & \ms{60.50}{0.51} & \ms{$+1.13$}{2.56} & \ms{59.09}{0.62} & \ms{60.11}{0.37} & \ms{$+1.01$}{0.43} \\
SAR  & \ms{57.91}{0.00} & \ms{60.95}{0.00} & \ms{61.51}{0.00} & \ms{$+0.56$}{0.00} & \ms{59.71}{0.00} & \ms{61.07}{0.00} & \ms{$+1.35$}{0.00} \\
FOA  & \ms{62.16}{0.13} & \ms{61.80}{0.02} & \ms{61.84}{0.07} & \ms{$+0.04$}{0.09} & \ms{59.80}{0.10} & \ms{60.41}{0.14} & \ms{$+0.61$}{0.09} \\
SPA  & \ms{65.41}{0.03} & \ms{64.77}{0.03} & \ms{65.27}{0.03} & \ms{$+0.50$}{0.04} & \ms{63.53}{0.08} & \ms{64.81}{0.06} & \ms{$+1.28$}{0.13} \\
NEO  & \ms{59.00}{0.00} & \ms{58.23}{0.00} & \ms{58.77}{0.00} & \ms{$+0.54$}{0.00} & \ms{56.81}{0.00} & \ms{58.16}{0.00} & \ms{$+1.34$}{0.00} \\
\bottomrule
\end{tabular}%
}
\end{table}

At both rates late beats flat for every method on both backbones (all 24 cells positive over 3
seeds); the residual seed variance is confined to the const baselines of the gradient methods, where
adaptation occasionally collapses on a single corruption$\times$seed run. SPA and NEO carry none of
it, SPA because no run of it collapses and NEO because it is deterministic. Table~\ref{tab:ttapercorr}
breaks the $r_{\mathrm{base}}{=}8$ result down by corruption.

\begin{table}[ht]
\centering
\caption{TTA $\times$ schedule, per-corruption top-1 (\%) at $r_{\mathrm{base}}{=}8$, $\gamma{=}2$
  (sev=5, 3-seed mean). For each method, flat = const, late = $\gamma{=}2$ at iso-GFLOPs.
  Knife-edge collapses ($<\!20$\%) are where adaptation, not the schedule, fails; the schedule
  shifts them between methods (e.g.\ $\gamma{=}2$ repairs Tent's ViT-S zoom, $18.2{\to}45.1$). NEO's
  ViT-S contrast cell is low but is not one of them: it optimizes nothing and so cannot diverge, and
  it improves that cell over the $14.4$ of no adaptation.}
\label{tab:ttapercorr}
\resizebox{\linewidth}{!}{%
\begin{tabular}{ll rrrrrrrrrrrrrrr r}
\toprule
 & & gauss & shot & imp & defoc & glass & motn & zoom & snow & frost & fog & brt & cont & elas & pix & jpeg & mean \\
\midrule
\rowcolor{groupgrey} \multicolumn{18}{l}{\emph{ViT-S}} \\
\multirow{2}{*}{Tent} & flat & 42.4 & 44.3 & 44.5 & 46.5 & 45.2 & 51.3 & 18.2 & 10.9 & 55.8 & 60.8 & 73.4 & 48.9 & 57.5 & 64.2 & 63.3 & 48.5 \\
                 & late & 45.5 & 47.1 & 47.2 & 47.7 & 46.1 & 52.2 & 45.1 & 21.7 & 56.8 & 61.3 & 73.5 & 51.0 & 57.4 & 64.5 & 63.5 & 52.0 \\
\multirow{2}{*}{EATA} & flat & 42.1 & 44.1 & 44.2 & 46.3 & 45.6 & 50.5 & 46.8 & 55.5 & 54.9 & 59.8 & 73.6 & 41.6 & 57.8 & 63.9 & 63.1 & 52.7 \\
                 & late & 45.0 & 46.5 & 46.6 & 46.7 & 46.1 & 51.1 & 47.4 & 55.8 & 55.8 & 60.6 & 73.8 & 44.1 & 57.7 & 64.1 & 63.5 & 53.7 \\
\multirow{2}{*}{SAR}  & flat & 40.9 & 42.5 & 43.1 & 45.1 & 43.9 & 50.0 & 45.0 & 53.5 & 53.8 & 58.1 & 71.9 & 45.1 & 56.0 & 62.5 & 61.5 & 51.5 \\
                 & late & 43.7 & 45.2 & 45.5 & 46.2 & 44.9 & 50.8 & 45.7 & 54.2 & 54.7 & 58.8 & 72.2 & 37.0 & 55.8 & 63.0 & 61.7 & 52.0 \\
\multirow{2}{*}{FOA}  & flat & 31.4 & 31.9 & 32.4 & 36.5 & 30.2 & 40.5 & 34.4 & 45.4 & 45.8 & 50.2 & 68.4 & 38.8 & 40.9 & 51.4 & 57.9 & 42.4 \\
                 & late & 33.2 & 32.6 & 34.7 & 35.8 & 29.6 & 40.7 & 34.4 & 45.5 & 46.1 & 51.2 & 67.6 & 39.5 & 41.7 & 51.4 & 58.1 & 42.8 \\
\multirow{2}{*}{SPA}  & flat & 43.0 & 45.2 & 45.0 & 41.6 & 36.6 & 28.7 & 49.2 & 58.2 & 57.6 & 60.3 & 73.7 & 43.0 & 60.1 & 63.8 & 62.4 & 51.2 \\
                 & late & 45.6 & 47.5 & 47.9 & 43.0 & 38.6 & 35.8 & 50.0 & 58.9 & 58.6 & 61.5 & 74.0 & 45.0 & 60.8 & 64.6 & 62.8 & 53.0 \\
\multirow{2}{*}{NEO}  & flat & 33.2 & 32.8 & 32.9 & 34.3 & 25.3 & 41.5 & 34.2 & 46.1 & 45.0 & 49.4 & 70.4 & 17.6 & 41.2 & 54.1 & 56.6 & 41.0 \\
                 & late & 34.8 & 34.2 & 34.5 & 35.3 & 25.5 & 42.7 & 34.8 & 46.8 & 46.2 & 50.4 & 71.1 & 19.2 & 41.6 & 55.3 & 57.2 & 42.0 \\
\midrule
\rowcolor{groupgrey} \multicolumn{18}{l}{\emph{ViT-B}} \\
\multirow{2}{*}{Tent} & flat & 54.8 & 55.7 & 56.3 & 52.5 & 44.3 & 55.9 & 48.8 & 61.3 & 56.1 & 12.9 & 77.1 & 56.2 & 50.3 & 67.1 & 68.0 & 54.5 \\
                 & late & 57.2 & 58.2 & 59.0 & 54.0 & 45.1 & 56.9 & 50.3 & 62.6 & 57.3 & 11.8 & 77.6 & 58.9 & 50.4 & 68.0 & 68.5 & 55.7 \\
\multirow{2}{*}{EATA} & flat & 56.1 & 58.3 & 56.1 & 50.7 & 54.9 & 60.3 & 57.9 & 67.3 & 62.7 & 67.1 & 77.4 & 6.5 & 67.4 & 72.6 & 71.2 & 59.1 \\
                 & late & 57.8 & 59.4 & 59.7 & 52.1 & 55.6 & 60.9 & 58.3 & 67.6 & 63.5 & 68.0 & 79.1 & 5.6 & 68.0 & 73.8 & 72.2 & 60.1 \\
\multirow{2}{*}{SAR}  & flat & 55.1 & 56.1 & 56.3 & 53.6 & 50.7 & 57.7 & 52.3 & 62.8 & 58.4 & 64.3 & 77.2 & 56.2 & 57.4 & 69.0 & 68.6 & 59.7 \\
                 & late & 57.5 & 58.5 & 58.9 & 54.7 & 51.4 & 58.7 & 53.7 & 63.9 & 60.6 & 66.6 & 77.8 & 57.6 & 57.8 & 69.6 & 69.0 & 61.1 \\
\multirow{2}{*}{FOA}  & flat & 56.6 & 57.4 & 57.6 & 53.5 & 43.8 & 55.8 & 50.6 & 64.9 & 60.5 & 66.2 & 77.0 & 59.7 & 55.0 & 68.9 & 69.4 & 59.8 \\
                 & late & 58.1 & 59.4 & 59.6 & 54.5 & 43.4 & 56.3 & 49.5 & 65.1 & 61.5 & 65.5 & 77.5 & 61.6 & 53.3 & 69.9 & 70.8 & 60.4 \\
\multirow{2}{*}{SPA}  & flat & 59.3 & 60.4 & 60.9 & 54.3 & 52.0 & 61.9 & 57.4 & 68.5 & 65.9 & 65.0 & 79.1 & 61.0 & 63.1 & 73.7 & 70.7 & 63.5 \\
                 & late & 61.4 & 62.7 & 63.2 & 55.6 & 53.0 & 63.1 & 58.3 & 69.4 & 66.6 & 66.2 & 79.4 & 64.0 & 63.9 & 74.3 & 71.0 & 64.8 \\
\multirow{2}{*}{NEO}  & flat & 54.0 & 54.4 & 54.6 & 47.3 & 37.1 & 52.5 & 44.9 & 62.9 & 61.4 & 69.0 & 77.2 & 54.1 & 48.7 & 66.0 & 68.1 & 56.8 \\
                 & late & 56.7 & 56.6 & 57.3 & 49.1 & 37.6 & 53.8 & 46.2 & 63.7 & 62.7 & 70.0 & 77.6 & 57.1 & 48.5 & 67.0 & 68.5 & 58.2 \\
\bottomrule
\end{tabular}%
}
\end{table}

\paragraph{Knife-edge collapses.} Per corruption, the late schedule wins on the large majority of
cells (166/180). For the three gradient methods the only failures under adaptation are
\emph{knife-edge} corruptions that collapse to near-chance ($<\!20$\%). At $\gamma{=}2$ these shifts
are favorable on balance: Tent's ViT-S zoom rises from $18.2\%$ to $45.1\%$, while Tent's ViT-B fog
and EATA's ViT-B contrast are structural, staying near-chance on both schedules at every rate. The schedule thus \emph{moves} these collapses between methods
rather than creating them, which is exactly the regime merge-aware sample selection targets
(Section~\ref{sec:limitations}). The two methods that take no gradient step on the network never
collapse. FOA's few late-versus-flat regressions are modest ($\le\!2.1$pp, e.g.\ ViT-S defocus and
ViT-B elastic) and do not concentrate in any corruption family, and NEO's single regression is ViT-B
elastic at $-0.2$pp. NEO's lowest cell, ViT-S contrast at $17.6$ flat and $19.2$ late, reads as a
collapse only against the $20\%$ line: with nothing to optimize there is nothing to diverge, and no
adaptation at all scores $14.4$ there, so NEO improves it and the late schedule improves it again.
SPA collapses nowhere and regresses nowhere, the only method in the table with a positive cell for
every corruption on both backbones.

\paragraph{What the two token budgets cost.}
\label{app:ttacost}
Appendix~\ref{app:latency} measures the schedule at inference. Under adaptation the accounting is
per method, since each one wraps the backbone differently. Every cell below is the same method at the
same learning rate; only the token budget changes. GFLOPs are counted with PyTorch's flop counter, so
a method's extra forward passes, its backward pass and its auxiliary modules are all included, and
every cell was identical across ten batches. Times are the mean over $100$ timed steps
after $20$ warmup steps, averaged over two repeated runs at batch~64 on one GPU, one process per cell,
with deterministic kernels disabled; the two runs differ by $0.06\%$ in the median cell and by at
most $1.0\%$.

The saving is the same for every method. GFLOPs fall by $26\%$ throughout, within $0.9$pp, because a
method's mechanism multiplies the reduced and the unreduced model alike and only the token profile
survives: SAR costs exactly twice Tent at either budget, SPA about two and a half times, FOA
thirteen to fourteen times, and all of them shed the same fraction. Wall-clock follows at $17.5$ to $24.7\%$. So the composability
in Table~\ref{tab:tta} is not bought back in compute: whichever method is layered on top, the late
schedule runs it at about three quarters of the cost of full tokens.

\begin{table}[ht]
\centering
\caption{Cost of one adaptation step on full tokens against the late schedule
  ($r_{\mathrm{base}}{=}8$, $\gamma{=}2$), ImageNet-C severity~5, batch~64. Each percentage is the
  fraction the quantity shrank. Full tokens = $197$ in every block.}
\label{tab:ttacost}
\small
\centering
\begin{tabular}{l r r r r r r}
\toprule
 & \multicolumn{3}{c}{GFLOPs / step} & \multicolumn{3}{c}{ms / step} \\
\cmidrule(lr){2-4} \cmidrule(lr){5-7}
Method & full & late & cut & full & late & cut \\
\midrule
\rowcolor{groupgrey} \multicolumn{7}{l}{\emph{ViT-S}} \\
Tent  & 1216 & 890 & $26.8\%$ & 156.9 & 124.5 & $20.7\%$ \\
EATA  & 1216 & 890 & $26.8\%$ & 171.5 & 139.2 & $18.8\%$ \\
SAR   & 2431 & 1781 & $26.8\%$ & 318.5 & 256.1 & $19.6\%$ \\
FOA   & 15555 & 11533 & $25.9\%$ & 1927.4 & 1590.8 & $17.5\%$ \\
SPA   & 3028 & 2226 & $26.5\%$ & 394.3 & 319.0 & $19.1\%$ \\
NEO   & 589 & 435 & $26.2\%$ & 67.5 & 54.2 & $19.8\%$ \\
\rowcolor{groupgrey} \multicolumn{7}{l}{\emph{ViT-B}} \\
Tent  & 4573 & 3359 & $26.5\%$ & 459.1 & 353.4 & $23.0\%$ \\
EATA  & 4573 & 3359 & $26.5\%$ & 524.4 & 394.7 & $24.7\%$ \\
SAR   & 9146 & 6718 & $26.5\%$ & 934.1 & 719.4 & $23.0\%$ \\
FOA   & 63943 & 47383 & $25.9\%$ & 6378.6 & 4984.7 & $21.9\%$ \\
SPA   & 11410 & 8401 & $26.4\%$ & 1155.8 & 892.0 & $22.8\%$ \\
NEO   & 2248 & 1658 & $26.2\%$ & 209.7 & 161.5 & $23.0\%$ \\
\bottomrule
\end{tabular}
\end{table}

\section{Reduction Distortion Is Front-Loaded for Every Reducer}
\label{app:mechgen}

The \emph{reduction distortion} $D(\ell)$, the output error a single reduction at layer $\ell$
produces, is the per-step quantity the schedule trades off. We measure it directly for each
training-free reducer: starting from the full $197$-token DeiT-S, at layer $\ell$ (and only there)
we let the reducer remove a fixed $16$ tokens and read the relative change in the output logits,
$D(\ell)=\mathbb{E}_x\,\lVert f^{(\ell)}(x)-f(x)\rVert/\lVert f(x)\rVert$ over $N{=}1024$ images
($f$ is the unreduced model). This reads the model output rather than operation-specific merged
features, so it applies uniformly to merging, pruning, sampling, and clustering reducers.

The main-text Figure~\ref{fig:mechtoken}a plots $D(\ell)$ under gaussian noise (severity~5) for all
five reducers, every curve front-loaded, an early reduction distorting far more than a late one.
This holds across the corruption taxonomy: Table~\ref{tab:mechgen} gives the early/late ratio per
reducer for one corruption from each ImageNet-C group (noise, blur, weather, digital), $\ge 6\times$
in every cell (Spearman against depth $\le -0.90$ throughout). Because the amplification $A(\ell)$
is operation-agnostic (Section~\ref{sec:mechtoken}), this front-loaded $D(\ell)$ is what makes late
concentration beneficial for every reducer, not only the merging case in the main text, generalizing
the account to the full set of reducers in Table~\ref{tab:baselines}.

\begin{table}[ht]
\centering
\caption{Front-loading of the per-layer reduction distortion $D(\ell)$ under each ImageNet-C group (DeiT-S,
  $N{=}1024$, severity~5): early/late ratio (early $=\ell\in\{1,\ldots,4\}$, late $=\ell\in\{9,\ldots,12\}$
  means) for one representative corruption per group (noise = gaussian, blur = defocus, weather = fog,
  digital = jpeg). Every cell is strongly front-loaded (ratio $\ge 6\times$; Spearman against depth
  $\le -0.90$ throughout).}
\label{tab:mechgen}
\begin{tabular*}{0.86\linewidth}{@{\extracolsep{\fill}}ll rrrr}
\toprule
Reducer & Family & noise & blur & weather & digital \\
\midrule
ToMe   & merging    & $31.5$ & $21.1$ & $35.2$ & $20.8$ \\
EViT   & pruning    & $13.4$ & $10.5$ & $15.8$ & $13.5$ \\
ATS    & sampling   & $10.9$ & $8.9$  & $9.5$  & $10.1$ \\
ATC    & clustering & $38.2$ & $22.6$ & $38.2$ & $21.6$ \\
PiToMe & merging    & $14.2$ & $6.6$  & $9.0$  & $10.4$ \\
\bottomrule
\end{tabular*}
\end{table}

\section{Per-Layer Probe: Summary Values}
\label{app:probe}

Table~\ref{tab:mechtoken} gives the exact per-condition summary behind Figure~\ref{fig:mechtoken}b
(the amplification curves of Section~\ref{sec:mechtoken}): early- and late-layer means of the
amplification $A(\ell)$, with the Spearman rank correlation against layer index over all $12$
layers. The strongly negative Spearman values ($\le -0.92$) confirm $A(\ell)$ is non-increasing in
$\ell$ (front-loaded). The full per-layer efficiency $\rho(\ell)=D(\ell)/S(\ell)$, from the
directly-measured per-layer reduction distortion $D(\ell)$ (Appendix~\ref{app:mechgen}) and the
marginal-saving denominator $S(\ell)$ of Appendix~\ref{app:sweetspot}, is reported in
Table~\ref{tab:rho}.

\paragraph{Why $A(\ell)$ is front-loaded (linearized view).}
\begin{lemma}[Linearized amplification]\label{lem:amp}
To first order, linearizing the layers above $\ell$ with input--output Jacobians $J_k$, a perturbation
$\delta$ at layer $\ell$'s output reaches the readout as $\big(\prod_{k>\ell}J_k\big)\delta$; its gain is
therefore at most the operator norm $\big\lVert\prod_{k>\ell}J_k\big\rVert$, attained only along the top
singular direction, and the isotropic probe measures the average-case gain
$A(\ell)\le\big\lVert\prod_{k>\ell}J_k\big\rVert$.
\end{lemma}
In words, an error introduced earlier passes through more layers on its way to the classifier, so it
has more opportunity to be amplified. Two consequences follow. The bound
$\big\lVert\prod_{k>\ell}J_k\big\rVert$ is a product of $L{-}\ell$ downstream factors, one fewer at
each deeper layer, so it is non-increasing in $\ell$ whenever those layers are on average
non-contractive, matching the measured front-loading of $A(\ell)$. And since
$\big\lVert\prod_{k>\ell}J_k\big\rVert\le\prod_{k>\ell}\lVert J_k\rVert$, a shift that scales the
per-layer norms by $(1{+}\epsilon)$ scales this bound by $(1{+}\epsilon)^{L-\ell}$, which compounds
over more factors at small $\ell$, inflating the \emph{early}-layer amplification most and leaving
the latest layers almost unchanged (the realized inflation also depends on singular-vector
alignment). This depth-compounding of a perturbation through a product of layer Jacobians is the
mechanism studied in mean-field analyses of deep
networks~\citep{poole2016exponential,pennington2017isometry}, here measured in a trained ViT. This
reduces the question of why shift raises $A(\ell)$ to the local, measurable claim that corruption
raises the downstream Jacobian norms, which Table~\ref{tab:mechtoken} confirms (early $A(\ell)$
inflates under corruption while late $A(\ell)$ barely moves).

\paragraph{Why the per-layer norms inflate off the clean distribution.}\label{app:ood-amp}
Self-attention can amplify off-manifold perturbations in a way a local operator cannot. Standard
dot-product self-attention is not Lipschitz~\citep{kim2021lipschitz}: its output token $o_i=\sum_j
p_{ij}\,v_j$ is a convex combination of value vectors with input-dependent weights
$p_i=\mathrm{softmax}(q_i K^\top/\sqrt{d})$, and a perturbation reroutes these weights through
$\partial p_i/\partial(\text{logits})=\mathrm{diag}(p_i)-p_i p_i^\top$ scaled by $K/\sqrt{d}$, a
term the unbounded bilinear form $q_i K^\top$ leaves unbounded. The coupling is all-to-all, so a
change to one token can shift every token's weights, unlike the fixed, bounded Jacobian of a
convolution's local receptive field. On the clean manifold this rerouting is organized by training;
off it, where the network was never optimized, it is unconstrained, consistent with the early-layer
inflation in Table~\ref{tab:mechtoken}.

We stress the scope. This is a capacity argument: it explains why a transformer can amplify
off-manifold shifts, not that a given shift must, which component (attention, LayerNorm, or the MLP)
dominates, or a signed magnitude per corruption; those the probe measures. The theory supplies only
the structural reason amplification is available, while the depth profile (early-most) follows from
Lemma~\ref{lem:amp} under the on-average non-contractive condition and is confirmed empirically
(Table~\ref{tab:mechtoken}).

\begin{table}[ht]
\centering
\caption{Amplification $A(\ell)$ summary (DeiT-S, $N{=}1024$, compute-neutral), for clean inputs
  and one corruption from each ImageNet-C group (severity~5). $A(\ell)$ is front-loaded (early $\gg$ late)
  in every condition, with clean the flattest, so the network's amplification concentrates a reduction's
  distortion in the early layers. early $=\ell\in\{1,\ldots,4\}$, late $=\ell\in\{9,\ldots,12\}$ (means); Spearman is
  over all 12 layers.}
\label{tab:mechtoken}
\centering
\begin{tabular}{lrrr}
\toprule
Condition & early $A(\ell)$ & late $A(\ell)$ & Spearman \\
\midrule
clean                       & 0.60 & 0.25 & $-0.92$ \\
gaussian noise (noise)      & 0.92 & 0.20 & $-1.00$ \\
defocus blur (blur)         & 0.74 & 0.20 & $-1.00$ \\
fog (weather)               & 0.83 & 0.17 & $-1.00$ \\
jpeg compression (digital)  & 0.77 & 0.23 & $-1.00$ \\
\bottomrule
\end{tabular}
\end{table}

\section{Clean Merge-Path Control: Is the Matching Criterion the Lever?}
\label{app:cleanpath}

We test whether the OOD token-reduction penalty is driven by degraded merge \emph{decisions} (which
tokens are merged) rather than by the schedule. For each ImageNet-C image we capture, layer by
layer, the merge path (the token groupings) that ToMe takes on the image's \emph{clean} counterpart,
and force the corrupted image to follow that same ``oracle'' path, ignoring its own corrupted
features when choosing what to merge. The clean path is unavailable at test time (it requires the
clean image); it upper-bounds what restoring the uncorrupted matching could recover (a
clean-decision oracle, not a proven global optimum). We compare, at iso-GFLOPs, against the standard
own-path baseline and against our late schedule. DeiT-S, the 15-corruption ImageNet-C mean
(severity~5), full $50$k; the own-path column reproduces the canonical flat numbers exactly, a
correctness check on the replay. The uncompressed model scores $39.40$.

\begin{table}[ht]
\centering
\caption{Clean merge-path control (15-corruption ImageNet-C mean, severity~5, DeiT-S, iso-GFLOPs). Forcing
  the clean (oracle) merge decisions recovers some OOD accuracy, but less than the late schedule at every
  $r_{\mathrm{base}}$. Parentheses give the gain over the own-path
  baseline; the late schedule uses its own (corrupted) decisions and needs no clean reference.}
\label{tab:cleanpath}
\begin{tabular}{lrrr}
\toprule
$r_{\mathrm{base}}$ & own-path (flat) & oracle clean-decision & late schedule ($\gamma{=}2$) \\
\midrule
$4$  & $38.83$ & $39.22$ $(+0.39)$ & $\mathbf{39.37}$ $(+0.54)$ \\
$6$  & $38.44$ & $39.03$ $(+0.59)$ & $\mathbf{39.32}$ $(+0.88)$ \\
$8$  & $37.99$ & $38.79$ $(+0.80)$ & $\mathbf{39.16}$ $(+1.17)$ \\
$10$ & $37.41$ & $38.36$ $(+0.95)$ & $\mathbf{38.66}$ $(+1.25)$ \\
\bottomrule
\end{tabular}
\end{table}

Even an oracle that uses the uncorrupted image's merge decisions recovers less than the late
schedule, which uses its own decisions and needs no clean reference. Relative to this clean-decision
oracle, the schedule, not the matching criterion, is the stronger lever, which motivates holding the
matching rule fixed while studying allocation. A more corruption-robust matching criterion is a
separate design axis that we do not pursue here, and leave to future work
(Section~\ref{sec:limitations}).

\section{Robustness to the Schedule's Functional Form}
\label{app:shapeform}

The power law (Eq.~\ref{eq:schedule}) is a convenient single-parameter probe of the
late-concentrated axis, not a shape we claim is optimal (Section~\ref{sec:model}). To check that the
gain reflects the late-concentrated \emph{direction} rather than the power-law form, we replace the
power law with two other monotone late-weighted families at the main operating point
($r_{\mathrm{base}}{=}8$, DeiT-S, ToMe, ImageNet-C severity~5), each calibrated to the same
iso-GFLOPs rule (at or below the flat baseline's GFLOPs) and with its shape parameter set so that
its removal center-of-mass matches the $\gamma{=}2$ schedule's: an \emph{exponential} profile
$r_\ell\!\propto\!e^{\beta \ell/L}$ ($\beta{=}3$; per-layer removals
$[2,3,4,6,7,9,12,16,21,28,36,48]$), and a non-smooth \emph{two-level step} that removes about $2$
tokens per layer in the first half and about $30$ per layer in the second half
($[2,2,2,2,2,2,30,30,30,30,30,30]$).

Both recover gains on par with the power law (Table~\ref{tab:shapeform}): the exponential adds
$+1.05$pp and the step $+1.18$pp over flat, against the power law's $+1.17$pp, and all 15
corruptions improve under each shape (Table~\ref{tab:shapeform-percorr}). The benefit is thus a
property of the late-concentrated direction, not of the power-law parameterization; we keep the
power law because its single scalar $\gamma$ admits the analytic iso-GFLOPs calibration and the
continuous concentration sweep used in the mechanism analysis (Figure~\ref{fig:calib}c), not because
it is the most accurate shape. The small spread across shapes is itself consistent with the
front-loaded-error account (Section~\ref{sec:mechtoken}): the step, which defers early removal most
($\sim$2 tokens per layer in the first half), matches the power law, while the exponential, which
removes slightly more in the earliest layers, gives back a little. This is a check at one operating
point on one backbone, not an exhaustive search over schedule families.

\begin{table}[ht]
\centering
\caption{Late-concentration is robust to the schedule's functional form (DeiT-S, ToMe, ImageNet-C
  sev=5, 15-corruption mean top-1, \%; $r_{\mathrm{base}}{=}8$, inference-only, iso-GFLOPs $\le$ flat).
  Each alternative shape is center-of-mass-matched to the $\gamma{=}2$ schedule. $\Delta$ = shape $-$ flat.}
\label{tab:shapeform}
\footnotesize
\begin{tabular}{llrrr}
\toprule
Schedule & Family & GFLOPs & mean & $\Delta$ vs.\ flat \\
\midrule
flat (const $r_{\mathrm{base}}{=}8$) & -- & $3.199$ & $37.99$ & -- \\
power law ($\gamma{=}2$) & power law (convex) & $3.147$ & $39.16$ & $+1.17$ \\
exponential ($\beta{=}3$) & exponential (convex) & $3.194$ & $39.04$ & $+1.05$ \\
two-level step & step (non-smooth) & $3.178$ & $39.17$ & $+1.18$ \\
\bottomrule
\end{tabular}
\end{table}

\begin{table}[ht]
\centering
\caption{Per-corruption top-1 (\%) for the alternative late shapes (DeiT-S, ToMe, sev=5,
  $r_{\mathrm{base}}{=}8$, iso-GFLOPs). uncompressed, flat and the power law ($\gamma{=}2$) are the same runs as
  Figure~\ref{fig:percorruption}. All 15 corruptions improve over flat under every late shape.}
\label{tab:shapeform-percorr}
\resizebox{\linewidth}{!}{%
\begin{tabular}{l rrrrrrrrrrrrrrr r}
\toprule
 & gauss & shot & imp & defoc & glass & motn & zoom & snow & frost & fog & brt & cont & elas & pix & jpeg & mean \\
\midrule
uncompressed    & 33.89 & 33.79 & 34.92 & 27.73 & 18.92 & 32.23 & 28.41 & 47.53 & 53.16 & 54.22 & 70.25 & 38.55 & 34.75 & 29.87 & 52.72 & 39.40 \\
\midrule
flat            & 31.62 & 31.70 & 32.40 & 26.53 & 18.42 & 31.11 & 27.31 & 46.37 & 51.10 & 51.43 & 69.57 & 36.59 & 34.58 & 28.88 & 52.19 & 37.99 \\
power law ($\gamma{=}2$) & 33.79 & 33.58 & 34.85 & 27.57 & 18.72 & 32.24 & 28.75 & 47.62 & 52.45 & 52.97 & 70.18 & 37.31 & 34.79 & 29.74 & 52.82 & 39.16 \\
exponential     & 33.43 & 33.28 & 34.53 & 27.42 & 18.72 & 32.11 & 28.36 & 47.49 & 52.42 & 52.85 & 70.05 & 37.43 & 35.07 & 29.63 & 52.77 & 39.04 \\
two-level step  & 33.58 & 33.45 & 34.59 & 27.61 & 18.73 & 32.20 & 28.67 & 47.71 & 52.53 & 52.94 & 70.09 & 37.87 & 35.13 & 29.72 & 52.79 & 39.17 \\
\bottomrule
\end{tabular}%
}
\end{table}

\section{A Searched Schedule Is Also Late-Concentrated}
\label{app:learned}

DiffRate~\citep{chen2023diffrate} searches a per-layer compression rate for a given architecture and
FLOP target by differentiable optimization on labeled ImageNet, and is therefore the learned
counterpart of the schedule Eq.~\ref{eq:schedule} specifies in closed form;
LTMP~\citep{bonnaerens2023ltmp} learns per-layer thresholds whose removal varies per input, a
different object. DiffRate reports its searched per-layer rates and shows that they outperform
constant schedules on clean ImageNet; it does not characterize the depth profile of those rates or
evaluate under distribution shift. Section~\ref{sec:mechanism} predicts what an accuracy-maximizing
search at fixed compute should find, namely removal concentrated in late layers, heaviest at the
last layer where a removal costs almost no accuracy (Appendix~\ref{app:finalblock}), yet spread over
several layers by per-layer capacity.

We ran DiffRate's released search with its default recipe (full ImageNet, three epochs) at our main
operating point for the two backbones its code supports. Table~\ref{tab:learned} lists the searched
per-layer removal next to the $\gamma{=}2$ power law at the same budget. Both searched schedules
have the predicted shape. The last three layers take $57\%$ of all removal on DeiT-S and $69\%$ on
DeiT-B, against $56\%$ for the power law, and the last layer alone takes $38\%$ and $29\%$, against
$22\%$, so the searched schedules are late-concentrated to a similar degree and heavier still at the
final layer. For accuracy we also evaluate a flat schedule at the same budget; all three run in
DiffRate's own prune-and-merge implementation, so their numbers are comparable with each other but
not with Table~\ref{tab:main}. At matched THOP GFLOPs on ImageNet-C (severity~5, 15-corruption mean;
within $0.4\%$ on DeiT-S, and with $4.8\%$ less compute for the power law on DeiT-B), the fixed
$\gamma{=}2$ schedule reaches $36.30$ on DeiT-S against $36.82$ for the searched schedule and
$34.33$ for flat, and $43.27$ on DeiT-B against $43.59$ and $41.17$. The one-parameter formula thus
delivers most of the gain that a labeled, gradient-based search obtains over flat, with no search,
labels, or gradients.

The searched schedule is also tied to the setting it was searched for. It comes from a
differentiable search that needs the labeled training set and gradient access to the model, is run
separately for each architecture, token count, and FLOP budget ($3.4$ and $7.2$ GPU-hours for DeiT-S
and DeiT-B with the default recipe, paid again for every budget), and yields rates specific to
DiffRate's own prune-and-merge operator; the released schedules cover the DeiT and MAE families at
$197$ tokens and a few budgets. None exists for the ViT-S/B backbones, the video model ($1568$
space-time tokens), the vision-language encoder ($576$ tokens), or the four further reducers of this
paper. The closed-form schedule needs none of this. Given only the layer count and the budget, it is
evaluated in under a millisecond on a CPU and applied unchanged in every setting above.

\begin{table}[ht]
\centering
\caption{Per-layer tokens removed at the main operating point (DeiT, $197$ tokens). DiffRate's searched
  schedules, from a labeled search run per architecture and budget, and the closed-form $\gamma{=}2$
  schedule; all three concentrate removal in the last layers, the searched ones most at the final layer, as
  Section~\ref{sec:mechanism} predicts.}
\label{tab:learned}
\footnotesize
\begin{tabular}{lrrrrrrrrrrrrrr}
\toprule
layer & 1 & 2 & 3 & 4 & 5 & 6 & 7 & 8 & 9 & 10 & 11 & 12 & total & last 3 \\
\midrule
DeiT-S, searched & 0 & 2 & 6 & 12 & 11 & 12 & 10 & 12 & 18 & 5 & 31 & 72 & 191 & 57\% \\
DeiT-B, searched & 0 & 1 & 2 & 2 & 7 & 8 & 11 & 11 & 17 & 49 & 29 & 55 & 192 & 69\% \\
power law, $\gamma{=}2$ (both) & 0 & 1 & 3 & 5 & 7 & 11 & 14 & 19 & 24 & 30 & 36 & 42 & 192 & 56\% \\
\bottomrule
\end{tabular}
\end{table}

\section{Schedule Placement at the Aggressive Rate}
\label{app:placer12}

\begin{wrapfigure}{r}{0.36\linewidth}
  \centering
  \includegraphics[width=\linewidth]{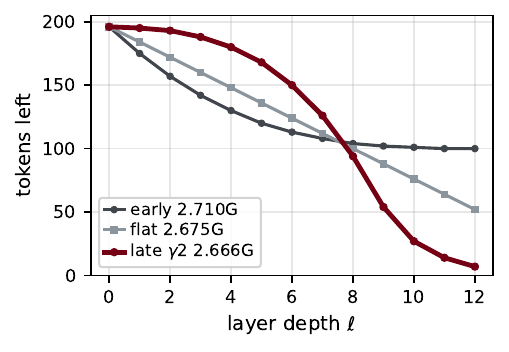}
  \vspace{-25pt}
  \caption{Tokens remaining after each layer at $r_{\mathrm{base}}{=}12$ (DeiT-S), for the
    early-concentrated, flat and late-concentrated schedules; the legend gives each schedule's GFLOPs.}
  \label{fig:placer12}
  \vspace{-25pt}
\end{wrapfigure}

Figure~\ref{fig:placer12} gives the token profiles at $r_{\mathrm{base}}{=}12$ that accompany the
$r_{\mathrm{base}}{=}8$ panel of Figure~\ref{fig:placement}. The ordering is the same at both rates:
early-concentrated front-loads its removals and still uses the most compute ($2.710$G versus flat's
$2.675$G and late's $2.666$G), yet it loses to flat by $2.23$pp, while the late schedule uses the
least compute and wins by $1.10$pp.

The early-concentrated control is the same power law reflected, so its removal rate falls with
depth, and we calibrate its per-layer budget against flat's compute as we do for the late schedules.
Because a removal made early frees capacity in every later block, matching flat's cost forces a much
smaller total removal: $60$ tokens versus flat's $96$ at $r_{\mathrm{base}}{=}8$, and $96$ versus
$144$ here. 
The calibrated control also ends up slightly above flat rather than below it, $3.290$G
versus $3.199$G at $r_{\mathrm{base}}{=}8$ and $2.710$G versus $2.675$G here, so the placement
comparison runs against our own claim on both the compute and the token axis.

\section{Why the Optimum is Interior: an iso-GFLOPs Account of the Sweet Spot}
\label{app:sweetspot}

At a fixed token budget $B=\sum_\ell r_\ell$, the front-loaded per-step distortion
(Section~\ref{sec:model}) is minimized by pushing all removals to the latest layers. Taken alone
this would predict ever-more-extreme concentration. The experiments instead show an inverted-U in
$\gamma$ (Table~\ref{tab:main}, Table~\ref{tab:calib-appendix}). The resolution is that our
comparisons are at \emph{iso-GFLOPs}, under which the budget $B$ is not a free constant.

\paragraph{The iso-GFLOPs constraint.} Because ToMe merges between a layer's attention and its MLP, a
removal at layer $k$ shrinks the token stream seen by that layer's own MLP \emph{and} by every
subsequent layer. To first order the per-token cost of a layer is constant, so the marginal compute
saved by one removal at layer $k$ is
\begin{equation}
  S(k) \;=\; c \;+\; (L-k),
  \label{eq:saved}
\end{equation}
in units of a full layer, where $(L-k)$ counts the downstream layers and the constant
$c=c_{\mathrm{MLP}}/c_{\mathrm{layer}}\approx0.61$ is the layer's own MLP saving (DeiT-S: MLP $8d^2$
MACs per token over a full layer's $12d^2+2Nd$, with $d{=}384$, $N{=}197$). Crucially $S(L)=c>0$: a
removal at the last layer still saves that layer's MLP. (The self-attention term, quadratic in the
token count, only sharpens the depth dependence: early removals shrink the sequence where attention
is most expensive.) The same form holds for any reducer that removes tokens inside a block, with $c$
the fraction of that block's per-token cost downstream of the removal point; the account uses the
shape of $S(k)$, not the value of $c$. The total compute saved is $C_{\mathrm{saved}}=\sum_k
r_k\,S(k)$, and matching a flat baseline's GFLOPs fixes $\sum_k r_k\,S(k)=S_{\mathrm{tot}}$,
\emph{not} $\sum_k r_k$.

\paragraph{Optimal allocation is governed by distortion-per-compute.} Writing $D(k)$ for the per-layer
reduction distortion (Section~\ref{sec:model}), minimizing the total distortion $\sum_k r_k\,D(k)$
subject to $\sum_k r_k\,S(k)=S_{\mathrm{tot}}$ and $0\le r_k\le \kappa_k$ (per-layer capacity) is,
as a first-order diagnostic relaxation (not an optimization of the actual network), a linear program
whose solution fills layers in increasing order of the ratio
\begin{equation}
  \rho(k) \;=\; \frac{D(k)}{S(k)},
  \label{eq:rho}
\end{equation}
the distortion incurred per unit of compute saved, until the constraint $S_{\mathrm{tot}}$ is met (a
water-filling solution). We evaluate $D(k)$ with its directly-measured values
(Appendix~\ref{app:mechgen}); Table~\ref{tab:rho} reports the resulting $\rho$. This ratio falls
with depth (later reductions are cheaper), and at the last layer it \emph{vanishes} ($D\approx0$,
i.e.\ $A(L)=0$): the readout sees only the class token, so a perturbation to the final layer's patch
tokens never reaches it (the class token is never merged), while $S(L)=c>0$. The last layer
therefore reduces at near-zero distortion, $\rho(L)\to0$, and unconstrained the water-filling
solution would pour the entire budget into the latest layers. What makes the optimum interior is the
per-layer capacity $\kappa_k$: only so many mutually similar tokens exist to merge at any one layer,
so once the cheap late layers are filled the budget must spill into earlier, higher-$\rho$ layers.
The optimal schedule is thus late-concentrated but not maximally so. This is the opposite of a
spurious blow-up at the last layer: the endpoint is the cheapest place to reduce, and it is
capacity, not efficiency, that bounds the concentration.

\paragraph{The efficient region is late, in clean and corrupted inputs alike.}
Table~\ref{tab:rho} evaluates $\rho(\ell)$ with the corrected denominator $S(\ell)$ from the
directly-measured per-layer reduction distortion $D(\ell)$ (the probe of Appendix~\ref{app:mechgen};
DeiT-S, $N{=}1024$, $L{=}12$). Layer $L$ is the degenerate free reduction ($D\approx0$, confirming
$A(L){=}0$, $\rho(L){\approx}0$). Among the capacity-bounded interior layers the efficient region is
late in both conditions ($\arg\min_{\ell<L}\rho$ at $\ell{=}11$): the distortion is steeply
front-loaded already on clean ($\rho$ at $\ell{=}1$ about $11\times$ that at $\ell{=}11$), so late
removal is the efficient choice with or without corruption. Corruption sharpens this rather than
relocating it: the early-layer distortion inflates (early $D$ rising about $40\%$ from clean to
gaussian noise while the late layers stay low), raising $\rho$ at $\ell{=}1$ to about $13\times$ the
late value. The efficiency diagnostic thus identifies \emph{where} the budget is efficient (the
late, capacity-bounded layers, in every condition); the growth of the \emph{gain} with severity is
established separately by the gains themselves (Section~\ref{sec:severity}).

\begin{table}[ht]
\centering
\caption{Distortion per unit of compute saved, $\rho(\ell)=D(\ell)/S(\ell)$ ($\times 10^{3}$), from the
  directly-measured per-layer reduction distortion $D(\ell)$ (relative output-logit change from removing tokens
  at layer~$\ell$ only; DeiT-S, $N{=}1024$; Appendix~\ref{app:mechgen}) and the exact
  marginal saving $S(\ell)=(L-\ell)+c$, $c{=}0.61$ (the layer's own-MLP fraction, since ToMe merges before the
  MLP). The distortion is steeply front-loaded in both conditions and steepens under noise. Layer $L{=}12$ is
  the degenerate free reduction: $D\approx0$ confirms $A(L){=}0$, giving $\rho(L){\approx}0$ despite
  $S(L){=}c{>}0$. Among the interior layers the minimum (\textbf{bold}) is late ($\ell{=}11$) in both
  conditions; it is per-layer capacity, not a blow-up of $\rho$, that keeps the optimum off the free last
  layer. The table is a directional diagnostic, showing the cheap-to-reduce region sits in the late (but
  capacity-bounded) layers, not a predictor of the exact optimal $\gamma$.}
\label{tab:rho}
\begin{tabular}{rrr}
\toprule
layer $\ell$ & clean & gaussian \\
\midrule
1  & 16.4 & 25.8 \\
2  & 12.7 & 18.4 \\
3  & 8.3  & 10.9 \\
4  & 6.3  & 5.9 \\
5  & 4.1  & 5.7 \\
6  & 3.4  & 4.7 \\
7  & 2.8  & 3.1 \\
8  & 2.4  & 3.1 \\
9  & 2.7  & 2.8 \\
10 & 2.3  & 2.9 \\
11 & \textbf{1.5}  & \textbf{2.0} \\
12 & 0.0  & 0.0 \\
\bottomrule
\end{tabular}
\end{table}

\paragraph{Capacity and saturation.} Two features keep the optimum interior, at different levels. The
hard capacity $\kappa_k$ in the linear program above already prevents collapse onto the free late
layers (the water-filling spill noted above), but with the per-unit distortion held fixed it would
fill each layer bang-bang to its cap. A refinement of that fixed-distortion assumption smooths the
solution. Treating the per-layer distortion $D(k)$ as independent of $r_k$ is first-order; in
practice a layer has a limited supply of mutually similar tokens, so as $r_k$ grows toward
$\kappa_k$, bipartite matching is forced onto less-similar pairs and the marginal distortion rises,
i.e.\ $D(k)$ increases (and becomes convex) in $r_k$. We treat this as a modeling assumption rather
than a directly measured quantity. This convexity penalizes dumping the entire budget into a few
late layers (exactly what large $\gamma$ does, where the removal budget balloons: its per-layer
average rises from $12$ at $\gamma{=}0$ to $44$ at $\gamma{=}3$ for $r_{\mathrm{base}}{=}12$), and
favors spreading removal smoothly across the efficient late-but-interior band, i.e.\ a moderate
power law.

\paragraph{Dependence on $r_{\mathrm{base}}$.} A larger overall budget (higher $r_{\mathrm{base}}$, hence
larger $S_{\mathrm{tot}}$) reaches the capacity limits of the most efficient late layers sooner,
forcing some removal into earlier, higher-$\rho$ layers; the optimal concentration $\gamma$ thus
\emph{decreases} with $r_{\mathrm{base}}$. Conversely a smaller budget leaves more slack to
concentrate, so the optimal $\gamma$ increases. This matches our measurements: on the 15-corruption
mean the optimum migrates from $\gamma\!\approx\!3$ at the lightest $r_{\mathrm{base}}{=}2$ to
$\gamma\!\approx\!1.5$ at $r_{\mathrm{base}}{=}12$ (Table~\ref{tab:main}), and on gaussian noise
(Table~\ref{tab:calib-appendix}) over-concentration (pushing $\gamma$ past the optimum) already
hurts at the most aggressive setting we report ($r_{\mathrm{base}}{=}14$, where $\gamma{=}1.5$ falls
below $\gamma{=}1.0$). The account is first-order; it predicts the \emph{shape} (an interior optimum
that shifts with budget), which is what we observe, not a precise $\gamma^\star$.

\section{The Late Gain is Not a Final-Layer or Readout Artifact}
\label{app:finalblock}

ToMe inserts its merge between a layer's attention and its MLP and protects the class token. At the
final layer the attention has already run, so the class token has finished aggregating patch
information; merging patch tokens afterward passes only through the token-wise MLP and LayerNorm,
with no further patch-to-class attention. A final-layer patch merge therefore cannot change the
logits (Section~\ref{sec:model}: $A(L)=0$), while still saving that layer's MLP on the discarded
patch tokens. Because the late-concentrated schedule places its largest removals at the deepest
layers, one might worry that part of its iso-GFLOPs advantage comes for free from this degenerate
reduction, or more generally from the class-token readout. We rule this out with three controls.

\paragraph{The final layer is a structurally free reduction.}
Merging tokens at only the last layer leaves the logits \emph{bit-identical} to the unreduced model
(max $|\Delta\mathrm{logit}|=0$ over $300$ gaussian-noise images at severity~5, for every merge
count up to the per-layer clamp; a $1.9\times10^{-6}$ rounding residue appears only at an extreme
$r{=}40$). The same merge at layer~$1$ changes $29\%$ of predictions (max
$|\Delta\mathrm{logit}|=5.5$), and at layer~$10$ it is already tiny ($0.076$). This confirms
$A(L)=0$ is structural (no patch-to-class path remains), not merely small. The strictly-free
final-layer saving is itself small (only that layer's MLP on the discarded patches, $0.2$--$0.7\%$
of total GFLOPs, versus the full downstream stack that early removals save); more to the point, the
late gain survives after this endpoint is removed entirely (next paragraph).

\paragraph{Excluding the final layer(s) leaves the gain intact.}
We drop the last layer (and the last two) from \emph{both} schedules and re-evaluate on the full
ImageNet-C validation set at the same operating point as the main comparison (DeiT-S, flat const
$r_{\mathrm{base}}{=}8$ and late $\gamma{=}2$, each recalibrated over the remaining layers so that
flat stays at its baseline compute and late at $\le$ flat; $15$ corruptions, severity~5, $50{,}000$
images each; Table~\ref{tab:exclast}). The late-over-flat gain is essentially unchanged: $+1.17$
(full, reproducing the main result) versus $+1.15$ (exclude last) and $+1.09$ (exclude last two),
positive on all $15$ corruptions in every condition. The advantage thus comes from the front-loaded
amplification across the interior layers, not from the free final reduction. This matches the
distortion account, in which the last layer contributes $r_L\,D(L)=0$ to the direct error (since
$A(L)=0$), so only the small compute it frees could matter, and the recalibrated control absorbs
that.

\paragraph{The effect reproduces under global-average-pooling readout.}
The free-final-layer phenomenon is specific to class-token readout. Under global average pooling
(GAP) the final layer's patch tokens are read out, so $A(L)\neq0$ and no free reduction exists
(merging at only the last layer now shifts the logits, max $|\Delta\mathrm{logit}|=0.13$). We
evaluate two MAE-pretrained models with GAP readout, ViT-B and ViT-L, the setting ToMe itself
studied, reusing ToMe's size-proportional pooling. At a matched $\sim25\%$ GFLOPs reduction (ViT-B
$r_{\mathrm{base}}{=}8$; ViT-L $r_{\mathrm{base}}{=}4$ over its $24$ layers; $\gamma{=}2$ late
calibrated to the closest schedule at or below each model's flat GFLOPs), the late schedule beats
flat on all $15$ corruptions for both models, by $+1.28$ (ViT-B) and $+1.22$ (ViT-L) mean
(Table~\ref{tab:maegap}, which also lists the compute and token budgets), reproducing the
class-token DeiT-S result ($+1.17$) in direction and magnitude. The schedule effect is therefore not
specific to the class-token readout.

\begin{table}[ht]
\centering
\caption{Excluding the final layer(s) from both schedules (DeiT-S, full ImageNet-C, severity~5, $50$k
  images; flat const $r_{\mathrm{base}}{=}8$ at $3.20$\,G, late $\gamma{=}2$ at $3.15$\,G, late $\le$
  flat). flat/late columns are top-1 (\%) for the full schedule; $\Delta$ = late $-$ flat for the full,
  exclude-last, and exclude-last-two variants, each recalibrated to the same compute. The full column
  reproduces Table~\ref{tab:severity}; the gain persists when the (free) final layer(s) are removed.}
\label{tab:exclast}
\begin{tabular}{lrrrrr}
\toprule
corruption & flat & late & $\Delta$ full & $\Delta$ excl-1 & $\Delta$ excl-2 \\
\midrule
gaussian noise    & 31.62 & 33.79 & +2.17 & +2.16 & +1.89 \\
shot noise        & 31.70 & 33.58 & +1.88 & +1.92 & +1.70 \\
impulse noise     & 32.40 & 34.85 & +2.45 & +2.33 & +2.21 \\
defocus blur      & 26.53 & 27.57 & +1.04 & +1.11 & +1.13 \\
glass blur        & 18.42 & 18.72 & +0.30 & +0.27 & +0.29 \\
motion blur       & 31.11 & 32.24 & +1.13 & +1.16 & +0.98 \\
zoom blur         & 27.31 & 28.75 & +1.44 & +1.42 & +1.12 \\
snow              & 46.37 & 47.62 & +1.25 & +1.15 & +1.12 \\
frost             & 51.10 & 52.45 & +1.36 & +1.19 & +1.16 \\
fog               & 51.43 & 52.97 & +1.54 & +1.54 & +1.46 \\
brightness        & 69.57 & 70.18 & +0.61 & +0.58 & +0.53 \\
contrast          & 36.59 & 37.32 & +0.73 & +0.78 & +1.16 \\
elastic transform & 34.58 & 34.79 & +0.21 & +0.24 & +0.26 \\
pixelate          & 28.88 & 29.74 & +0.86 & +0.86 & +0.69 \\
jpeg compression  & 52.19 & 52.82 & +0.64 & +0.53 & +0.60 \\
\midrule
\textbf{mean}     & \textbf{37.99} & \textbf{39.16} & \textbf{+1.17} & \textbf{+1.15} & \textbf{+1.09} \\
\bottomrule
\end{tabular}
\end{table}

\begin{table}[ht]
\centering
\caption{Global-average-pooling readout (MAE ViT-B and ViT-L, full ImageNet-C, severity~5, $50$k images).
  Each model's flat / late columns carry, in the upper rows, its budget (GFLOPs, with the reduction from the uncompressed model
  in parentheses, $16.9$G for ViT-B and $59.7$G for ViT-L; tokens removed and
  final, out of $196$ patches) and, in the lower rows, its per-corruption top-1 (\%); late is
  $\gamma{=}2$ calibrated to the closest schedule at or below the flat GFLOPs. At iso-GFLOPs
  (late $\le$ flat) late removes more tokens in total yet keeps fewer at the end, and still beats flat on
  all $15$ corruptions for both scales (mean $+1.28$ / $+1.22$): the gain is not from a larger token
  budget and reproduces the class-token result.}
\label{tab:maegap}
\resizebox{\linewidth}{!}{%
\begin{tabular}{lrrrrrr}
\toprule
 & \multicolumn{3}{c}{MAE ViT-B} & \multicolumn{3}{c}{MAE ViT-L} \\
\cmidrule(lr){2-4}\cmidrule(lr){5-7}
 & flat & late & $\Delta$ & flat & late & $\Delta$ \\
\midrule
\multicolumn{7}{@{}l}{\emph{Budget}} \\
\quad GFLOPs         & 12.67 ($-25\%$) & 12.62 ($-25\%$) & -- & 44.98 ($-25\%$) & 44.90 ($-25\%$) & -- \\
\quad tokens removed & 96    & 169   & -- & 96    & 180   & -- \\
\quad final tokens   & 100   & 27    & -- & 100   & 16    & -- \\
\midrule
\multicolumn{7}{@{}l}{\emph{Top-1 accuracy (\%) by corruption}} \\
gaussian noise    & 40.64 & 44.50 & +3.86 & 54.14 & 57.08 & +2.94 \\
shot noise        & 41.62 & 44.57 & +2.95 & 56.91 & 59.37 & +2.46 \\
impulse noise     & 42.13 & 45.88 & +3.75 & 56.47 & 59.24 & +2.77 \\
defocus blur      & 27.74 & 29.15 & +1.41 & 36.75 & 37.57 & +0.82 \\
glass blur        & 18.38 & 18.42 & +0.05 & 24.42 & 24.78 & +0.36 \\
motion blur       & 35.38 & 36.23 & +0.85 & 50.83 & 52.00 & +1.16 \\
zoom blur         & 29.24 & 30.27 & +1.02 & 43.32 & 44.16 & +0.84 \\
snow              & 53.04 & 53.92 & +0.87 & 64.19 & 65.21 & +1.02 \\
frost             & 54.47 & 55.12 & +0.65 & 65.04 & 65.84 & +0.80 \\
fog               & 56.54 & 57.89 & +1.36 & 60.48 & 63.22 & +2.74 \\
brightness        & 73.55 & 73.88 & +0.32 & 78.18 & 78.35 & +0.17 \\
contrast          & 46.69 & 46.96 & +0.26 & 54.58 & 54.61 & +0.03 \\
elastic transform & 26.19 & 26.30 & +0.10 & 38.96 & 39.48 & +0.53 \\
pixelate          & 43.61 & 45.09 & +1.48 & 49.98 & 51.34 & +1.35 \\
jpeg compression  & 52.93 & 53.17 & +0.24 & 60.29 & 60.55 & +0.25 \\
\midrule
\textbf{mean}     & \textbf{42.81} & \textbf{44.09} & \textbf{+1.28} & \textbf{52.97} & \textbf{54.19} & \textbf{+1.22} \\
\bottomrule
\end{tabular}%
}
\end{table}

\section{What the Late Schedule Recovers: a Two-Channel Decomposition}
\label{app:channels}

The body (Table~\ref{tab:main}) reports that late concentration closes $83\%$ of the accuracy the
flat schedule loses to compression. That is an aggregate accounting: in principle a schedule could
close the same gap by getting a \emph{different} set of images right. This appendix shows it does
not.

\paragraph{Setup.} We split every image-corruption case by what the same model does with no reduction
at all. In region~$A$ the unreduced model is correct, so any error there was manufactured by the
reduction and an advantage in $A$ is literally less reduction damage. In region~$B$ the unreduced
model is already wrong, so an advantage there is the schedule arriving at a new correct answer,
which is the ``it got lucky'' explanation. The two regions partition the flat-to-late accuracy
difference exactly,
\begin{equation*}
\underbrace{\Pr(A)\big[\mathrm{acc}_{\text{late}}(A)-\mathrm{acc}_{\text{flat}}(A)\big]}_{\text{recovery}}
\;+\;
\underbrace{\Pr(B)\big[\mathrm{acc}_{\text{late}}(B)-\mathrm{acc}_{\text{flat}}(B)\big]}_{\text{new correctness}}
\;=\;\mathrm{acc}_{\text{late}}-\mathrm{acc}_{\text{flat}},
\end{equation*}
so no part of the measured difference is left unattributed. All pairs are the calibrated iso-GFLOPs
schedules of Table~\ref{tab:isocalib} (late GFLOPs $\le$ flat), full ImageNet-C at severity~5,
$50{,}000$ images $\times$ $15$ corruptions $=750{,}000$ cases per backbone.

\begin{table}[ht]
\centering
\caption{Two-channel decomposition of the late-minus-flat accuracy difference (pp), full ImageNet-C
  severity~5. \emph{recovery} = cases the unreduced model already got right; \emph{new correctness} =
  cases it got wrong. In all twelve cells the recovery channel is positive and larger than the net,
  and the new-correctness channel is negative: the gain comes from not destroying predictions the
  model already had, not from finding new ones.}
\label{tab:channels}
\small\setlength{\tabcolsep}{6pt}
\begin{tabular}{ll rrr}
\toprule
backbone & pair ($r_{\mathrm{base}}$, $\gamma$) & net & recovery & new correctness \\
\midrule
DeiT-S & $4$, $\gamma{=}1$  & $+0.375$ & $\mathbf{+0.740}$ & $-0.366$ \\
DeiT-S & $4$, $\gamma{=}2$  & $+0.542$ & $\mathbf{+1.169}$ & $-0.627$ \\
DeiT-S & $8$, $\gamma{=}1$  & $+0.902$ & $\mathbf{+1.257}$ & $-0.355$ \\
DeiT-S & $8$, $\gamma{=}2$  & $+1.173$ & $\mathbf{+1.798}$ & $-0.625$ \\
DeiT-S & $12$, $\gamma{=}1$ & $+1.213$ & $\mathbf{+1.417}$ & $-0.204$ \\
DeiT-S & $12$, $\gamma{=}2$ & $+1.103$ & $\mathbf{+1.331}$ & $-0.228$ \\
\midrule
ViT-B  & $4$, $\gamma{=}1$  & $+0.348$ & $\mathbf{+0.685}$ & $-0.337$ \\
ViT-B  & $4$, $\gamma{=}2$  & $+0.576$ & $\mathbf{+1.238}$ & $-0.662$ \\
ViT-B  & $8$, $\gamma{=}1$  & $+0.991$ & $\mathbf{+1.325}$ & $-0.334$ \\
ViT-B  & $8$, $\gamma{=}2$  & $+1.567$ & $\mathbf{+2.111}$ & $-0.544$ \\
ViT-B  & $12$, $\gamma{=}1$ & $+1.774$ & $\mathbf{+1.780}$ & $-0.006$ \\
ViT-B  & $12$, $\gamma{=}2$ & $+1.224$ & $\mathbf{+1.249}$ & $-0.025$ \\
\bottomrule
\end{tabular}
\end{table}

\paragraph{The gain is recovery, not new correctness.} Table~\ref{tab:channels} gives the
decomposition. In every one of the twelve cells the recovery channel exceeds the net difference and
the new-correctness channel is negative, so late concentration actually concedes ground on cases the
unreduced model was already getting wrong and wins purely by preserving the ones it was getting
right. At the main operating point ($r_{\mathrm{base}}{=}8$, $\gamma{=}2$) the unreduced DeiT-S is
correct on $295{,}479$ of the $750{,}000$ cases. The flat schedule destroys $8.5\%$ of those
predictions ($25{,}017$) and ours destroys $3.9\%$ ($11{,}531$), at no extra compute; the
intermediate $\gamma{=}1$ sits between them at $5.3\%$. On ViT-B the rates are $8.0\%$ and $4.2\%$
of $416{,}401$ correct predictions.

\begin{figure}[ht]
\centering
\includegraphics[width=0.95\linewidth]{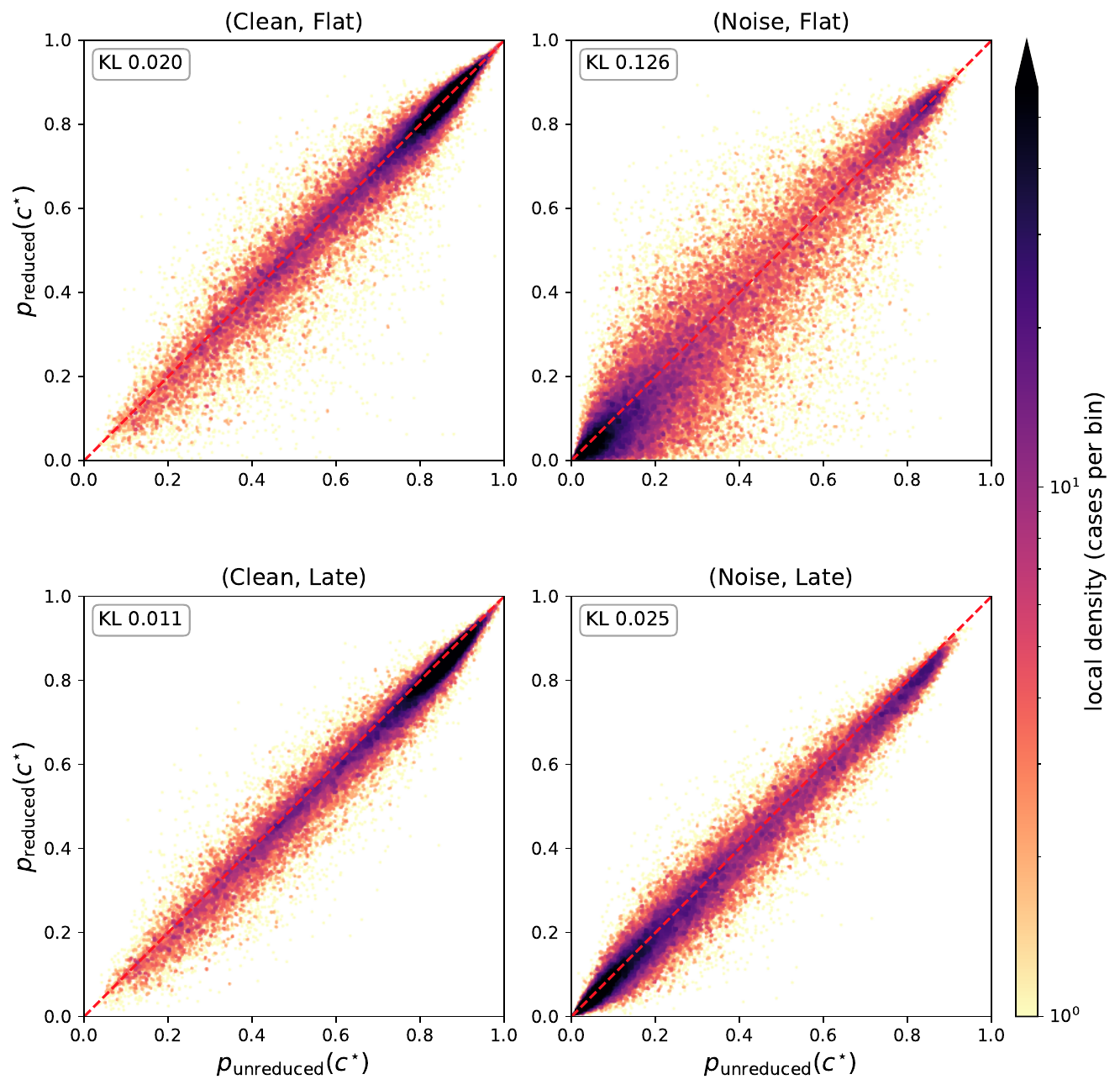}
\caption{Per-image agreement with the unreduced model (DeiT-S, $r_{\mathrm{base}}{=}8$, iso-GFLOPs,
  $50{,}000$ images per panel). Horizontal axis: the probability the unreduced model assigns to its own
  predicted class $c^{\star}$; vertical axis: the probability the reduced model assigns to that same
  class; exact agreement lies on the dashed line. Color is the local density of cases, on a shared
  scale across panels, saturating above its $75$th percentile. Reduction is nearly harmless on clean
  input under either schedule and remains so for ours under corruption; the flat schedule departs from
  the unreduced model only where reduction and corruption meet. KL is the mean per-case
  $\mathrm{KL}(p_{\text{unreduced}}\,\|\,p_{\text{reduced}})$ over the full class distribution on the
  same input. Gaussian noise at severity~5 is shown for legibility; the aggregate over all $15$
  corruptions is quoted above.}
\label{fig:predagree}
\end{figure}

\paragraph{The damage is a subpopulation, not a uniform drift.} The same fact is visible per image.
Figure~\ref{fig:predagree} plots, for each case, the probability the unreduced model assigns to its
own predicted class $c^{\star}$ against the probability the reduced model assigns to that same
class, so exact agreement lies on the diagonal. Most cases sit on the diagonal under either
schedule; what distinguishes them is a subpopulation that the flat schedule scatters off it.
Averaged over the $15$ corruptions the mean per-case
$\mathrm{KL}(p_{\mathrm{unreduced}}\,\|\,p_{\mathrm{reduced}})$ is $0.085$ for flat and $0.029$ for
ours, and the flat schedule changes the unreduced model's top-1 prediction on $19.1\%$ of cases
against ours on $10.8\%$ (ViT-B: $17.4\%$ and $11.0\%$), with ours lower on $15/15$ corruptions for
both backbones. The panels also show that this is specific to the corrupted condition: on clean
input the two schedules are equally close to the unreduced model, consistent with the
clean-to-corrupted ratio of Section~\ref{sec:severity}.

\section{Calibration: Per-Cell Detail (gaussian noise)}
\label{app:calib}

Figure~\ref{fig:calib} summarizes the iso-GFLOPs calibration; the tables below list the per-cell
numbers. Table~\ref{tab:calib-appendix} gives the gaussian-noise sweep; the 15-corruption mean sweep
for all six operating points is the body Table~\ref{tab:main}. $\Delta$ = late $-$ flat. The gain is
an inverted-U in $\gamma$ whose optimum shifts to higher $\gamma$ as $r_{\mathrm{base}}$ decreases,
and over-concentration begins to hurt (the optimum drops below $\gamma{=}1.5$) at the most
aggressive settings we report ($r_{\mathrm{base}}{=}14$). Table~\ref{tab:isocalib} lists the exact
calibrated schedules used throughout the paper, confirming that every late-concentrated operating
point sits at or below its flat baseline's GFLOPs.

\begin{table}[ht]
\centering
\caption{Calibrated iso-GFLOPs schedules used throughout the paper. For each operating point
  $r_{\mathrm{base}}$ and exponent $\gamma$ we list the late-concentrated per-layer-mean removal budget
  $\bar{r}$, chosen so the schedule's GFLOPs do not exceed the flat (const $r_{\mathrm{base}}$)
  baseline. Parentheses give the GFLOPs deviation from flat (DeiT-S, THOP): \emph{every entry is
  $\le 0$}, so every late schedule is at most as expensive as its flat counterpart (strictly
  conservative). $\bar{r}$ is backbone-independent (197 tokens, 12 layers at 224px), so the
  same schedules apply to DeiT-S/B and ViT-S/B; the flat column shows DeiT-S GFLOPs. The larger
  under-shoots at small $r_{\mathrm{base}}$ with high $\gamma$ (e.g.\ $r_{\mathrm{base}}{=}6$,
  $\gamma{=}3$) are an integer-granularity effect and remain conservative.}
\label{tab:isocalib}
\small\setlength{\tabcolsep}{4pt}
\begin{tabular}{r l rrrrrr}
\toprule
$r_{\mathrm{base}}$ & flat (GFLOPs) & $\gamma{=}0.5$ & $\gamma{=}1.0$ & $\gamma{=}1.5$ & $\gamma{=}2.0$ & $\gamma{=}2.5$ & $\gamma{=}3.0$ \\
\midrule
2 & 2 (3.99G) & 3 ($-1.6\%$) & 3 ($-0.0\%$) & 4 ($-1.4\%$) & 4 ($-0.4\%$) & 5 ($-1.3\%$) & 5 ($-0.6\%$) \\
4 & 4 (3.72G) & 5 ($-0.7\%$) & 6 ($-1.1\%$) & 7 ($-0.7\%$) & 8 ($-0.9\%$) & 9 ($-1.1\%$) & 10 ($-1.4\%$) \\
6 & 6 (3.46G) & 8 ($-2.7\%$) & 9 ($-1.2\%$) & 10 ($-0.2\%$) & 13 ($-3.6\%$) & 16 ($-5.2\%$) & 20 ($-7.6\%$) \\
8 & 8 (3.20G) & 10 ($-1.2\%$) & 12 ($-2.0\%$) & 14 ($-1.8\%$) & 16 ($-1.6\%$) & 18 ($-0.5\%$) & 21 ($-0.7\%$) \\
10 & 10 (2.94G) & 13 ($-3.6\%$) & 15 ($-2.2\%$) & 17 ($-0.6\%$) & 21 ($-1.2\%$) & 26 ($-1.8\%$) & 33 ($-2.9\%$) \\
12 & 12 (2.67G) & 15 ($-2.2\%$) & 18 ($-1.6\%$) & 22 ($-0.4\%$) & 27 ($-0.3\%$) & 34 ($-0.1\%$) & 44 ($-0.4\%$) \\
\bottomrule
\end{tabular}
\end{table}

\begin{table}[ht]
\centering
\caption{iso-GFLOPs calibration on gaussian noise (sev=5, 50k), DeiT-S. const rows are flat
  baselines; late rows are calibrated to the matching flat GFLOPs. $\Delta$ = late $-$ flat.}
\label{tab:calib-appendix}
\begin{tabular}{rlrr}
\toprule
$r_{\mathrm{base}}$ & schedule & gaussian top-1 & $\Delta$ \\
\midrule
 6 & const             & 32.35 & -- \\
 6 & late $\gamma{=}2.0$ & 33.84 & $+$1.49 \\
 8 & const             & 31.62 & -- \\
 8 & late $\gamma{=}1.0$ & 33.12 & $+$1.50 \\
 8 & late $\gamma{=}1.5$ & 33.49 & $+$1.86 \\
 8 & late $\gamma{=}2.0$ & 33.79 & $+$2.17 \\
10 & const             & 30.73 & -- \\
10 & late $\gamma{=}1.0$ & 32.96 & $+$2.23 \\
10 & late $\gamma{=}1.5$ & 33.25 & $+$2.52 \\
10 & late $\gamma{=}2.0$ & 33.44 & $+$2.71 \\
10 & late $\gamma{=}2.5$ & 33.36 & $+$2.63 \\
12 & const             & 29.68 & -- \\
12 & late $\gamma{=}0.5$  & 31.41 & $+$1.72 \\
12 & late $\gamma{=}0.75$ & 32.15 & $+$2.47 \\
12 & late $\gamma{=}1.0$  & 32.33 & $+$2.65 \\
12 & late $\gamma{=}1.5$  & 32.64 & $+$2.96 \\
12 & late $\gamma{=}2.0$  & 32.72 & $+$3.04 \\
12 & late $\gamma{=}2.5$  & 32.62 & $+$2.93 \\
12 & late $\gamma{=}3.0$  & 32.44 & $+$2.75 \\
14 & const             & 28.51 & -- \\
14 & late $\gamma{=}1.0$ & 31.35 & $+$2.84 \\
14 & late $\gamma{=}1.5$ & 30.99 & $+$2.48 \\
\bottomrule
\end{tabular}
\end{table}

Across operating points the 15-corruption mean optimum migrates from $\gamma\!\approx\!3$ at the
lightest $r_{\mathrm{base}}{=}2$ to $\gamma\!\approx\!1.5$ at the aggressive
$r_{\mathrm{base}}{=}12$ (Table~\ref{tab:main}), exactly as in Figure~\ref{fig:calib}c. The default
$\gamma{=}2$ is the per-point optimum at the main $r_{\mathrm{base}}{=}8$ (and
$r_{\mathrm{base}}{=}6$) and stays within $0.13$pp of the per-point optimum at every operating point
we test, which is why a single static $\gamma{=}2$ suffices without per-shift tuning.

\section{Wall-clock Latency at iso-GFLOPs}
\label{app:latency}

We verify that the iso-GFLOPs match translates into matched real latency, addressing the concern
that analytic FLOPs need not predict wall-clock speed. Table~\ref{tab:latency} reports throughput at
the main operating point ($r_{\mathrm{base}}{=}8$, batch~64) on a single GPU, averaged over $200$
timed iterations after $30$ warmup iterations. Latency is data-independent (the per-layer token
counts are fixed by the schedule, not by the input). The late ($\gamma{=}2$) and flat schedules are
wall-clock-indistinguishable (within $0.5\%$), so every iso-GFLOPs comparison in the paper is also
an iso-latency comparison; token reduction itself speeds up inference over the uncompressed model at
this throughput-oriented batch size.

\begin{table}[ht]
\centering
\caption{Wall-clock latency at the main operating point ($r_{\mathrm{base}}{=}8$), batch~64, single GPU
  ($200$ timed iterations, $30$ warmup). At iso-GFLOPs the late ($\gamma{=}2$) and flat schedules match in
  throughput to within $0.5\%$. The $384$-dim row covers DeiT-S and ViT-S; the $768$-dim row covers DeiT-B
  and ViT-B (identical architecture, hence identical latency).}
\label{tab:latency}
\small
\begin{tabular}{l l r r r}
\toprule
backbone & schedule & GFLOPs & ms / batch & img / s \\
\midrule
$d{=}384$ (DeiT-S, ViT-S) & uncompressed        & 4.25 & 67.1 & 954 \\
                          & flat (const $r{=}8$) & 3.20 & 54.6 & 1172 \\
                          & late ($\gamma{=}2$)  & 3.15 & 54.6 & 1172 \\
\midrule
$d{=}768$ (DeiT-B, ViT-B) & uncompressed        & 16.86 & 212.0 & 302 \\
                          & flat (const $r{=}8$) & 12.67 & 167.0 & 383 \\
                          & late ($\gamma{=}2$)  & 12.46 & 167.5 & 382 \\
\bottomrule
\end{tabular}
\end{table}

\end{document}